\documentclass[twoside]{article}

\usepackage[preprint]{aistats2027}

\usepackage{color}
\usepackage{algpseudocode}
\usepackage[skins]{tcolorbox}

\usepackage[round]{natbib}

\usepackage{xcolor}

\definecolor{mycyan}{RGB}{0,204,204}

\usepackage{hyperref}
\usepackage{amssymb}
\usepackage{amsthm}
\usepackage{mathtools}
\usepackage{mathrsfs}
\usepackage{graphicx}
\usepackage{subcaption}
\usepackage[space]{grffile}
\usepackage{url}
\usepackage{bm}
\usepackage{booktabs}
\usepackage{algorithm}
\usepackage{algpseudocode}
\usepackage{comment}
\usepackage{float}

\newtheorem{assumption}{Assumption}[section]
\newtheorem{theorem}{Theorem}[section]
\newtheorem{Lemma}{Lemma}[section]

\newtheorem{definition}{Definition}
\newtheorem{corollary}{Corollary}[section]
\newtheorem{remark}{Remark}[section]

\newcommand{\R}{\mathbb{R}}
\newcommand{\E}{\mathbb{E}}
\newcommand{\PP}{\mathbb{P}}
\newcommand{\St}{\mathcal{S}}
\newcommand{\Ac}{\mathcal{A}}
\newcommand{\bt}{\boldsymbol{\theta}}
\newcommand{\bv}{\boldsymbol{v}}
\newcommand{\Bbar}{B_{\eta}}
\newcommand{\Gbar}{\bar G}
\newcommand{\Lv}{L_{V}}
\newcommand{\Mv}{M_{V}}
\newcommand{\hb}{\widehat{\boldsymbol h}}
\newcommand{\hg}{\widehat{\boldsymbol g}}
\newcommand{\sinv}{\bar\sigma^{-1}}
\newcommand{\Iset}{\mathcal{I}}
\newcommand{\wc}{w_c}
\newcommand{\wg}{w_g}
\newcommand{\iot}{\iota_\delta}

\usepackage{pifont}
\newcommand{\DefinedAs}{\mathrel{\mathop:}=}

\begin{document}



\twocolumn[

\aistatstitle{Provable Zeroth-Order Safe Exploration for\\ Reinforcement Learning from Human Feedback}
\aistatsauthor{ Weiqin Chen \And Xinmeng Huang \And  Dongsheng Ding \And  Santiago Paternain }

\aistatsaddress{ Purdue University \And UPenn \And University of Tennessee, Knoxville \And RPI} ]

\begin{abstract}
Safe reinforcement learning from human feedback (Safe-RLHF) seeks to improve helpfulness while satisfying harmlessness constraints. 
% Existing primal-dual approaches can converge to feasible policies but do not prevent constraint violations during training. 
Existing primal-dual approaches can converge to feasible policies, but generally do not guarantee constraint satisfaction throughout training.
We study a stronger requirement, safe exploration for RLHF, under which {\bf every} policy executed during training, including auxiliary exploration policies, remains feasible.
We propose Zeroth-Order Log-Barrier Policy Gradient (ZO-LBPG), a reward-model-free method that optimizes a log-barrier objective using zeroth-order estimates derived directly from human preferences. To ensure safety under noisy feedback, we combine conservative constraint margin estimates with coupled rollouts and an adaptive step size that prevents updates from crossing constraint boundaries.
% We prove that, with high probability, all iterates and exploration policies remain strictly feasible, and that ZO-LBPG globally converges to an approximately optimal constrained policy.
We establish a high-probability guarantee that all iterates and auxiliary exploration policies remain strictly feasible throughout training, and that ZO-LBPG globally converges to an approximately optimal constrained policy.
We establish a sample complexity of $\widetilde{\mathcal O}(\varepsilon^{-6}+d^3\varepsilon^{-5})$ trajectories and $\widetilde{\mathcal O}(\varepsilon^{-8}+d^5\varepsilon^{-7})$ human queries. 
% Experiments on a synthetic CMDP and a language-model alignment task corroborate the  training-time safety guarantees.
Experiments via a synthetic setup and a language-model alignment task corroborate the  training-time safety.

\end{abstract}

% \weiqin{Plan for 8 Pages: abs+intro: 1.5, SAFEX-RLHF formulation: 1.25, ZO-LBPG Algo: 2, theoretical guarantees: 2, experiment+conclusion: 1.25}
%%%%%%%%%%%%%%%%%%%%%%%%%%%%%%%%%%%%%%%%%%%%%%%%%%%%%%%%%%%%%%%%%%%%%%%%%%%%

% \vspace{-2mm}
\section{Introduction}
\label{sec:intro}

% \vspace{-10mm}
%

Reinforcement learning from human feedback (RLHF) has become a standard approach for aligning language models and other interactive agents with human preferences~\citep{christiano2017deep,ouyang2022training,bai2022training}. A central objective of RLHF is to improve helpfulness, but optimizing helpfulness alone can produce undesirable or harmful behavior. Safe-RLHF addresses this issue by formulating alignment as a constrained reinforcement-learning problem: maximize helpfulness subject to one or more harmlessness requirements, such as limits on toxicity, deception, or privacy violations~ \citep{dai2023safe}.

% Reinforcement learning from human feedback (RLHF) is widely used to align agents such as large language models (LLMs) with human preferences \citep{christiano2017deep,ouyang2022training,bai2022training}. Optimizing helpfulness alone, however, can produce responses that are helpful yet harmful \citep{dai2023safe}.

Recent work has established global convergence guarantees for Safe-RLHF without explicitly learning reward models~\citep{liu2026policy}. In particular,  formulate Safe-RLHF as an infinite-horizon constrained Markov decision process (CMDP) and estimate policy gradients directly from human feedback using zeroth-order perturbations. Their primal-dual method can converge to a feasible solution while avoiding explicit reward and utility models. This reward-model-free formulation is attractive because learned reward models can be vulnerable to reward hacking and other forms of misspecification~\citep{casper2023open,wang2026reward,helff2026llms}.

% Safe-RLHF \citep{dai2023safe} addresses this problem by decoupling helpfulness and harmlessness: it maximizes helpfulness subject to harmlessness constraints, e.g., toxicity, deception, and privacy violations, each with its own threshold. In this setting, \citet{liu2026policy} establish global convergence guarantees for Safe-RLHF using an infinite-horizon constrained Markov decision process (CMDP) without reward modeling. They estimate the primal-dual policy gradient via the discrepancy from human preferences directly~\citep{zhang2024zeroth}. Avoiding explicit reward models is appealing because learned reward models suffer from reward hacking~\citep{casper2023open,wang2026reward,helff2026llms}.

{\bf However, convergence to a feasible policy does not guarantee safety during training.} A primal-dual method~\citep{zhang2026alignment, liu2026policy} may temporarily violate a constraint while its dual variables adapt. In conventional constrained optimization, such violations may be acceptable if only the final solution matters. In RLHF, however, policies encountered during training may themselves be evaluated, inspected, or deployed. Moreover, zeroth-order RLHF methods execute auxiliary perturbed policies to obtain human feedback. Thus, even if the current iterate is feasible, the policies used to estimate its gradient may be unsafe. These considerations motivate a stronger requirement, which we call \underline{Safe} e\underline{X}ploration for RLHF (SafeX-RLHF): every policy executed during training, including both optimization iterates and auxiliary exploration policies, satisfy all harmlessness constraints.

% Convergence to a feasible policy does not by itself ensure safety during training. Existing Lagrangian-based approaches to safe-RLHF include primal--dual policy gradient methods~\citep{liu2026policy} and iterative dual-based methods~\citep{zhang2026alignment}. They all lack the safety guarantee for every intermediate policy iterate in the training. This is unsuitable for safety-critical scenarios, where intermediate checkpoints are routinely exposed to human evaluators for evaluation and may sometimes be deployed. We therefore seek for a stronger property, termed as  \underline{Safe} e\underline{X}ploration for RLHF (SafeX-RLHF). It requires \emph{every} policy executed during training, including the auxiliary policies queried for optimization purposes, to satisfy all harmlessness constraints.

Safe exploration has been studied in constrained reinforcement learning using interior-point methods. Log-barrier objectives provide a natural mechanism: as a constraint margin approaches zero, the barrier gradient repulses increasingly, while a conservative step size prevents the update from crossing the boundary~\citep{usmanova2024log}. \cite{ni2025safe}, for example, establish global convergence for log-barrier policy optimization in CMDPs with explicit reward signals. Extending this approach to SafeX-RLHF, however, introduces two fundamental difficulties.

% Safe exploration is well understood when explicit reward signals are available. Interior-point (log-barrier) methods \citep{usmanova2024log,ni2025safe} keep every iterate strictly feasible by shrinking the step near a constraint boundary and by exploiting the repulsion of the barrier gradient, and \citet{ni2025safe} establish global convergence for CMDPs with $\widetilde{\mathcal O}(\varepsilon^{-6})$ samples.
%

First, the objective and constraint values are not directly observed. RLHF provides only noisy binary judgments. Both the helpfulness gradient and the harmlessness gradients must be estimated from human preferences, while the constraint margins themselves must be inferred from absolute safety judgments. The latter is particularly challenging because the margins appear in the denominator of the log-barrier gradient. An error in a constraint margin is consequently amplified as the policy approaches the boundary.

%\vspace{-1mm}
Second, a zeroth-order gradient estimate must be accurate enough to certify an individual update, rather than merely being accurate on average~\citep{zhang2024zeroth}.
Existing zeroth-order RLHF methods perturb the policy along a single random direction at each iteration. 
Although such an estimator is unbiased after averaging over the random direction, an individual estimate can deviate substantially from the true gradient. Such an estimate is insufficient for certifying that a particular policy update, or the auxiliary policies used to obtain it, remains safe.

%\vspace{-1mm}
We address these challenges with \underline{Z}eroth-\underline{O}rder \underline{L}og-\underline{B}arrier \underline{P}olicy \underline{G}radient (ZO-LBPG). The method combines three ingredients. We first estimate constraint margins from absolute harmlessness judgments and use lower confidence bounds in the barrier gradient, which conservatively strengthens the repulsion from each constraint boundary. We then estimate reward and harmlessness gradients using independently sampled perturbation directions and coupled trajectory pairs. Finally, we construct an adaptive step size from confidence bounds on both the current constraint margins and their directional changes, ensuring that no update can cross a constraint boundary. This combination allows us to establish high-probability safety simultaneously for the iterates and for all auxiliary policies executed to collect human feedback. Our main contributions are as follows.

% Extending the log-barrier framework from CMDPs to SafeX-RLHF is non-trivial. First, reward and utility signals are absent with only binary judgments from noisy evaluators are observed. Although zeroth-order methods can estimate policy gradients by inverting the link function \citep{zhang2024zeroth}, the resulting bias must be controlled at every iterate rather than averaged out. Moreover, the zeroth-order estimators used in RLHF \citep{zhang2024zeroth,liu2026policy} perturb the policy along only a single random direction at each iteration. While such estimators are unbiased in expectation, an individual estimate can be far from the true gradient, making it insufficient to certify the safety of a particular update. To address these entangled challenges, we propose a \underline{Z}eroth-\underline{O}rder \underline{L}og-\underline{B}arrier \underline{P}olicy \underline{G}radient method (ZO-LBPG) for achieving SafeX-RLHF. Our contributions are summarized as follows.
\begin{itemize}
\vspace{-3mm}
    \item {\bf SafeX-RLHF algorithm.} We introduce ZO-LBPG, a reward-model-free zeroth-order method for safe exploration with multiple harmlessness constraints. The method estimates barrier gradients directly from pairwise preferences and absolute safety judgments, and uses confidence-aware step sizes to preserve feasibility.
\vspace{-2mm}
    \item {\bf Training-time safety guarantees.} We prove that, under a feasible initialization and an extended MFCQ condition, every iterate remains strictly feasible with high probability. Moreover, every perturbed policy executed for gradient estimation remains feasible as well. Thus, the guarantee applies to the entire data-collection process, rather than only to the final policy.
\vspace{-1mm}
    \item {\bf Global convergence.} Under standard RL assumptions, we establish global convergence to an approximately optimal constrained policy. The resulting sample complexity is \(\widetilde {\mathcal O}(\epsilon^{-6}+d^3\epsilon^{-5})\) trajectories and \(\widetilde {\mathcal O}(\epsilon^{-8}+d^5\epsilon^{-7})\) human queries. The leading terms are independent of the policy dimension and match the trajectory complexity of safe exploration with explicit reward and utility signals~\citep{ni2025safe}.
    \vspace{-1mm}
    \item {\bf Empirical validation.} 
    % Experiments on a synthetic CMDP and a language-model alignment task demonstrate the distinction between training-time safety and eventual feasibility. 
    Experiments on a synthetic CMDP and a language-model alignment task illustrate the practical distinction between training-time safety and eventual feasibility.
    ZO-LBPG maintains strictly positive constraint margins throughout training, whereas the primal-dual baseline repeatedly crosses the constraint boundary while achieving a smaller reward gap.
\end{itemize}

\vspace{-3.5mm}
\section{SafeX-RLHF}
\label{sec:prelim}
\vspace{-2mm}
%
%
%
% \blue{SP: Maybe I’m missing something about this
% formulation. If the state is a text/tokens and it is
% allow to grow at each timestep. How is it that the
% MDP has finite states? Is there a hard bound on
% how big this could grow? If this is starndard in the
% literature I suggest we cite other works.} \weiqin{An LLM has a finite vocabulary and a finite context window, and generation stops at the end-of-sequence token or when the context length is reached. So the set of all possible prompts, histories and responses is finite, and the horizon is bounded. This is the standard way of modeling LLM generation as an MDP in the RLHF theory literature, and it is exactly the CMDP model of \citep{liu2026policy} that we build on.} \blue{SP: Understood, the absorbing states fix this. }
%
%
%
\textbf{CMDP.} We formulate Safe-RLHF as an infinite-horizon discounted CMDP \citep{liu2026policy}, namely $(\St,\Ac,P,r,\{g_i\}_{i\,=\,1}^m,\{b_i\}_{i\,=\,1}^m,\gamma,\rho)$ with finite state space $\St$ (prompts and dialogue histories) and action space $\Ac$ (responses), transition kernel $P$, reward (helpfulness) $r$: $\St\times\Ac\to[0,1]$, costs (harmlessness) $g_i$: $\St\times\Ac\to[0,1]$, constraint thresholds $b_i$, discount factor $\gamma\in(0,1)$ and initial distribution $\rho$. Finiteness follows from the finite vocabulary and context window of an LLM, with the end of a generation modeled by an absorbing state. A policy $\pi$ induces trajectories $\tau=\{(s_t,a_t)\}_{t\,\ge\,0}$ with $s_0\sim\rho$, $a_t\sim\pi(\cdot \,\vert\,  s_t)$, $s_{t+1}\sim P(\cdot \,\vert\,  s_t,a_t)$. For $H\in\mathbb N\cup\{\infty\}$ and $\diamond\in\{r\}\cup\{g_i\}_{i\,=\,1}^m$, define the length-$H$ return and value function
\begin{align}
R_\diamond^H(\tau)\DefinedAs\sum_{t\,=\,0}^{H-1}\gamma^t\diamond(s_t,a_t), ~ V_\diamond^{\pi,H}(\rho)\DefinedAs\E_{\tau\sim\pi}\big[R_\diamond^H(\tau)\big], \notag
\end{align}
and let $V^\pi_\diamond(\rho) \DefinedAs V^{\pi,\infty}_\diamond(\rho)$. Since $\diamond\in[0,1]$, $R^H_\diamond(\tau)\in[0,G(H)]$ with $G(H) \DefinedAs \frac{1-\gamma^H}{1-\gamma}$ and $V^\pi_\diamond(\rho)\in[0,\Gbar]$ with $\Gbar \DefinedAs \frac{1}{1-\gamma}$. Without loss of generality, let $b_i\in(0,\Gbar)$.

\textbf{Problem.} Let $\{\pi_{\bt}:\bt\in\Theta\}$ be a parameterized policy class with $\Theta\subseteq\R^d$ (e.g., softmax, $\pi_{\bt}(a \,\vert\,  s)\propto e^{\theta_{s,a}}$, $\Theta=\R^{|\St||\Ac|}$). We take $\Theta=\R^d$ and write $V^{\bt}_\diamond(\rho)\DefinedAs V^{\pi_{\bt}}_\diamond(\rho)$. Safe-RLHF seeks a policy that maximizes helpfulness subject to the harmlessness constraints,
\begin{equation}
\label{eq:primal}
\max_{\bt\,\in\,\Theta}\ V_r^{\bt}(\rho)
\quad
\text{s.t.}
\quad 
V_{g_i}^{\bt}(\rho)
\;\ge\; 
b_i,\ i\in[m]. \tag{OP}
\end{equation}
Denote by $\bar V^{\bt}_{g_i}(\rho) \DefinedAs V^{\bt}_{g_i}(\rho)-b_i$ the constraint margins. SafeX-RLHF distinguishes from standard CMDPs and Safe-RLHF in two aspects: (i) rewards and utilities are never observed; and (ii) every policy executed during training must be feasible. We formalize them below.

\vspace{-0.5mm}
\textbf{Human feedback.} RLHF does not return explicit reward or utility signals, and all information about $r$ and $g_i$ comes from binary human judgments. Following the discrete-choice model of \citet{ben1985discrete}, a judgment about two trajectories $\tau_0,\tau_1$ is related to their latent returns through a \emph{link function}, which maps the return difference to the probability of the observed choice. We consider three channels of one--bit feedback: pairwise helpfulness and harmlessness $o_\diamond$, and absolute harmlessness $o_{c_i}$, each with its own monotone link $\sigma_\diamond$, or $\sigma_{c_i}$: $\R\to[0,1]$,
\vspace{-0.5mm}
\begin{align}\label{eq:channels}
% &\PP(o_r=1)
% \; = \;
% \sigma_r\big(R^H_r(\tau_1)-R^H_r(\tau_0)\big), \notag \\
&\PP(o_{\diamond}=1)
\; = \; \sigma_{\diamond}\big(R^H_{\diamond}(\tau_1)-R^H_{\diamond}(\tau_0)\big), \\
&\PP(o_{c_i}=1)
\; = \;
\sigma_{c_i}\big(R^H_{g_i}(\tau)-b_i\big), \notag
\setlength{\belowdisplayskip}{2pt}
\end{align}
for $i\in[m]$. The absolute channel measures whether a trajectory satisfies the $i$th constraint, which is equivalent to a comparison with a virtual trajectory of return $b_i$~\citep{chatterji2021theory}, and primal--dual methods rely on the same channel in their dual updates \citep{liu2026policy}. The most common link in RLHF is the B-T model $\sigma(x)=1/(1+e^{-x})$ \citep{bradley1952rank}. Our analysis is not tied to this specific choice and holds for the broader class below.

%

% \blue{SP: Why not write all assumptions in terms of $\sigma$ instead of the inverse?. If strictly increasing and twice
% continuous diff then the inverse also has that property. Four the boundeedness I have some problems,
% not so much with the assumption but with the example. Given $L$ I can always find a $p$ close enough to 0
% or 1 that makes $\frac{1}{p(1-p)} > L$. If the strict containment of $P_\sigma$ in $(0,1)$ is there to avoid this point it
% needs to be clear. It could be strictly contained but
% the inf or the sup could still be 0 or 1. Because the
% $1/(1+e^{-x})$ definition of the example it is clear that this can’t be. But the connection is not super clear.} \weiqin{We state it on the inverse because that is what the algorithm and proofs use (we invert frequencies), and a strictly increasing twice-differentiable link does not guarantee a differentiable inverse (e.g., $x^3$ at 0) unless its derivative is bounded away from zero, which is exactly the Lipschitz condition on the inverse. Boundedness is not an issue because $\mathcal P_\sigma=[\sigma(-\bar G),\sigma(\bar G)]$ is a closed interval with endpoints strictly inside $(0,1)$ (return differences are bounded by $\bar G$), so $p$ near 0 or 1 never occurs and for Bradley–Terry $L$ is finite, of order $e^{\bar G}$. Please feel free to make it more clear though.} \blue{SP: I'm okay with this. Maybe worth making explicit in the assumption that $\underline{p}>0$ and $\bar{p}<1$. I'm not sure if you used the monotonicity of $\sigma$ anywhere else. }
%
%
\vspace{-0.5mm}
\begin{assumption}[Link functions]
\label{assum:link}
Each link $\sigma$ in~\eqref{eq:channels} is strictly increasing on $\bar{\mathcal I} \DefinedAs [-\Gbar,\Gbar]$, and its inverse $\sigma^{-1}$ is twice continuously differentiable on $\mathcal P_\sigma \DefinedAs \sigma(\bar{\mathcal I})= [\underline{p}, \overline{p}]$ with $|(\sigma^{-1})'|\le L$ and $|(\sigma^{-1})''|\le L_2$ on $\mathcal P_\sigma$, where $\underline{p} \DefinedAs \sigma(-\Gbar)$ and $\overline{p} \DefinedAs \sigma(\Gbar)$.
\end{assumption}
\vspace{-1mm}
All return differences and threshold-adjusted returns in~\eqref{eq:channels} lie in $\bar{\mathcal I}$. In the B-T model, $(\sigma^{-1})'(p)=\frac{1}{p(1-p)}$ and $(\sigma^{-1})''(p)=\frac{2p-1}{p^2(1-p)^2}$ are bounded on $\mathcal P_\sigma\subset(0,1)$. 
%
% The bound on the second derivative is the only requirement beyond those of \citet{zhang2024zeroth,liu2026policy}, and it is what allows a bias of order $1/M$ in Section~\ref{sec:estimators}.

\vspace{-0.5mm}
\textbf{Sampler.} We control the randomness with which our policies are executed, as in the case of LLM response generation, and formalize this by a seeded sampler.
\vspace{-0.5mm}
\begin{assumption}[Seeded sampler]
\label{assum:seeds}
Trajectories are generated by a sampler that maps a seed $\omega=(s_0,(u_h,u'_h)_{h<H})$ with i.i.d.\ uniform components to a trajectory $\tau(\bt,\omega)$ through $a_h=F^{-1}_{\pi_{\bt}(\cdot \,\vert\,  s_h)}(u_h)$ (inverse-CDF sampling for a fixed ordering of $\Ac$) and $s_{h+1}=\Phi(s_h,a_h,u'_h)$, where $\Phi(s,a,U)\sim P(\cdot \,\vert\,  s,a)$ for $U\sim\mathrm{Unif}[0,1]$. A coupled pair consists of two trajectories generated from a common seed under two different parameters $\tau(\bt,\omega)$ and $\tau(\bt',\omega)$.
\end{assumption}
\vspace{-1mm}
For an LLM, $u_h$ drives token sampling and the transitions are deterministic, so Assumption~\ref{assum:seeds} amounts to sampling the two responses of a pair with the same random seed. Each trajectory of a coupled pair has the same marginal distribution as an independently sampled one. The coupling only makes the two trajectories coincide until the first step at which the two policies sample different actions, which is what reduces the variance of their return difference in Section~\ref{sec:estimators}.

\vspace{-1mm}
\textbf{Safe exploration.} Solving \eqref{eq:primal} from human feedback requires executing policies to collect judgments, and some methods, in particular zeroth-order methods, execute not only the iterates but also perturbed policies whose responses are shown to the evaluators. We thus require every such policy to be feasible.
\begin{definition}[Safe exploration]
\label{def:safe}
% An algorithm ensures safe exploration if every policy it executes to collect data (such as the iterates $\pi_{\bt^{(t)}}$ and any auxiliary policies) 
% satisfies the constraint $i$ in \eqref{eq:primal}, $\forall i\in[m]$.
An algorithm ensures safe exploration if every policy it executes to collect data (including the iterates \(\pi_{\theta^{(t)}}\) and any auxiliary policies) satisfies all constraints in (OP).
\end{definition}
\vspace{-2mm}
This requirement is much stronger than constraint satisfaction on average \citep{liu2026policy}, and it strengthens the notion of safety in \citet{ni2025safe}, which concerns the iterates only and assumes explicit reward and utility signals. Safe exploration needs a safe starting point. As is standard \citep{liu2026policy}, we assume that one is known, for instance a supervised fine-tuned (SFT) reference model that meets every threshold with some margin \citep{dai2023safe}.
\begin{assumption}[Feasible initialization]
\label{assum:slater}
There exist a known parameter $\bt^{(0)}\in\Theta$ and $\nu>0$ such that $\bar V^{\bt^{(0)}}_{g_i}(\rho)\ge\nu$ for all $i\in[m]$.
\end{assumption}

\vspace{-1mm}
\textbf{Regularity of the value functions.} To keep every executed policy feasible, the algorithm must know how much a value function can change when the policy parameter moves, both to bound the effect of an update on the constraint margins and to control the error of the zeroth-order gradient estimates. Both are governed by Lipschitz and smoothness constants of the value functions, which follow from the standard policy smoothness condition \citep{Xu2020Sample,agarwal2021theory,yuan2022general}. The condition holds for softmax, log-linear and clipped Gaussian policies.
\begin{assumption}[Policy smoothness]
\label{assum:smooth}
There exist $M_g,M_h>0$ such that $\|\nabla_{\bt}\log\pi_{\bt}(a \,\vert\,  s)\|\le M_g$ and $\|\nabla^2_{\bt}\log\pi_{\bt}(a \,\vert\,  s)\|\le M_h$ for all $\bt\in\Theta$, $s\in\St$, $a\in\Ac$.
\end{assumption}
\begin{Lemma}[Value regularity]
\label{lem:smooth}
Under Assumption~\ref{assum:smooth}, for every $\diamond$ and $H\in\mathbb N\cup\{\infty\}$, $\bt\mapsto V^{\bt,H}_\diamond(\rho)$ is $\Lv$-Lipschitz and $\Mv$-smooth on $\Theta$ with
%\begin{align}
$
\Lv \DefinedAs \frac{M_g}{(1-\gamma)^{3/2}},\, \Mv \DefinedAs \frac{M_g^2+M_h}{(1-\gamma)^2},
$
%\end{align}
and the truncation errors satisfy $0\le V^{\bt}_\diamond(\rho)-V^{\bt,H}_\diamond(\rho)\le\epsilon_V(H)$ and $\|\nabla_{\bt}V^{\bt}_\diamond(\rho)-\nabla_{\bt}V^{\bt,H}_\diamond(\rho)\|\le\epsilon_\nabla(H)$, where
% \begin{align}
$
\epsilon_V(H) \DefinedAs \frac{\gamma^H}{1-\gamma},\, 
\epsilon_\nabla(H) \DefinedAs \frac{M_g\gamma^H\sqrt{1+H(1-\gamma)}}{(1-\gamma)^{3/2}}.
$
% \end{align}
\end{Lemma}
%
%
% \blue{SP: You will use $\epsilon$ later on. Make it its own equation so that you can reference it later. }
\vspace{-2mm}
The smoothness modulus $\Mv$ controls the second-order change of a margin along an update and appears in the confidence width and step size in Section~\ref{sec:method}, while the Lipschitz constant $\Lv$ bounds the first-order change of a constraint margin in the worst case. The truncation errors account for the fact that only trajectories of length $H$ are observed.
% , and decay geometrically in $H$.
\vspace{-2mm}

%%%%%%%%%%%%%%%%%%%%%%%%%%%%%%%%%%%%%%%%%%%%%%%%%%%%%%%%%%%%%%%%%%%%%%%%%%%%%%%%%%%%%%%%%%%%%%%%%%%%%%%%%%%%%%%%%%%%%%%%%%%%%%%%%%%%%%%%%%%%%%%%%%%%%%%%%%%%%%%%%%%%%%%%%%%%%%%%%%%%%%%%%%%%%%%%%%%%%%%%%%%%%%%%%%%%%%%%%%%%%%%%%%%%%%%%%%%%%%%%%%%%%%%%%%%%%%%%%%%%%%%%%%%%%%%%%%%%%%%%%%%%%%%%%%%%%%%%%%%%%%%%%%%%%%%%%%%%%%%%%%%%%%%%%%%%%%%%
%%%%%%%%%%%%%%%%%%%%%%%%%%%%%%%%%%%%%%%%%%%%%%%%%%%%%%%%%%%%%%%%%%%%%%%%%%%%%%%%%%%%

% \section{Zeroth-Order Log-Barrier Policy Gradient Approach}
\section{Algorithm Development}
\label{sec:method}
\vspace{-2mm}
We address the SafeX-RLHF problem \eqref{eq:primal} by maximizing the following log-barrier surrogate over the strictly feasible region,
% \citep{usmanova2024log,ni2025safe}
%
\vspace{-2mm}
\begin{equation}
\label{eq:barrier}
\max_{\bt\,\in\,\Theta}\ \Bbar^{\bt}(\rho) 
\;\DefinedAs\; 
V_r^{\bt}(\rho)+\eta\sum_{i\,=\,1}^m\log\bar V_{g_i}^{\bt}(\rho), \tag{SP}
\setlength{\belowdisplayskip}{2pt}
\end{equation}
where $\eta>0$ is the barrier parameter. A natural way to solve \eqref{eq:barrier} is gradient ascent on $\bt$ \citep{usmanova2024log,ni2025safe}, i.e., $\bt^{(t+1)}=\bt^{(t)}+\psi_t\,\nabla_{\bt}\Bbar^{\bt^{(t)}}(\rho)$ with step size $\psi_t$ and gradient
\vspace{-1mm}
\begin{align}\label{eq:barrier-grad}
    \nabla_{\bt} \Bbar^{\bt^{(t)}}(\rho)
    \; = \; 
    \nabla_{\bt} V_r^{\bt^{(t)}}(\rho)
    +
    \eta
    \sum_{i\,=\,1}^m
    \frac{\nabla_{\bt} \bar V_{g_i}^{\bt^{(t)}}(\rho)}{\bar V_{g_i}^{\bt^{(t)}}(\rho)}.
    \setlength{\belowdisplayskip}{1pt}
\end{align}
As a constraint margin $\bar V_{g_i}^{\bt^{(t)}}(\rho)$ approaches zero, its term dominates \eqref{eq:barrier-grad} and pushes the update away from the corresponding constraint boundary. This repulsion is what makes log-barrier methods suitable for safe exploration, provided that the step size is small enough so that the repulsion is never overrun. In standard RL, the gradients in \eqref{eq:barrier-grad} are estimated through the policy gradient theorem \citep{sutton1999policy} and the constraint margins through discounted cumulative utilities, both of which require explicit reward and utility signals. In SafeX-RLHF, only binary human feedback is available, and the rest of this section presents how to build the estimates of the constraint margins and gradients and the step size from the human feedback alone, which results in Algorithm~\ref{alg_zolbpg}.
%
%
%
% \subsection{Estimating constraint margins and gradients from binary human feedback}
\vspace{-2mm}
\subsection{Estimations from human feedback}
\label{sec:estimators}
\vspace{-1mm}
\textbf{Inverting the link function.} 
In RLHF, an empirical frequency $\hat p$ of $M$ binary answers can be turned into a return difference by inverting the link function. Prior work \citep{zhang2024zeroth,liu2026policy} clips $\hat p$ to $\mathcal P_\sigma= [\underline{p}, \overline{p}]$ of Assumption~\ref{assum:link} and then applies $\sigma^{-1}$. We instead extend $\sigma^{-1}$ from $\mathcal P_\sigma$ to the whole interval $[0,1]$ by its tangents at the two endpoints, i.e.,
%
%
% \blue{SP: Different definition now. What are $p_{\min}$ and $p_{\max}$? \red{WC: Done} }
%
\vspace{-1mm}
\begin{equation}
\label{eq:ext-inv}
\sinv(p) 
\DefinedAs \!
\begin{cases}
\sigma^{-1}(\underline{p})+(\sigma^{-1})'(\underline{p})(p-\underline{p}), & p<\underline{p},\\
\sigma^{-1}(p), & p\in\mathcal P_\sigma,\\
\sigma^{-1}(\overline{p})+(\sigma^{-1})'(\overline{p})(p-\overline{p}), & p>\overline{p}.
\end{cases}
\end{equation}
Its properties are formalized as follows. 
% This extension is part of our contribution, and its properties are formalized as follows. 
% \blue{SP: What follows needs is not very clear, references a technical result in the future, and it is not clear that is needed before (11). I would either remove it or claim the Lemma as a proposition here. This new definition is part of the contribution, so I think it is fair to give the properties of this object here. Then you can explain why this is good (which by the way is part of the things that are not clear). \red{WC: Done} } \DD{This extension allows us to establish Lipschitz continuity and bound the estimation error, which are essential for constructing confidence bounds that ensure safe policy updates??}
%
%
\begin{Lemma}[Inverse-link estimation]
\label{lem:link}
Let Assumption~\ref{assum:link} hold, $x\in\bar{\mathcal I}$, $\hat p$ the mean of $M$ i.i.d.\ Bernoulli$(\sigma(x))$ bits and $\hat x \DefinedAs \sinv(\hat p)$. Then $|\hat x-x|\le L|\hat p-\sigma(x)|\le L$ almost surely,
% \begin{align}
% \label{eq:link-bias}
$
\big|\E[\hat x]-x\big|
% \; \le \;
% \frac{L_2\sigma(x)(1-\sigma(x))}{2M}
\le
\beta_M 
\DefinedAs
\frac{L_2}{8M}.
$
% \end{align}
%
\end{Lemma}
%
% \DD{define $\beta_M$ when we need it? \red{WC: I think we can give it a name here, and reference it whenever use it.}}
%

\vspace{-2mm}
Clipping is not differentiable at the endpoints of $\mathcal P_\sigma$, so its error can only be controlled through the deviation $|\hat p-\sigma(x)|$, which is of order $1/\sqrt{M}$. Our extension $\sinv$ is differentiable on $[0,1]$ with an $L_2$-Lipschitz derivative, so a second-order expansion applies, in which the first-order term vanishes because $\E[\hat p]=\sigma(x)$. The error of an inverted answer therefore splits into a bias of order $1/M$ and a bounded fluctuation that averages out over trajectories. With this inversion, we present two phases of each iteration in Algorithm~\ref{alg_zolbpg}.

\vspace{-1mm}
\textbf{Phase 1 (Constraint margins).} Draw $N_c$ seeds $\{\omega_n\}_{n=1}^{N_c}$, which give initial states $s_{n,0}\sim\rho$ and base trajectories $\tau_{n,0}=\tau(\bt^{(t)},\omega_n)$ of length $H$. For each $\tau_{n,0}$ and each constraint $i$, query $M_c$ evaluators for absolute harmlessness, denote the frequency of positive answers by $\hat p^{(t)}_{c_i,n}$, and estimate constraint margins by 
%
% \blue{SP: I would give this estimate a name that reminds us of the constraints. $\hat{v}$?} \red{You mean a name for constraint margin estimate? Isn't $\widehat h^{(t)}_{c_i} $ below the name? } \blue{SP: The name $h$ is not related to the quantities it estimates. Isn't it estimating V? or V bar? I would call it $\hat{V}$ instead of $h$.}
%

\vspace{-5mm}
\begin{equation}
\label{eq:hc}
\widehat h^{(t)}_{c_i} 
\;\DefinedAs\; 
\frac{1}{N_c}\sum_{n\,=\,1}^{N_c}\sinv_{c_i}\big(\hat p^{(t)}_{c_i,n}\big), ~ i\in[m].
\setlength{\belowdisplayskip}{1pt}
\end{equation}
By \eqref{eq:channels} and \eqref{eq:ext-inv}, $\sinv_{c_i}(\hat p^{(t)}_{c_i,n})$ estimates $R^H_{g_i}(\tau_{n,0})-b_i$, whose mean is the truncated margin $V^{\bt^{(t)}, H}_{g_i}(\rho)-b_i$.

\vspace{-1mm}
\textbf{Phase 2 (Gradients).} Draw $N\le N_c$ i.i.d.\ directions $\{ \bv^{(t)}_n \}_{n=1}^N \sim\mathrm{Unif}(\mathbb V^{d-1})$ on the unit sphere. Given a perturbation magnitude $\mu$, run the perturbed policy $\pi_{\bt^{(t)}+\mu\bv^{(t)}_n}$ on the seed $\omega_n$ of the $n$-th base trajectory to obtain $\tau_{n,1}=\tau(\bt^{(t)}+\mu\bv^{(t)}_n,\omega_n)$, and query $M$ evaluators for pairwise helpfulness and pairwise harmlessness on $(\tau_{n,1},\tau_{n,0})$. With the resulting frequencies $\hat p^{(t)}_{\diamond,n}$, the zeroth-order smoothing \citep{zhang2024zeroth} gives the gradient estimates for $\diamond\in\{r\}\cup\{g_i\}_{i\,=\,1}^m,$
\vspace{-3mm}
\begin{equation}
\label{eq:zo-grad}
\hb^{(t)}_\diamond 
\;\DefinedAs\; 
\frac{d}{\mu}\cdot\frac1N\sum_{n\,=\,1}^N\sinv_\diamond\big(\hat p^{(t)}_{\diamond,n}\big)\bv^{(t)}_n.
\setlength{\belowdisplayskip}{2pt}
\end{equation}
By \eqref{eq:channels} and \eqref{eq:ext-inv}, $\sinv_\diamond(\hat p^{(t)}_{\diamond,n})$ estimates the return difference $R^H_\diamond(\tau_{n,1})-R^H_\diamond(\tau_{n,0})$, whose conditional mean given $\bv^{(t)}_n$ is $V^{\bt^{(t)}+\mu\bv_n^{(t)},H}_\diamond(\rho)-V^{\bt^{(t)},H}_\diamond(\rho)$.

Two designs in \eqref{eq:zo-grad} are essential for our analysis. $(i)$ The two trajectories of a pair are coupled. They differ only if the two policies sample different actions at some step, which happens with probability at most $H|\Ac|M_g\mu$, so the second moment of the return difference is at most of order $\mu$, whereas for independent rollouts it is of order $\Gbar^2$ even when the two policies coincide. This gain matters since the estimator amplifies the variance by $d^2/\mu^2$. $(ii)$ Each pair uses its own independently drawn direction $\bv^{(t)}_n$, so \eqref{eq:zo-grad} is an average of $N$ i.i.d.\ bounded random vectors and concentrates around its mean at rate $1/\sqrt N$ (Lemma~\ref{lem:est}). With a single direction shared by all $N$ pairs, as in \citet{zhang2024zeroth,liu2026policy}, the estimate does not concentrate around the true gradient, and thus cannot certify that an update is safe.
%
%
%
% \subsection{A step that cannot cross the boundary}

\vspace{-3mm}
\subsection{Boundary-aware updates}
\label{sec:stepsize}
\vspace{-2mm}
The estimates in Section~\ref{sec:estimators} carry bias and noise from three sources: finite-horizon truncation ($H$), zeroth-order smoothing ($\mu$), and human feedback with finite evaluators and trajectories ($M,M_c,N,N_c$). 
To execute a step that is safe despite these errors, we first fix a confidence level $\delta\in(0,1)$ and let
% \begin{align}
% \label{eq:iota}
$
\iot 
\DefinedAs
\log\frac{2(2m+1)}{\delta}+\frac14.
$
% \end{align}
%
Each iteration forms $2m+1$ estimates, one reward gradient, $m$ constraint gradients and $m$ constraint margins. We require each of them to be accurate with probability at least $1-\frac{\delta}{2m+1}$, so that by a union bound all $2m+1$ estimates are accurate simultaneously with probability at least $1-\delta$.
Define the variance proxy of the coupled zeroth-order estimator
\vspace{-1mm}
\begin{equation}
\label{eq:sigma-g}
\sigma_g^2 
\;\DefinedAs\; 
\frac{2d^2}{\mu^2}\Big(\Gbar^2H|\Ac|M_g\,\mu+\frac{L^2}{4M}\Big),
\setlength{\belowdisplayskip}{2pt}
\end{equation}
whose two terms are the trajectory noise (of order $\mu$) and the human feedback noise, both amplified by the factor $d^2/\mu^2$.
We show in Section~\ref{sec:estimation_error} that, with probability at least $1-\delta$, the constraint margin estimates \eqref{eq:hc} and the gradient estimates \eqref{eq:zo-grad} lie within the following confidence widths of their targets.
\vspace{-2mm}
\begin{align}
\wc& 
\;\DefinedAs\; 
\epsilon_V(H)+\beta_{M_c}+(\Gbar+L)\sqrt{\frac{2\iot}{N_c}},\label{eq:wc}\\
\wg& 
\;\DefinedAs\; 
\epsilon_\nabla(H)+\frac{d\Mv\mu}{2}+\frac{d\beta_M}{\mu}+\sqrt{\frac{8\sigma_g^2\iot}{N}}.\label{eq:wg}
\end{align}
\textbf{Lower confidence bounds on constraint margins.} 
The constraint margins appear in the denominators of \eqref{eq:barrier-grad}, and thus underestimating them strengthens the repulsion from the boundary. 
We then replace each constraint margin in \eqref{eq:barrier-grad} by its lower confidence bound $\underline\alpha_i(t) \DefinedAs \widehat h^{(t)}_{c_i}-\wc$, which satisfies $\underline\alpha_i(t)\le\bar V^{\bt^{(t)}}_{g_i}(\rho)$ whenever the constraint margin estimates are $\wc$-accurate, and replace the gradients in \eqref{eq:barrier-grad} by their estimates \eqref{eq:zo-grad}. This gives the estimate of the barrier gradient \eqref{eq:barrier-grad} and the unit vector in its direction
\vspace{-1mm}
\begin{equation}
\label{eq:barrier-grad-hat-new}
\hg_t 
\; \DefinedAs \; \hb^{(t)}_r
+
\eta\sum_{i\,=\,1}^m
\frac{\hb^{(t)}_{g_i}}{\underline\alpha_i(t)}, ~
\hat u_t \DefinedAs \hg_t/\|\hg_t\|.
\end{equation}
%
% The update of Algorithm~\ref{alg_zolbpg} moves along $\hat u_t$, and the step size rule below uses $\hat u_t$ to measure how fast each constraint margin can change along the update.
%
%
%
\textbf{Upper confidence bounds on the directional derivatives.} 
Let $\beta_i(t) \DefinedAs \langle\nabla_{\bt}V^{\bt^{(t)}}_{g_i}(\rho),\hat u_t\rangle$ denote the directional derivative of the $i$-th constraint value along $\hat u_t$. It measures how fast the $i$-th constraint margin changes along the update, so the step must be limited by an upper bound on $|\beta_i(t)|$. Since $\beta_i(t)$ depends on the unknown true gradient, we bound it through the gradient estimate. The projection $\langle\hb^{(t)}_{g_i},\hat u_t\rangle$ is within $\wg$ of $\beta_i(t)$ whenever the gradient estimate is $\wg$-accurate. We therefore define
\vspace{-2mm}
\begin{align}
\label{eq:beta-ub}
\overline\beta_i(t) 
\;\DefinedAs\; 
\big|\langle\hb^{(t)}_{g_i},\hat u_t\rangle\big|+\wg, ~ i\in[m].
\end{align}
\textbf{Adaptive step size.} Following \citet{ni2025safe}, we define the local smoothness constant $\widehat M_t$ of \eqref{eq:barrier} along $\hat u_t$ and the step size $\psi_t$ as
\vspace{-2mm}
\begin{align}
\widehat M_t& 
\DefinedAs  \Mv+\sum_{i\,=\,1}^m\left(\frac{10\Mv\eta}{\underline\alpha_i(t)}+\frac{8\eta\overline\beta_i(t)^2}{\underline\alpha_i(t)^2}\right),\label{eq:Mhat}\\
\psi_t& 
\DefinedAs \!
\min \! \left\{ \! \frac{1}{\widehat M_t},\! \min_{i\,\in\,[m]} \! \frac{\underline\alpha_i(t)}{\sqrt{\Mv\underline\alpha_i(t)}\!+\!2\overline\beta_i(t)} \!\cdot\! \frac{1}{\|\hg_t\|} \! \right\} \!\label{eq:stepsize}
\end{align}
The two terms in \eqref{eq:stepsize} serve two purposes. The first term $1/\widehat M_t$ is small enough that the barrier objective increases along $\hat u_t$, since $\widehat M_t$ bounds the curvature of \eqref{eq:barrier} along the update. The second term ensures that no constraint margin loses more than half of its value in one update, so that a single step can never cross a constraint boundary. Both terms are computed from the confidence bounds, and no Hessian information is needed. Following \citet{ni2025safe}, Algorithm~\ref{alg_zolbpg} stops when $\|\hg_t\|\le\eta/2$ or after $T$ iterations.
\vspace{-5mm}
\begin{algorithm}[H]
% \caption{\underline{Z}eroth-\underline{O}rder \underline{L}og-\underline{B}arrier \underline{P}olicy \underline{G}radient (ZO-LBPG)}
\caption{ZO-LBPG}
\label{alg_zolbpg}
\begin{algorithmic}[1]
\Require feasible $\bt^{(0)}$, parameters $\eta,\mu,H,\delta,T$, batch sizes $N_c\ge N$, evaluator counts $M_c,M$, constants $\Mv,L,L_2,M_g$, the widths $\wc,\wg$ in \eqref{eq:wc} and \eqref{eq:wg}.
\For{$t=0,\dots,T-1$}
  \State \textbf{Phase 1.} Draw $N_c$ seeds $\{\omega_n\}_{n=1}^{N_c}$ and generate base trajectories $\tau_{n,0}=\tau(\bt^{(t)}, \omega_n)$. Query $M_c$ evaluators for absolute harmlessness of each $\tau_{n,0}$ w.r.t.\ each constraint. Compute $\widehat h^{(t)}_{c_i}$ by \eqref{eq:hc} and $\underline\alpha_i(t)\gets\widehat h^{(t)}_{c_i}-\wc$.
  \State \textbf{Phase 2.} Draw $\{\bv^{(t)}_n\}_{n=1}^N\sim\mathrm{Unif}(\mathbb V^{d-1})$ and generate \textit{coupled} trajectories $\tau_{n,1}=\tau(\bt^{(t)}+\mu\bv^{(t)}_n, \omega_n)$. Query $M$ evaluators for pairwise helpfulness and harmlessness on each $(\tau_{n,1},\tau_{n,0})$. Compute $\hb^{(t)}_\diamond$ by \eqref{eq:zo-grad}, $\hg_t$, $\hat u_t$ by \eqref{eq:barrier-grad-hat-new}, and $\overline\beta_i(t)$ by \eqref{eq:beta-ub}.
  \If{$\|\hg_t\|\le\eta/2$} \Return $\bt^{(t)}$ \EndIf
  \State Compute $\widehat M_t$ and $\psi_t$ by \eqref{eq:Mhat} and \eqref{eq:stepsize}.
  \State $\bt^{(t+1)}\gets\bt^{(t)}+\psi_t\,\hg_t$.
\EndFor
\State \Return $\bt^{(T)}$
\end{algorithmic}
\end{algorithm}

%%%%%%%%%%%%%%%%%%%%%%%%%%%%%%%%%%%%%%%%%%%%%%%%%%%%%%%%%%%%%%
%%%%%%%%%%%%%%%%%%%%%%%%%%%%%%%%%%%%%%%%%%%%%%%%%%%%%%%%%%%%%%
%%%%%%%%%%%%%%%%%%%%%%%%%%%%%%%%%%%%%%%%%%%%%%%%%%%%%%%%%%%%%%
%%%%%%%%%%%%%%%%%%%%%%%%%%%%%%%%%%%%%%%%%%%%%%%%%%%%%%%%%%%%%%
%%%%%%%%%%%%%%%%%%%%%%%%%%%%%%%%%%%%%%%%%%%%%%%%%%%%%%%%%%%%%%

\vspace{-5mm}
\section{Theoretical Guarantees}
\label{sec:theory}
\vspace{-3mm}
We state the remaining assumptions in Section~\ref{sec:assumptions}, quantify the estimation errors in Section~\ref{sec:estimation_error}, and establish safe exploration in Section~\ref{sec:safe_exploration_theory} and global convergence with its sample complexity in Section~\ref{sec:convergence}. All proofs are provided in the appendices.
\vspace{-3mm}
\subsection{Assumptions}
\label{sec:assumptions}
\vspace{-2mm}
The first assumption is the extended Mangasarian--Fromovitz constraint qualification (MFCQ) of \citet{ni2025safe}.
\begin{assumption}[Extended MFCQ]
\label{assum:mfcq}
There exist $0<\nu_{\mathrm{emf}}\le\nu$ and $\ell\in(0,\Lv]$ such that for every $\bt\in\Theta$ there is a unit vector $s_{\bt}$ with $\langle s_{\bt},\nabla_{\bt}V^{\bt}_{g_i}(\rho)\rangle\ge\ell$ for all $i\in\Iset_{\nu_{\mathrm{emf}}}(\bt) \DefinedAs \{i\,\vert\,0<\bar V^{\bt}_{g_i}(\rho)\le\nu_{\mathrm{emf}}\}$.
\end{assumption}
Assumption~\ref{assum:mfcq} is what allows the barrier to repel the iterates from the boundary, since the barrier gradient has a large component along $s_{\bt}$ whenever some constraint margins are small. The strength of this repulsion is quantified by the constants
\begin{align}
\label{eq:constants-c}
\bar c \;\DefinedAs\; \frac{\ell}{6m\Lv}
\;\;\text{ and }\;\;
c \;\DefinedAs\; \frac32\Big(\frac{\ell}{18m\Lv}\Big)^m ,
\end{align}
and we show in Section~\ref{sec:safe_exploration_theory} that no constraint margin ever drops below $c\eta$. The exponential dependence of $c$ on $m$ is shared with \citet{usmanova2024log,ni2025safe}.

The next two assumptions are the standard conditions under which the barrier objective satisfies a gradient dominance property, which we use for convergence to a globally optimal policy \citep{yuan2022general,masiha2022stochastic,ni2025safe}. Let $d^{\bt}_\rho(s,a) \DefinedAs (1-\gamma)\sum_{t\,\ge\,0}\gamma^t\PP^{\pi_{\bt}}(s_t=s,a_t=a)$ be the discounted state-action visitation distribution of $\pi_{\bt}$ and $F^{\bt}(\rho) \DefinedAs \E_{d^{\bt}_\rho}[\nabla_{\bt}\log\pi_{\bt}(a|s)\nabla_{\bt}\log\pi_{\bt}(a|s)^\top]$ the Fisher information matrix.
\begin{assumption}[Relaxed Fisher non-degeneracy]
\label{assum:fisher}
The smallest non-zero eigenvalue of $F^{\bt}(\rho)$ is at least $\mu_F>0$ for every $\bt\in\Theta$.
\end{assumption}
\begin{assumption}[Bounded transfer error]
\label{assum:transfer}
For $\diamond\in\{r\}\cup\{g_i\}_{i=1}^m$, let $A^{\bt}_\diamond$ denote the advantage function and $\mu^*_\diamond \DefinedAs (F^{\bt}(\rho))^\dagger\nabla_{\bt}V^{\bt}_\diamond(\rho)$. For every $\bt\in\Theta$ and every optimal policy $\pi^*$ of \eqref{eq:primal}, $\E_{(s,a)\sim d^{\pi^*}_\rho}[(A^{\bt}_\diamond(s,a)-(1-\gamma){\mu^*_\diamond}^\top\nabla_{\bt}\log\pi_{\bt}(a|s))^2]\le\varepsilon_{\mathrm{bias}}$.
\end{assumption}

\vspace{-3mm}
\subsection{Estimation guarantees}
\label{sec:estimation_error}
\vspace{-2mm}
Let $\mathcal F_t$ denote the $\sigma$-algebra generated by all seeds, directions and evaluator answers of iterations $\{0,\dots,t-1\}$. The following lemma demonstrates that the confidence widths of Section~\ref{sec:stepsize} are valid.
\begin{Lemma}[Accuracy of the estimators]
\label{lem:est}
Let Assumptions~\ref{assum:link},~\ref{assum:seeds} and~\ref{assum:smooth} hold, and let
\begin{equation}
\label{eq:range}
N \;\ge\; \frac{32d^2(\Gbar+L)^2\iot}{\mu^2\sigma_g^2}.
\setlength{\belowdisplayskip}{2pt}
\end{equation}
Then, conditionally on $\mathcal F_t$, the event
\begin{align}
\label{eq:good-event}
\mathcal E_t \DefinedAs \Big\{&\|\hb^{(t)}_\diamond-\nabla_{\bt}V^{\bt^{(t)}}_\diamond(\rho)\|\le\wg, \forall \, \diamond\in\{r\}\cup\{g_i\}_{i=1}^m, \notag\\
&|\widehat h^{(t)}_{c_i}-\bar V^{\bt^{(t)}}_{g_i}(\rho)|\le\wc,  \forall i\in[m]\Big\}
\end{align}
has probability at least $1-\delta$.
\end{Lemma}
\vspace{-2mm}
Condition \eqref{eq:range} follows from the Bernstein inequality and is of lower order than the sample sizes needed for accuracy. On $\mathcal E_t$ the upper confidence bounds \eqref{eq:beta-ub} satisfy $|\beta_i(t)|\le\overline\beta_i(t)\le|\beta_i(t)|+2\wg$ for all $i\in[m]$.

The barrier gradient estimate $\hg_t$ combines the individual estimates through \eqref{eq:barrier-grad-hat-new}, and the next result bounds the resulting error. Let $\alpha_i(t) \DefinedAs \bar V^{\bt^{(t)}}_{g_i}(\rho)$, $\alpha_{\min}(t) \DefinedAs \min_i\alpha_i(t)$ and $\Delta_t \DefinedAs \hg_t-\nabla_{\bt}\Bbar^{\bt^{(t)}}(\rho)$.
\begin{theorem}[Gradient error]
\label{theorem:barrier-err}
Under Assumptions~\ref{assum:link},~\ref{assum:seeds},~\ref{assum:smooth} and~\ref{assum:mfcq}, on $\mathcal E_t$, if $\wc\le\alpha_{\min}(t)/8$, then $\frac34\alpha_i(t)\le\underline\alpha_i(t)\le\alpha_i(t)$ for all $i\in[m]$ and
\vspace{-2mm}
\begin{equation}
\label{eq:delta-bound}
\|\Delta_t\|
\; \le \;
\wg\Big(1+\sum_{i=1}^m \frac{4\eta}{3\alpha_i(t)}\Big)+\sum_{i=1}^m \frac{8\eta\Lv\wc}{3\alpha_i(t)^2}.
\setlength{\belowdisplayskip}{2pt}
\end{equation}
If moreover $\alpha_{\min}(t)\ge c\eta$ and
\vspace{-2mm}
\begin{align}
\label{eq:width-req}
\wg&
\;\le\;
\bar w_g \DefinedAs C_1\min\{c\eta,2\ell\}, \notag\\
\wc&
\;\le\;
\bar w_c \DefinedAs \eta\, C_2\,\min\{c\eta,2\ell\},
\end{align}
where $C_1,C_2>0$ depend only on $m$, $c$ and $\Lv$ (Appendix~\ref{app:estimation}), then $\|\Delta_t\|\le\eta/4$.
\end{theorem}
The two terms of \eqref{eq:delta-bound} come from the numerators and the denominators of the barrier gradient and expose an asymmetry that drives the sample complexity. A gradient error is multiplied by at most $\eta/\alpha_i(t)\le1/c$, so the gradients only need accuracy $\mathcal O(\eta)$, whereas a constraint margin error is multiplied by $\eta/\alpha_i(t)^2$, which is of order $1/\eta$, so the constraint margins need accuracy $\mathcal O(\eta^2)$. The more demanding requirement consequently falls on the constraint margin channel, which needs no perturbation and no factor $d/\mu$.

\vspace{-3mm}
\subsection{Safe exploration}
\label{sec:safe_exploration_theory}
\vspace{-2mm}
On the good events $\mathcal E_t$, the step size \eqref{eq:stepsize} keeps every executed policy feasible throughout the run. The proof (Appendix~\ref{app:safe}) combines three facts. One update never reduces a constraint margin by more than half of its lower confidence bound, the barrier objective is $\widehat M_t$-smooth along the update so that the step size $1/\widehat M_t$ increases it, and whenever the smallest constraint margin drops below $\bar c\eta$, the repulsion of Assumption~\ref{assum:mfcq} dominates the barrier gradient, so that the \emph{product} of the nearly active constraint margins does not decrease.
\begin{theorem}[Safe exploration]
\label{thm:safe}
Under Assumptions~\ref{assum:link},~\ref{assum:seeds},~\ref{assum:slater},~\ref{assum:smooth},~\ref{assum:mfcq} and that $\eta\le\min\{1,\nu_{\mathrm{emf}}\}$. Choose the hyperparameters so that \eqref{eq:range} holds, $\wg\le\bar w_g$, $\wc\le\bar w_c$ and $\mu\le c\eta/(2\Lv)$. Then with probability at least $1-T\delta$, the following holds for every iteration $t$ reached by Algorithm~\ref{alg_zolbpg}. $(i)$ $\alpha_i(t)\ge c\eta$ for all $i\in[m]$. $(ii)$ Every exploration policy $\pi_{\bt^{(t)}+\mu\bv^{(t)}_n}$ has all constraint margins at least $c\eta/2$. $(iii)$ $\|\Delta_t\|\le\eta/4$. $(iv)$ $\psi_t\ge C\eta$ for a constant $C>0$ that depends only on $m,c,\Mv,\Lv,\Gbar$ (Appendix~\ref{app:safe}). Therefore, ZO-LBPG ensures safe exploration with probability at least $1-T\delta$.
\end{theorem}
\vspace{-2mm}
Parts $(i)$ and $(ii)$ are the safety guarantee, and parts $(iii)$ and $(iv)$ are what the convergence analysis needs. Since the constraint margin floor $c\eta$ does not depend on $t$, the requirements \eqref{eq:width-req} are uniform over the run and the sample sizes can be fixed in advance.

\vspace{-2mm}
\subsection{Convergence and sample complexity}
\label{sec:convergence}
\vspace{-1mm}
Global convergence rests on the gradient dominance property, connecting the optimality gap of a strictly feasible policy to the norm of its barrier gradient.
\begin{Lemma}[Gradient dominance]
\label{lem:gd}
Under Assumptions~\ref{assum:smooth},~\ref{assum:fisher},~\ref{assum:transfer}, for every strictly feasible $\bt$ and optimal $\pi^*$, it holds that 
\vspace{-2mm}
\begin{align}
V_r^{\pi^*}(\rho)-V_r^{\bt}(\rho) 
&\le m\eta+\frac{\sqrt{\varepsilon_{\mathrm{bias}}}}{1-\gamma}\Big(1+\sum_{i=1}^m \frac{\eta}{\bar V^{\bt}_{g_i}(\rho)}\Big) \notag\\
&\quad +\frac{M_g}{\mu_F}\|\nabla_{\bt}\Bbar^{\bt}(\rho)\|.
\setlength{\belowdisplayskip}{2pt}
\end{align}
\end{Lemma}
\vspace{-2mm}
Theorem~\ref{thm:safe} keeps the barrier gradient estimate accurate and the step size bounded below, and together with the ascent property of the step this yields a linear contraction of the optimality gap.
\begin{theorem}[Global convergence]
\label{thm:main}
Let Assumptions~\ref{assum:link},~\ref{assum:seeds},~\ref{assum:slater},~\ref{assum:smooth},~\ref{assum:mfcq},~\ref{assum:fisher} and~\ref{assum:transfer} hold, $\eta\le\min\{1,\nu_{\mathrm{emf}}\}$, and let the hyperparameters satisfy the conditions of Theorem~\ref{thm:safe}. Set $\bar C \DefinedAs \min\{C,8M_g/(\mu_F\eta^2)\}$ and $\kappa \DefinedAs \frac{\bar C\mu_F\eta^2}{8M_g}\in(0,1]$. With probability at least $1-T\delta$, the output $\bt_{\mathrm{out}}$ satisfies
\vspace{-1mm}
\begin{align}
V_r^{\pi^*}(\rho)-V_r^{\bt_{\mathrm{out}}}(\rho) &\le (1-\kappa)^T\big(V_r^{\pi^*}(\rho)-V_r^{\bt^{(0)}}(\rho)\big) \notag \\ 
&\quad + \frac{\sqrt{\varepsilon_{\mathrm{bias}}}}{1-\gamma}\Big(1+\frac mc\Big)+\frac{3M_g\eta}{4\mu_F} \notag \\
&\quad +m\eta\Big(1+\log\frac{1}{c\eta(1-\gamma)}\Big).
\end{align}
\end{theorem}
\vspace{-3mm}
The four terms are the optimization error decaying linearly, the approximation error vanishing for the softmax class, the loss incurred at stopping triggers, and the bias of the barrier, which is $\mathcal O(\eta\log\frac1\eta)$. An $\varepsilon$-optimal policy hence requires $\eta\propto\varepsilon$, and converting \eqref{eq:range} and \eqref{eq:width-req} into sample sizes yields the following.
\begin{corollary}[Sample complexity]
\label{cor:samples}
Under the assumptions of Theorem~\ref{thm:main}, fix $\varepsilon\in(0,1]$ and $\delta_0\in(0,1)$, set $\eta=\varepsilon/m$ (assuming $\varepsilon/m\le\min\{\nu_{\mathrm{emf}},2\ell/c\}$) and $\delta=\delta_0/T$, and choose $T,\mu,M_c,M,N_c,N,H$ as in Appendix~\ref{app:gd}, with $T=\widetilde{\mathcal O}(\varepsilon^{-2})$, $N_c=\widetilde{\mathcal O}(\varepsilon^{-4})$, $N=\widetilde{\mathcal O}(d^3\varepsilon^{-3})$, $M_c=\mathcal O(\varepsilon^{-2})$, $M=\mathcal O(d^2\varepsilon^{-2})$ and $H=\mathcal O(\frac{1}{1-\gamma}\log\frac{d}{\varepsilon})$. Then with probability at least $1-\delta_0$, ZO-LBPG ensures safe exploration and
\vspace{-0.5mm}
\begin{align}
&V_r^{\pi^*}(\rho)-V_r^{\bt_{\mathrm{out}}}(\rho) \notag \\
&\le2\varepsilon+\frac{3M_g\varepsilon}{4m\mu_F}+\varepsilon\log\frac{m}{c\varepsilon(1-\gamma)}+\frac{2m\sqrt{\varepsilon_{\mathrm{bias}}}}{c(1-\gamma)},
\setlength{\belowdisplayskip}{2pt}
\end{align}
using $\widetilde{\mathcal O}(\varepsilon^{-6}+d^3\varepsilon^{-5})$ trajectories and $\widetilde{\mathcal O}(\varepsilon^{-8}+d^5\varepsilon^{-7})$ human queries in total.
\end{corollary}
\vspace{-2mm}
The margins need accuracy $\mathcal O(\varepsilon^2)$, which costs $N_c=\widetilde{\mathcal O}(\varepsilon^{-4})$ base trajectories per iteration, but they are estimated from absolute judgments on unperturbed trajectories, so $N_c$ and $M_c$ do not depend on $d$. The gradients need only $\mathcal O(\varepsilon)$ accuracy. Thus, the zeroth-order estimator, the only place where the dimension enters, costs $N=\widetilde{\mathcal O}(d^3\varepsilon^{-3})$ pairs per iteration. The leading terms $\varepsilon^{-6}$ and $\varepsilon^{-8}$ are thus dimension-free.
\begin{remark}[The price of human feedback]
\label{rem:complexity}
With explicit reward and utility signals, \citet{ni2025safe} need $\widetilde{\mathcal O}(\varepsilon^{-6})$ trajectories, and Corollary~\ref{cor:samples} attains the same leading rate from human feedback. The additional price is the factor $M_c=\mathcal O(\varepsilon^{-2})$ of evaluators per trajectory, which turns $\widetilde{\mathcal O}(\varepsilon^{-6})$ trajectories into $\widetilde{\mathcal O}(\varepsilon^{-8})$ human queries.
\end{remark}

%%%%%%%%%%%%%%%%%%%%%%%%%%%%%%%%%%%%%%%%%%%%%%%%%%%%%%%%%%%%%%%%%%%%%%%%%%%%%%%%%%%%%%%%%%%%%%%%%%%%%%%%%%%%%%%%%%%%%%%%%%%%%%%%%%%%%%%%%%%%%%%%%%%%%%%%%%%%%%%%%%%%%%%%%%%%%%%%%%%%%%%%%%%%%%%%%%%%%%%%%%%%%%%%%%%%%%%%%%%%%%%%%%%%%%%%%%%%%%%%%%%%%%%%%%%%%%%%%%%%%%%%%%%%%%%%%%%%%%%%%%%%%%%%%%%%%%%%%%%%%%%%%%%%%%%%%%%%%%%%%%%%%%%%%%%%%%%%%%%%%%%%%%%%%%%%%%%%%%%%%%%%%%%%%%%%%%%%%%%%%%%%%%%%%%%%%%%%%%%%%%%%%%%%%%%%%%%%%%%%

\vspace{-3mm}
\section{Experiments}
\label{sec:experiment}
\vspace{-2mm}
%
%
%
% \weiqin{Xinmeng added new experiments, so need to change.}
% \red{
% We compare ZO-LBPG with ZPGPD-HF \citep{liu2026policy}, the zeroth-order primal--dual method with average feasibility guarantees, on a tabular RLHF CMDP with simulated evaluators \citep{liu2026policy}. Additional details and results are
% provided in Appendix~\ref{app:experiments}.}

We evaluate ZO-LBPG in two settings: a synthetic CMDP that allows control of the optimization and feedback noise, and a language-model alignment task that examines the persistence training-time safety in a realistic autoregressive policy. Our primary question is not whether ZO-LBPG achieves the highest final helpfulness, but whether it can improve helpfulness while maintaining safety throughout training. We thus compare ZO-LBPG with ZPGPD-HF~\citep{liu2026policy}, a zeroth-order primal-dual method that only provides average-feasibility guarantees. Additional details and results are
provided in Appendix~\ref{app:experiments}.

% We compare ZO-LBPG with ZPGPD-HF \citep{liu2026policy}, the zeroth-order primal--dual method with average feasibility guarantees, in a synthetic setting~\citep{liu2026policy} and a language model setting. Additional details and results are
% provided in Appendix~\ref{app:experiments}.

\vspace{-2mm}
\subsection{Synthetic data}
\label{sec:sync-data}
\vspace{-2mm}
\textbf{Setup.} 
We first consider the synthetic CMDP used to illustrate the SafeX-RLHF setting. The environment has \(|\St|=6\) states, \(|\Ac|=3\) actions, discount factor \(\gamma=0.85\), and \(m=3\) harmlessness constraints, giving a policy dimension of \(d=18\) under tabular softmax parameterization. Transition probabilities are sampled from a Dirichlet distribution with concentration parameter \(0.5\). The reward is sampled uniformly from \([0,1]\), while the safety utilities are constructed as $g_i=0.4(1-r)+0.6\xi_i$ with $\xi_i\sim\mathrm{Unif}[0,1]$,
% \vspace{-1mm}
% \[
% g_i=0.4(1-r)+0.6\xi_i,\qquad
% \xi_i\sim\mathrm{Unif}[0,1],
% \]
creating a conflict between helpfulness and harmlessness. We set \(b_i=3\) for all \(i\).
The constrained optimum has all three constraints active and reward value \(V_r^\star(\rho)=5.31\), whereas the unconstrained optimum violates all three constraints. 
Both algorithms are initialized at the uniform policy, which is strictly feasible with initial margins \((0.52,0.40,0.26)\). 
% This initialization plays the role of a safe SFT reference model.
% To isolate the effect of the optimization method, both algorithms use the same human-feedback simulation. 
Trajectories have horizon \(H=40\), and pairwise and absolute judgments are generated using Bradley--Terry evaluators. We consider a low-noise regime with \(N=1024\), \(N_c=2048\), and \(M=64\) evaluators per query, and a high-noise regime with \(N=256\), \(N_c=512\), and \(M=8\). We run \(T=2000\) iterations with ten random seeds. 

\begin{figure}[ht]
    \centering
    \begin{subfigure}{0.49\textwidth}
        \centering\includegraphics[width=1.0\linewidth]{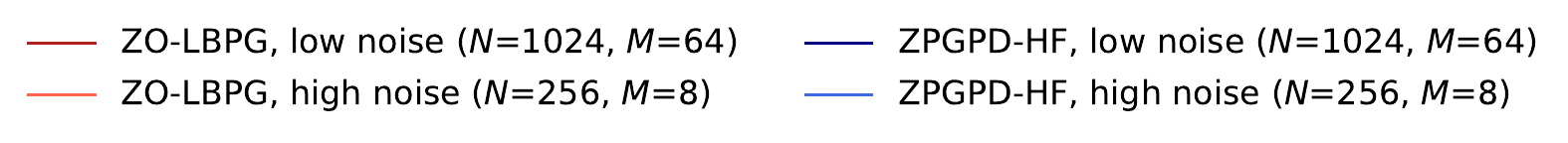}
    \end{subfigure}
    \begin{subfigure}{0.235\textwidth}
        \centering\includegraphics[width=1.0\linewidth]{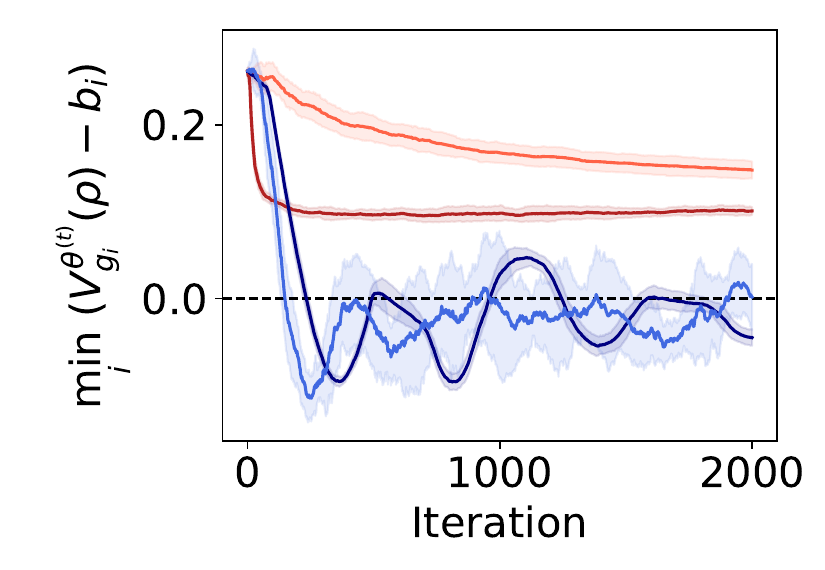}
        \caption{Min constraint margin.}
        \label{fig:margin}
    \end{subfigure}
    \begin{subfigure}{0.235\textwidth}
        \centering\includegraphics[width=1.0\linewidth]{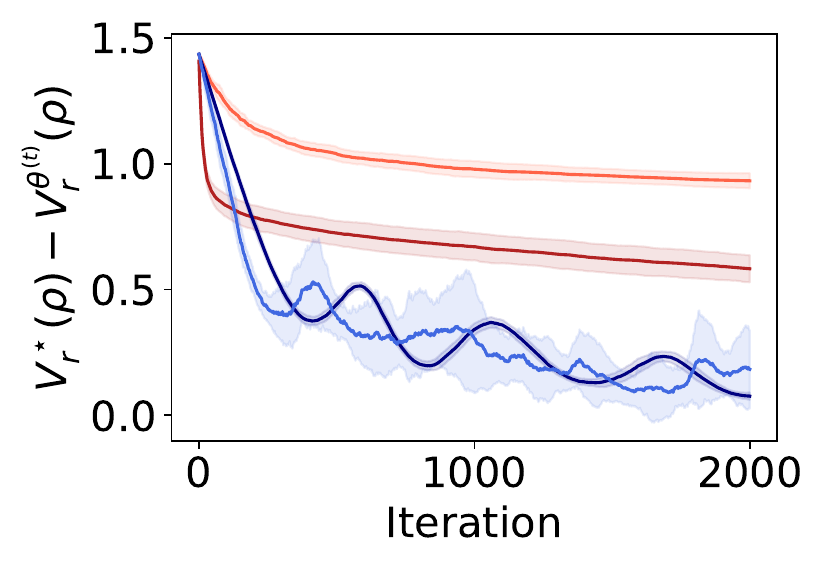}
        \caption{Reward optimality gap.}
        \label{fig:optgap}
    \end{subfigure}
    \vspace{-5mm}
\caption{Synthetic CMDP. ZO-LBPG keeps every iterate strictly feasible, whereas ZPGPD-HF constantly violates the constraints and attains its smaller reward gap through infeasible policies. Curves show means over ten seeds with one-standard-deviation bands.}
\label{fig:main}
\end{figure}

\vspace{-1mm}
\textbf{Results.} 
Figure~\ref{fig:main} highlights the central distinction between the two methods. ZPGPD-HF achieves a smaller reward optimality gap, but its iterates repeatedly cross the constraint boundary. In the low- and high-noise regimes, respectively, 66.6\% and 68.9\% of its iterates are infeasible, with worst constraint margins of \(-0.118\) and \(-0.179\). These oscillations are consistent with the primal-dual mechanism: the policy can move outside the feasible region while the dual variables react to the resulting constraint violation.
In contrast, ZO-LBPG maintains a positive margin throughout training in both regimes. No iterate violates any constraint, and the worst margin over all executed exploration policies remains positive as well: \(+0.052\) in the low-noise regime and \(+0.048\) in the high-noise regime. The price is a larger final reward gap, reflecting the conservative steps required to maintain a certified safety margin.
More ablations in Appendix~\ref{app:experiments} further clarify the source of this tradeoff. 
% Removing the coupling between trajectory pairs increases the gradient-estimation noise and slightly worsens reward optimization, but does not compromise the safety guarantee. Increasing the number of perturbation pairs from \(N=1024\) to \(N=2048\) improves the reward while leaving safety unchanged. 
% These observations are consistent with the theory: the gradient-estimation width \(w_g\) controls optimization accuracy, whereas the confidence bounds on the constraint margins determine the safety of the updates.

% ZPGPD-HF attains a small reward gap at the cost of leaving the feasible region on more than $66.6\%$ of its iterates, with the worst constraint margin $-0.18$, and the oscillations around the boundary that are typical of primal--dual methods (see Figure~\ref{fig:main} and Table~\ref{tab:results} in Appendix~\ref{app:experiments}).
% %
% % Its average margin over the second half of training is $-0.01$, so it satisfies the constraints on average, exactly as its theory predicts.
% %
% In particular, despite the larger optimality gap, ZO-LBPG never violates any constraint. The smallest constraint margin over $20000$ iterates per regime is $+0.081$, and over all executed exploration policies it is $+0.048$ (see Table~\ref{tab:results} in Appendix~\ref{app:experiments}). 
% %
% Appendix~\ref{app:experiments} also reports ablation studies on the coupling of the trajectory pairs and on the batch size $N$, which demonstrate that the reward performance of ZO-LBPG is governed by the width $\wg$ of the gradient estimate while its safety is unaffected.

\vspace{-2mm}

\vspace{-5mm}
\subsection{Experiments on language models}
\vspace{-2mm}
% \weiqin{I think this subsection needs to be reduced a bit. Something like Section 6.1 is good.}

\textbf{Setup.}
We next test ZO-LBPG on a language-model alignment task. We use Qwen2.5-3B-Instruct~\citep{qwen2025qwen25} with prompts from PKU-SafeRLHF~\citep{ji2024pku}, treating each prompt as a state and the generated tokens as the action. The base model is frozen, and we optimize a low-dimensional LoRA-style adapter with \(d=8\) trainable parameters. This low-dimensional parameterization allows us to study the zeroth-order method while keeping the computational cost manageable.
We use the Beaver-7B reward and cost models~\citep{dai2023safe} to provide latent helpfulness and harmlessness scores. These scores are used only to simulate noisy human judgments and to evaluate the learned policies; neither algorithm has access to the latent scores during optimization. Pairwise and absolute feedback are generated according to the Bradley--Terry model with \(M=M_c=256\) simulated evaluators. Both methods use the same prompts, feedback pipeline, and batch sizes, with \(N=512\), \(N_c=1024\), and \(T=300\). We  include an unconstrained helpfulness-only run as a reference.
% The initial adapter is strictly feasible and lies in a region where helpfulness and harmlessness conflict. 
% Thus, improving helpfulness without accounting for the safety constraint eventually leads to a violation. We implement the coupled sampler by sharing the randomness cut used for token generation between each base and perturbed policy, as required by the zeroth-order estimator.

% We apply ZO-LBPG to Qwen2.5-3B-Instruct~\citep{qwen2025qwen25} on prompts from PKU-SafeRLHF~\citep{ji2024pku}, with
% the prompt as the state and the response as the action. The seeded sampler of Assumption~\ref{assum:seeds} is
% implemented by inverse-CDF token sampling, so the two responses of a pair share the prompt and the random uniforms.
% The policy parameter is a LoRA-type adapter~\citep{hu2022lora,balazy2024loraxs} with frozen factors and a trainable
% core of dimension $d=8$. Helpfulness and harmlessness are latent and given by the Beaver-7B reward and cost
% models~\citep{dai2023safe}, with one constraint ($m=1$), and every query is answered by $M=M_c=256$ simulated
% Bradley--Terry evaluators. The initial adapter $\theta^{(0)}$ is strictly feasible (Assumption~\ref{assum:slater}) and
% lies where helpfulness and harmlessness conflict. Both methods use $N=512$, $N_c=1024$ and $T=300$ with three seeds,
% ZO-LBPG uses $\eta=0.05$, and unconstrained ascent on the helpfulness alone serves as a reference.
% Appendix~\ref{app:llm} gives the full setup and Table~\ref{tab:llm}.

\begin{figure}[ht]
\vspace{-2mm}
  \centering
  \begin{subfigure}{0.5\textwidth}
        \centering  \hspace{-5mm}\includegraphics[width=1\textwidth]{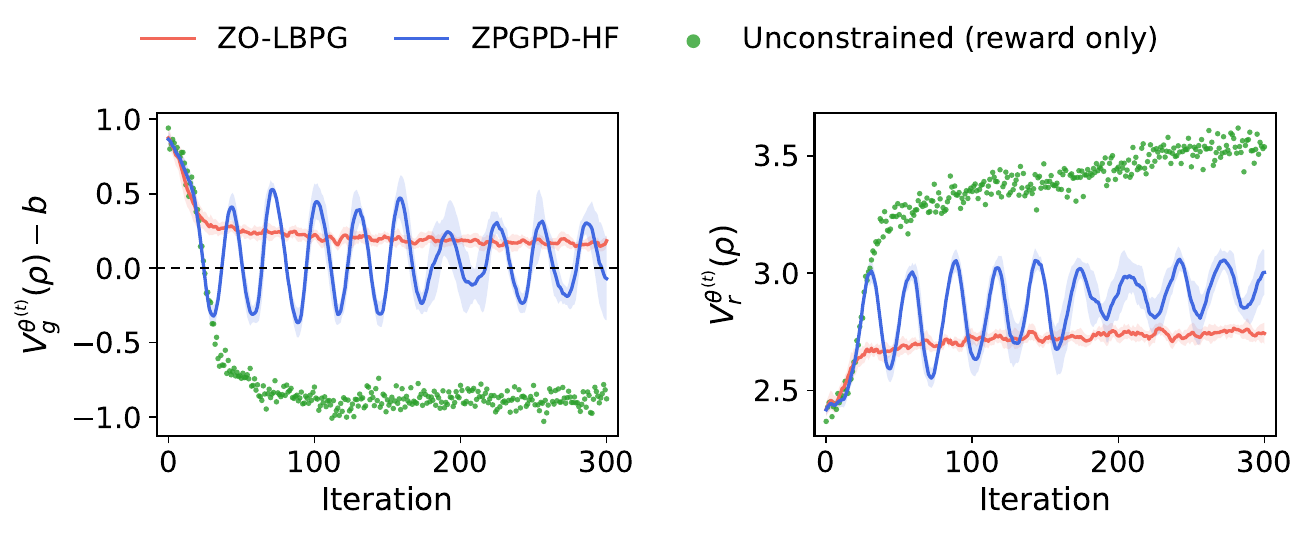}
        % \label{fig:margin}
    \end{subfigure}
    \vspace{-7mm}
  \caption{LLM experiment. ZO-LBPG keeps every iterate strictly feasible and improves the helpfulness, whereas
  ZPGPD-HF repeatedly crosses the constraint boundary. Means over three seeds 
  with one-standard-deviation bands, smoothed over five iterations.}
  \label{fig:llm}
\end{figure}

\textbf{Results.}
Figure~\ref{fig:llm} shows that the synthetic-CMDP phenomenon persists in the language-model setting. The unconstrained reference rapidly improves helpfulness but drives the safety margin to approximately \(-0.9\). ZPGPD-HF also oscillates around the constraint boundary and produces infeasible iterates: 43.5\% of its iterates have negative estimated margins, and its worst margin is \(-0.51\).
ZO-LBPG, in contrast, maintains a positive safety margin throughout the entire run. Across all three seeds, its smallest estimated iterate margin is \(+0.025\), and none of its exploration policies is infeasible. At the same time, ZO-LBPG improves helpfulness from 2.39 to 2.74. ZPGPD-HF reaches a higher final helpfulness of 2.94, but does so while repeatedly entering the infeasible region; even when restricted to its feasible iterates, its average helpfulness is 2.86.
These results illustrate the central tradeoff of training-time safety. A primal-dual method can move more aggressively toward the reward optimum but constantly violate constraint, whereas ZO-LBPG deliberately limits its updates using confidence bounds on the safety margin. 
% The resulting reward gap is thus the cost of obtaining a high-probability safety certificate for every policy executed during training.

% The unconstrained reference drives the margin to about $-0.9$, so the constraint is active
% (Figure~\ref{fig:llm}). ZPGPD-HF oscillates around the boundary and leaves the feasible region on $43.5\%$ of its
% iterates, with the worst margin $-0.51$. ZO-LBPG never violates the constraint: the smallest estimated margin over
% all $903$ iterates is $+0.025$, and no exploration policy is significantly infeasible. Meanwhile it raises the
% helpfulness from $2.39$ to $2.74$. As in Section~\ref{sec:sync-data}, ZPGPD-HF reaches a higher helpfulness
% ($2.94$), partly through infeasible iterates, and the remaining gap is the price of training-time safety: ZO-LBPG only
% takes steps it can certify and therefore stays about one confidence width away from the boundary.

% ====================================================================================
% Appendix block
% ====================================================================================

\vspace{-5mm}
\section{Conclusion}
\label{sec:conclusion}
\vspace{-2mm}
We developed ZO-LBPG, a reward-model-free method for safe exploration in RLHF. By combining a log-barrier objective with confidence-aware zeroth-order gradient estimates and adaptive step sizes, ZO-LBPG guarantees, with high probability, that both training iterates and auxiliary exploration policies remain strictly feasible throughout training. We further establish global convergence to an approximately optimal constrained policy with  dimension-free leading-order sample complexity. Experiments on a synthetic CMDP and a language-model alignment task demonstrate the central tradeoff: ZO-LBPG maintains training-time safety, while the primal-dual baseline achieves higher helpfulness by allowing intermediate constraint violations.

\clearpage

\bibliography{main}

@book{ben1985discrete,
  title={Discrete choice analysis: theory and application to travel demand},
  author={Ben-Akiva, Moshe E and Lerman, Steven R},
  volume={9},
  year={1985},
  publisher={MIT press}
}

@article{bradley1952rank,
  title={Rank analysis of incomplete block designs: I. The method of paired comparisons},
  author={Bradley, Ralph Allan and Terry, Milton E},
  journal={Biometrika},
  volume={39},
  number={3/4},
  pages={324--345},
  year={1952},
  publisher={JSTOR}
}

@article{zhang2024zeroth,
  title={Zeroth-order policy gradient for reinforcement learning from human feedback without reward inference},
  author={Zhang, Qining and Ying, Lei},
  journal={arXiv preprint arXiv:2409.17401},
  year={2024}
}

@article{liu2018zeroth,
  title={Zeroth-order stochastic variance reduction for nonconvex optimization},
  author={Liu, Sijia and Kailkhura, Bhavya and Chen, Pin-Yu and Ting, Paishun and Chang, Shiyu and Amini, Lisa},
  journal={Advances in neural information processing systems},
  volume={31},
  year={2018}
}

@article{agarwal2021theory,
  title={On the theory of policy gradient methods: Optimality, approximation, and distribution shift},
  author={Agarwal, Alekh and Kakade, Sham M and Lee, Jason D and Mahajan, Gaurav},
  journal={Journal of Machine Learning Research},
  volume={22},
  number={98},
  pages={1--76},
  year={2021}
}

@article{chatterji2021theory,
  title={On the theory of reinforcement learning with once-per-episode feedback},
  author={Chatterji, Niladri and Pacchiano, Aldo and Bartlett, Peter and Jordan, Michael},
  journal={Advances in Neural Information Processing Systems},
  volume={34},
  pages={3401--3412},
  year={2021}
}

@article{zhang2026alignment,
  title={Alignment of large language models with constrained learning},
  author={Zhang, Botong and Li, Shuo and Hounie, Ignacio and Bastani, Osbert and Ding, Dongsheng and Ribeiro, Alejandro},
  journal={Advances in Neural Information Processing Systems},
  volume={38},
  pages={30960--31011},
  year={2026}
}

@article{ding2020natural,
  title={Natural policy gradient primal-dual method for constrained markov decision processes},
  author={Ding, Dongsheng and Zhang, Kaiqing and Basar, Tamer and Jovanovic, Mihailo},
  journal={Advances in Neural Information Processing Systems},
  volume={33},
  pages={8378--8390},
  year={2020}
}

@inproceedings{kakade2002approximately,
  title={Approximately optimal approximate reinforcement learning},
  author={Kakade, Sham and Langford, John},
  booktitle={Proceedings of the nineteenth international conference on machine learning},
  pages={267--274},
  year={2002}
}

@article{dai2023safe,
  title={Safe rlhf: Safe reinforcement learning from human feedback},
  author={Dai, Josef and Pan, Xuehai and Sun, Ruiyang and Ji, Jiaming and Xu, Xinbo and Liu, Mickel and Wang, Yizhou and Yang, Yaodong},
  journal={arXiv preprint arXiv:2310.12773},
  year={2023}
}

@article{zhan2023provable,
  title={Provable reward-agnostic preference-based reinforcement learning},
  author={Zhan, Wenhao and Uehara, Masatoshi and Sun, Wen and Lee, Jason D},
  journal={arXiv preprint arXiv:2305.18505},
  year={2023}
}

@inproceedings{ding2022ACC,
  title={Convergence and optimality of policy gradient primal-dual method for constrained Markov decision processes},
  author={Ding, Dongsheng and Zhang, Kaiqing and Ba{\c{s}}ar, Tamer and Jovanovi{\'c}, Mihailo R},
  booktitle={2022 American Control Conference (ACC)},
  pages={2851--2856},
  year={2022},
  organization={IEEE}
}

@article{bai2022training,
  title={Training a helpful and harmless assistant with reinforcement learning from human feedback},
  author={Bai, Yuntao and Jones, Andy and Ndousse, Kamal and Askell, Amanda and Chen, Anna and DasSarma, Nova and Drain, Dawn and Fort, Stanislav and Ganguli, Deep and Henighan, Tom and others},
  journal={arXiv preprint arXiv:2204.05862},
  year={2022}
}

@article{ouyang2022training,
  title={Training language models to follow instructions with human feedback},
  author={Ouyang, Long and Wu, Jeffrey and Jiang, Xu and Almeida, Diogo and Wainwright, Carroll and Mishkin, Pamela and Zhang, Chong and Agarwal, Sandhini and Slama, Katarina and Ray, Alex and others},
  journal={Advances in neural information processing systems},
  volume={35},
  pages={27730--27744},
  year={2022}
}

@article{casper2023open,
  title={Open problems and fundamental limitations of reinforcement learning from human feedback},
  author={Casper, Stephen and Davies, Xander and Shi, Claudia and Gilbert, Thomas Krendl and Scheurer, J{\'e}r{\'e}my and Rando, Javier and Freedman, Rachel and Korbak, Tomasz and Lindner, David and Freire, Pedro and others},
  journal={arXiv preprint arXiv:2307.15217},
  year={2023}
}

@article{zhang2024reinforcement,
  title={Reinforcement learning from human feedback without reward inference: Model-free algorithm and instance-dependent analysis},
  author={Zhang, Qining and Wei, Honghao and Ying, Lei},
  journal={arXiv preprint arXiv:2406.07455},
  year={2024}
}

@article{du2024exploration,
  title={Exploration-driven policy optimization in rlhf: Theoretical insights on efficient data utilization},
  author={Du, Yihan and Winnicki, Anna and Dalal, Gal and Mannor, Shie and Srikant, R},
  journal={arXiv preprint arXiv:2402.10342},
  year={2024}
}

@article{paternain2019constrained,
  title={Constrained reinforcement learning has zero duality gap},
  author={Paternain, Santiago and Chamon, Luiz and Calvo-Fullana, Miguel and Ribeiro, Alejandro},
  journal={Advances in Neural Information Processing Systems},
  volume={32},
  year={2019}
}

@inproceedings{zhu2023principled,
  title={Principled reinforcement learning with human feedback from pairwise or k-wise comparisons},
  author={Zhu, Banghua and Jordan, Michael and Jiao, Jiantao},
  booktitle={International Conference on Machine Learning},
  pages={43037--43067},
  year={2023},
  organization={PMLR}
}

@article{wu2023making,
  title={Making rl with preference-based feedback efficient via randomization},
  author={Wu, Runzhe and Sun, Wen},
  journal={arXiv preprint arXiv:2310.14554},
  year={2023}
}

@inproceedings{achiam2017constrained,
  title={Constrained policy optimization},
  author={Achiam, Joshua and Held, David and Tamar, Aviv and Abbeel, Pieter},
  booktitle={International conference on machine learning},
  pages={22--31},
  year={2017},
  organization={PMLR}
}

@article{tessler2018reward,
  title={Reward constrained policy optimization},
  author={Tessler, Chen and Mankowitz, Daniel J and Mannor, Shie},
  journal={arXiv preprint arXiv:1805.11074},
  year={2018}
}

@article{hinder2019polynomial,
  title={A polynomial time log barrier method for problems with nonconvex constraints},
  author={Hinder, Oliver and Ye, Yinyu},
  journal={arXiv preprint arXiv:1807.00404},
  year={2018}
}

@inproceedings{bai2022achieving2,
  title={Achieving zero constraint violation for constrained reinforcement learning via conservative natural policy gradient primal-dual algorithm},
  author={Bai, Qinbo and Bedi, Amrit Singh and Aggarwal, Vaneet},
  booktitle={Proceedings of the AAAI Conference on Artificial Intelligence},
  volume={37},
  pages={6737--6744},
  year={2023}
}

@article{ding2024last,
  title={Last-iterate convergent policy gradient primal-dual methods for constrained mdps},
  author={Ding, Dongsheng and Wei, Chen-Yu and Zhang, Kaiqing and Ribeiro, Alejandro},
  journal={Advances in Neural Information Processing Systems},
  volume={36},
  year={2024}
}

@article{chow2018lyapunov,
  title={A lyapunov-based approach to safe reinforcement learning},
  author={Chow, Yinlam and Nachum, Ofir and Duenez-Guzman, Edgar and Ghavamzadeh, Mohammad},
  journal={Advances in neural information processing systems},
  volume={31},
  year={2018}
}

@article{chow2019lyapunov,
  title={Lyapunov-based safe policy optimization for continuous control},
  author={Chow, Yinlam and Nachum, Ofir and Faust, Aleksandra and Duenez-Guzman, Edgar and Ghavamzadeh, Mohammad},
  journal={arXiv preprint arXiv:1901.10031},
  year={2019}
}

@article{koller2019learning,
  title={Learning-based Model Predictive Control for Safe Exploration and Reinforcement Learning},
  author={Koller, Torsten and Berkenkamp, Felix and Turchetta, Matteo and Boedecker, Joschka and Krause, Andreas},
  journal={arXiv preprint arXiv:1906.12189},
  year={2019}
}

@inproceedings{koller2018learning,
  title={Learning-based model predictive control for safe exploration},
  author={Koller, Torsten and Berkenkamp, Felix and Turchetta, Matteo and Krause, Andreas},
  booktitle={2018 IEEE conference on decision and control (CDC)},
  pages={6059--6066},
  year={2018},
  organization={IEEE}
}

@article{berkenkamp2017safe,
  title={Safe model-based reinforcement learning with stability guarantees},
  author={Berkenkamp, Felix and Turchetta, Matteo and Schoellig, Angela and Krause, Andreas},
  journal={Advances in neural information processing systems},
  volume={30},
  year={2017}
}

@inproceedings{sui2015safe,
  title={Safe exploration for optimization with Gaussian processes},
  author={Sui, Yanan and Gotovos, Alkis and Burdick, Joel and Krause, Andreas},
  booktitle={International conference on machine learning},
  pages={997--1005},
  year={2015},
  organization={PMLR}
}

@inproceedings{wachi2018safe,
  title={Safe exploration and optimization of constrained mdps using gaussian processes},
  author={Wachi, Akifumi and Sui, Yanan and Yue, Yisong and Ono, Masahiro},
  booktitle={Proceedings of the AAAI Conference on Artificial Intelligence},
  volume={32},
  year={2018}
}

@inproceedings{yuan2022general,
  title={A general sample complexity analysis of vanilla policy gradient},
  author={Yuan, Rui and Gower, Robert M and Lazaric, Alessandro},
  booktitle={International Conference on Artificial Intelligence and Statistics},
  pages={3332--3380},
  year={2022},
  organization={PMLR}
}

@article{mondal2024sample,
  title={Sample-Efficient Constrained Reinforcement Learning with General Parameterization},
  author={Mondal, Washim Uddin and Aggarwal, Vaneet},
  journal={arXiv preprint arXiv:2405.10624},
  year={2024}
}

@inproceedings{zeng2022finite,
  title={Finite-time complexity of online primal-dual natural actor-critic algorithm for constrained Markov decision processes},
  author={Zeng, Sihan and Doan, Thinh T and Romberg, Justin},
  booktitle={2022 IEEE 61st Conference on Decision and Control (CDC)},
  pages={4028--4033},
  year={2022},
  organization={IEEE}
}

@inproceedings{xu2021crpo,
  title={{CRPO}: A new approach for safe reinforcement learning with convergence guarantee},
  author={Xu, Tengyu and Liang, Yingbin and Lan, Guanghui},
  booktitle={International Conference on Machine Learning},
  pages={11480--11491},
  year={2021},
  organization={PMLR}
}

@article{vaswani2022near,
  title={Near-optimal sample complexity bounds for constrained MDPs},
  author={Vaswani, Sharan and Yang, Lin and Szepesv{\'a}ri, Csaba},
  journal={Advances in Neural Information Processing Systems},
  volume={35},
  pages={3110--3122},
  year={2022}
}

@inproceedings{Xu2020Sample,
  title={Sample efficient policy gradient methods with recursive variance reduction},
  author={Xu, Pan and Gao, Felicia and Gu, Quanquan},
  booktitle={International Conference on Learning Representations},
  year={2020}
}

@article{masiha2022stochastic,
  title={Stochastic Second-Order Methods Improve Best-Known Sample Complexity of SGD for Gradient-Dominated Functions},
  author={Masiha, Saeed and Salehkaleybar, Saber and He, Niao and Kiyavash, Negar and Thiran, Patrick},
  journal={Advances in Neural Information Processing Systems},
  volume={35},
  pages={10862--10875},
  year={2022}
}

@book{altman1999constrained,
  title={Constrained Markov decision processes: stochastic modeling},
  author={Altman, Eitan},
  year={1999},
  publisher={Routledge}
}

@article{kalagarla2023safe,
  title={Safe Posterior Sampling for Constrained MDPs with Bounded Constraint Violation},
  author={Kalagarla, Krishna C and Jain, Rahul and Nuzzo, Pierluigi},
  journal={arXiv preprint arXiv:2301.11547},
  year={2023}
}

@article{usmanova2024log,
  title={Log Barriers for Safe Black-box Optimization with Application to Safe Reinforcement Learning},
  author={Usmanova, Ilnura and As, Yarden and Kamgarpour, Maryam and Krause, Andreas},
  journal={Journal of Machine Learning Research},
  volume={25},
  number={171},
  pages={1--54},
  year={2024}
}

@inproceedings{ni2025safe,
  title={A safe exploration approach to constrained Markov decision processes},
  author={Ni, Tingting and Kamgarpour, Maryam},
  booktitle={International Conference on Artificial Intelligence and Statistics},
  pages={3592--3600},
  year={2025},
  organization={PMLR}
}

@article{liu2026policy,
  title={Policy Gradient Primal-Dual Method for Safe Reinforcement Learning from Human Feedback},
  author={Liu, Qiang and Kline, Adrienne and Wei, Ermin},
  journal={arXiv preprint arXiv:2604.19024},
  year={2026}
}

@article{christiano2017deep,
  title={Deep reinforcement learning from human preferences},
  author={Christiano, Paul F and Leike, Jan and Brown, Tom and Martic, Miljan and Legg, Shane and Amodei, Dario},
  journal={Advances in neural information processing systems},
  volume={30},
  year={2017}
}

@inproceedings{stooke2020responsive,
  title={Responsive safety in reinforcement learning by pid lagrangian methods},
  author={Stooke, Adam and Achiam, Joshua and Abbeel, Pieter},
  booktitle={International Conference on Machine Learning},
  pages={9133--9143},
  year={2020},
  organization={PMLR}
}

@article{calvo2023state,
  title={State augmented constrained reinforcement learning: Overcoming the limitations of learning with rewards},
  author={Calvo-Fullana, Miguel and Paternain, Santiago and Chamon, Luiz FO and Ribeiro, Alejandro},
  journal={IEEE Transactions on Automatic Control},
  year={2023},
  publisher={IEEE}
}

@article{wang2026reward,
  title={Reward hacking in the era of large models: Mechanisms, emergent misalignment, challenges},
  author={Wang, Xiaohua and Tian, Muzhao and Zeng, Yuqi and Huang, Zisu and Yuan, Jiakang and Chen, Bowen and Xu, Jingwen and Zhou, Mingbo and Liu, Wenhao and Wu, Muling and others},
  journal={arXiv preprint arXiv:2604.13602},
  year={2026}
}

@article{helff2026llms,
  title={LLMs gaming verifiers: RLVR can lead to reward hacking},
  author={Helff, Lukas and Delfosse, Quentin and Steinmann, David and H{\"a}rle, Ruben and Shindo, Hikaru and Schramowski, Patrick and Stammer, Wolfgang and Kersting, Kristian and Friedrich, Felix},
  journal={arXiv preprint arXiv:2604.15149},
  year={2026}
}

@inproceedings{liu2020ipo,
  title={Ipo: Interior-point policy optimization under constraints},
  author={Liu, Yongshuai and Ding, Jiaxin and Liu, Xin},
  booktitle={Proceedings of the AAAI conference on artificial intelligence},
  volume={34},
  pages={4940--4947},
  year={2020}
}

@article{ghadimi2013stochastic,
  title={Stochastic first-and zeroth-order methods for nonconvex stochastic programming},
  author={Ghadimi, Saeed and Lan, Guanghui},
  journal={SIAM journal on optimization},
  volume={23},
  number={4},
  pages={2341--2368},
  year={2013},
  publisher={SIAM}
}

@inproceedings{kohler2017sub,
  title={Sub-sampled cubic regularization for non-convex optimization},
  author={Kohler, Jonas Moritz and Lucchi, Aurelien},
  booktitle={International Conference on Machine Learning},
  pages={1895--1904},
  year={2017},
  organization={PMLR}
}

@article{sutton1999policy,
  title={Policy gradient methods for reinforcement learning with function approximation},
  author={Sutton, Richard S and McAllester, David and Singh, Satinder and Mansour, Yishay},
  journal={Advances in neural information processing systems},
  volume={12},
  year={1999}
}

@article{qwen2025qwen25,
  title   = {Qwen2.5 Technical Report},
  author  = {{Qwen Team}},
  journal = {arXiv preprint arXiv:2412.15115},
  year    = {2024}
}

@inproceedings{huang2024one_shot,
 author = {Huang, Xinmeng and Li, Shuo and Dobriban, Edgar and Bastani, Osbert and Hassani, Hamed and Ding, Dongsheng},
 booktitle = {Advances in Neural Information Processing Systems},
 doi = {10.52202/079017-2680},
 editor = {A. Globerson and L. Mackey and D. Belgrave and A. Fan and U. Paquet and J. Tomczak and C. Zhang},
 pages = {84350--84383},
 publisher = {Curran Associates, Inc.},
 title = {One-Shot Safety Alignment for Large Language Models via Optimal Dualization},
 url = {https://proceedings.neurips.cc/paper_files/paper/2024/file/9979a69d2613ab98ad25d3849068f9f0-Paper-Conference.pdf},
 volume = {37},
 year = {2024}
}

@article{ji2024pku,
  title   = {{PKU-SafeRLHF}: Towards Multi-Level Safety Alignment for {LLMs} with Human Preference},
  author  = {Ji, Jiaming and Hong, Donghai and Zhang, Borong and Chen, Boyuan and Dai, Josef and Zheng, Boren and
             Qiu, Tianyi and Li, Boxun and Yang, Yaodong},
  journal = {arXiv preprint arXiv:2406.15513},
  year    = {2024}
}

\clearpage
\appendix
\thispagestyle{empty}

% Supplementary material: To improve readability, you must use a single-column format for the supplementary material.
\onecolumn
\aistatstitle{Appendices}

% \red{
% Note: You can choose whether to include you appendices as part of the main submission file (here) OR submit them separately as part of the supplementary material. The preferred option is to include appendices as part of the main submission file. It is the authors' responsibility that any supplementary material does not conflict in content with the main paper (e.g., the separately uploaded additional material is not an updated version of the one appended to the manuscript).

% Formatting Instructions:

% The appendices follow the same formatting instructions as in the main paper.
% The only difference is that the supplementary material must be in a \emph{single-column} format.
% }

%%%%%%%%%%%%%%%%%%%%%%%%%%%%%%%%%%%%%%%%%%%%%%%%%%%%%%%%%%%%%%%%%%%%%%%%%%%%%%%%%%%%%%%%%%%%%%%%%%%%%%%%%%%%%%%%%%%%%%%%%%%%%%%%%%%%%%%%%%%%%%%%%%%%%%%%%%%%%%%%%%%%%%%%%%%%%%%%%%%%%%%%%%%%%%%%%%%%%%%%%%%%%%%%%%%%%%%%%%%%%%%%%%%%%%%%%%%%%%%%%%%%%%%%%%%%%%%%%%%%%%%%%%%%%%%%%%%%%%%%%%%%%%%%%%%%%%%%%%%%%%%%%%%%%%%%%%%%%%%%%%%%%%%%%%%%%%%%%%%%%%%%%%%%%%%%%%%%%%%%%%%%%%%%%%

\section{Related Work}
\label{sec:related}
\textbf{RLHF and Safe-RLHF.} Most analyses of RLHF \citep{zhan2023provable,zhu2023principled,wu2023making,du2024exploration} fit a reward through the Bradley--Terry model \citep{bradley1952rank} and then optimize  reward-based objectives. \citet{huang2024one_shot,zhang2024zeroth,zhang2024reinforcement} estimate value differences directly through the inverse link and prove convergence to stationary points. \citet{dai2023safe} introduced Safe-RLHF, and \citet{liu2026policy} gave the first global guarantees, for one constraint, through natural and zeroth-order primal--dual methods, reaching averaged constraint satisfaction. 
% However, none of these work controls the safety of individual iterates during training.
However, none of these works explicitly guarantees the safety of individual iterates throughout training.

\textbf{Constrained MDPs.} 
% Primal--dual and Lagrangian methods for CMDPs \citep{altman1999constrained,paternain2019constrained,achiam2017constrained,tessler2018reward,ding2020natural,ding2022ACC,zeng2022finite,xu2021crpo,mondal2024sample} bound the average violation, \citet{bai2022achieving2} achieve zero violation in an averaged sense and \citet{ding2024last} establish last-iterate convergence. 
Primal-dual and Lagrangian methods for CMDPs \citep{altman1999constrained,paternain2019constrained,achiam2017constrained,tessler2018reward,ding2020natural,ding2022ACC,zeng2022finite,xu2021crpo,mondal2024sample} typically provide guarantees on average constraint violation. \citet{bai2022achieving2} achieve zero constraint violation in an averaged sense, while \citet{ding2024last} establish last-iterate convergence.
Individual iterates of these methods overshoot the boundary \citep{stooke2020responsive,calvo2023state}. Lower bounds and safe posterior sampling are studied by \citet{vaswani2022near,kalagarla2023safe}.

\textbf{Safe exploration.} Feasibility during learning \citep{koller2019learning} has been obtained with Gaussian-process models \citep{sui2015safe,berkenkamp2017safe,wachi2018safe,koller2018learning}, Lyapunov functions \citep{chow2018lyapunov,chow2019lyapunov} and interior-point methods \citep{hinder2019polynomial,liu2020ipo}. \citet{usmanova2024log} introduced LB-SGD, a log-barrier method with feasible iterates for black-box optimization with unknown constraints, and \citet{ni2025safe} extended it to CMDPs with a gradient-dominance argument that yields global optimality. We inherit the interior-point mechanism but leverage the human feedback only.
% replace the first-order oracle with a zeroth-order one.
% since values and gradients are reconstructed from binary judgments through inverse links and zeroth-order smoothing, whose errors are amplified by the factor $d/\mu$ and must be controlled at every iterate.
% \weiqin{mention the other type of Safe exploration.}

\textbf{Zeroth-order methods.} Random-perturbation gradient estimators \citep{ghadimi2013stochastic,liu2018zeroth} underlie the RLHF estimators of \citet{zhang2024zeroth,liu2026policy}, who use the same direction for all trajectory pairs per iteration. 
% We instead use one direction per trajectory pair in each iteration, enabling the estimates concentrate around the target gradient.
We instead use one direction per trajectory pair in each iteration, enabling the resulting estimates to concentrate around the target gradient.

%%%%%%%%%%%%%%%%%%%%%%%%%%%%%%%%%%%%%%%%%%%%%%%%%%%%%%%%%%%%%%%%%%%%%%%%%%%%%%%%%%%%
%%%%%%%%%%%%%%%%%%%%%%%%%%%%%%%%%%%%%%%%%%%%%%%%%%%%%%%%%%%%%%%%%%%%%%%%%%%%%%%%%%%%
%%%%%%%%%%%%%%%%%%%%%%%%%%%%%%%%%%%%%%%%%%%%%%%%%%%%%%%%%%%%%%%%%%%%%%%%%%%%%%%%%%%%
%%%%%%%%%%%%%%%%%%%%%%%%%%%%%%%%%%%%%%%%%%%%%%%%%%%%%%%%%%%%%%%%%%%%%%%%%%%%%%%%%%%%
%%%%%%%%%%%%%%%%%%%%%%%%%%%%%%%%%%%%%%%%%%%%%%%%%%%%%%%%%%%%%%%%%%%%%%%%%%%%%%%%%%%%

\section{Value-Function Regularity (Proof of Lemma~\ref{lem:smooth})}
\label{app:regularity}

For $t\ge0$, let $\tau_{0:t}=(s_0,a_0,\dots,s_t,a_t)$ denote a trajectory prefix, whose law under $\pi_{\bt}$ is
\begin{align}
p_{\bt}(\tau_{0:t})=\rho(s_0)\prod_{k=0}^{t}\pi_{\bt}(a_k \,\vert\,  s_k)\prod_{k=0}^{t-1}P(s_{k+1} \,\vert\,  s_k,a_k),
\end{align}
and define the cumulative score and its derivative
\begin{align}
\Psi_t(\bt) \DefinedAs \sum_{k=0}^{t}\nabla_{\bt}\log\pi_{\bt}(a_k \,\vert\,  s_k)=\nabla_{\bt}\log p_{\bt}(\tau_{0:t}), \qquad \notag\\
\Phi_t(\bt) \DefinedAs \sum_{k=0}^{t}\nabla^2_{\bt}\log\pi_{\bt}(a_k \,\vert\,  s_k)=\nabla^2_{\bt}\log p_{\bt}(\tau_{0:t}).
\end{align}

\begin{Lemma}[Moment bounds for the cumulative score]
\label{lem:score}
Under Assumption~\ref{assum:smooth}, for all $t\ge0$ and $\bt\in\Theta$ we have $\|\Phi_t(\bt)\|\le(t+1)M_h$ almost surely, and
\begin{align}
\E_{\tau_{0:t}\sim p_{\bt}}\big[\|\Psi_t(\bt)\|^2\big]\le(t+1)M_g^2, \qquad \E_{\tau_{0:t}\sim p_{\bt}}\big[\|\Psi_t(\bt)\|\big]\le\sqrt{t+1}\,M_g.
\end{align}
\end{Lemma}
\begin{proof}
The bound on $\Phi_t$ follows from the triangle inequality. For $\Psi_t$, expand
\begin{align}
\E\|\Psi_t\|^2=\sum_{k=0}^{t}\E\|\nabla_{\bt}\log\pi_{\bt}(a_k \,\vert\,  s_k)\|^2+2\sum_{0\le j<k\le t}\E\big\langle\nabla_{\bt}\log\pi_{\bt}(a_j \,\vert\,  s_j),\nabla_{\bt}\log\pi_{\bt}(a_k \,\vert\,  s_k)\big\rangle.
\end{align}
For $j<k$, conditioning on $(\tau_{0:k-1},s_k)$ and using that $a_k\sim\pi_{\bt}(\cdot \,\vert\,  s_k)$,
\begin{align}
\E\big[\nabla_{\bt}\log\pi_{\bt}(a_k \,\vert\,  s_k)\,\big|\,\tau_{0:k-1},s_k\big]=\sum_{a\in\Ac}\pi_{\bt}(a \,\vert\,  s_k)\frac{\nabla_{\bt}\pi_{\bt}(a \,\vert\,  s_k)}{\pi_{\bt}(a \,\vert\,  s_k)}=\nabla_{\bt}\sum_{a\in\Ac}\pi_{\bt}(a \,\vert\,  s_k)=0,
\end{align}
so every cross term vanishes, and $\E\|\Psi_t\|^2\le(t+1)M_g^2$. The bound on $\E\|\Psi_t\|$ follows from Jensen's inequality.

\end{proof}

\begin{proof}[Proof of Lemma~\ref{lem:smooth}]
Fix $\diamond$ and write $\diamond_t \DefinedAs \diamond(s_t,a_t)\in[0,1]$.

\emph{Step 1 (derivatives of the truncated value).} For finite $H$, $V^{\bt,H}_\diamond(\rho)=\sum_{t=0}^{H-1}\gamma^t\E_{p_{\bt}}[\diamond_t]$, and $\E_{p_{\bt}}[\diamond_t]=\sum_{\tau_{0:t}}p_{\bt}(\tau_{0:t})\diamond_t$ is a finite sum of smooth functions of $\bt$. Differentiating term by term and using $\nabla p_{\bt}=p_{\bt}\nabla\log p_{\bt}$ and $\nabla^2p_{\bt}=p_{\bt}(\nabla^2\log p_{\bt}+\nabla\log p_{\bt}\nabla\log p_{\bt}^\top)$,
\begin{align}
\nabla_{\bt}\E_{p_{\bt}}[\diamond_t]=\E_{p_{\bt}}\big[\diamond_t\Psi_t(\bt)\big], \qquad
\nabla^2_{\bt}\E_{p_{\bt}}[\diamond_t]=\E_{p_{\bt}}\big[\diamond_t\big(\Phi_t(\bt)+\Psi_t(\bt)\Psi_t(\bt)^\top\big)\big]. \label{eq:gpomdp}
\end{align}
Hence, using $\diamond_t\in[0,1]$, Lemma~\ref{lem:score} and $\|\E[\diamond_t\Psi_t\Psi_t^\top]\|\le\E[\diamond_t\|\Psi_t\|^2]$,
\begin{align}
\|\nabla_{\bt} V^{\bt,H}_\diamond(\rho)\|&\le\sum_{t=0}^{H-1}\gamma^t\E\|\Psi_t\|\le M_g\sum_{t=0}^{\infty}\gamma^{t/2}\cdot\gamma^{t/2}\sqrt{t+1} \notag\\
&\le M_g\sqrt{\sum_{t\ge0}\gamma^t}\sqrt{\sum_{t\ge0}\gamma^t(t+1)}=\frac{M_g}{(1-\gamma)^{3/2}}=\Lv, \label{eq:lip-proof}\\
\|\nabla^2_{\bt} V^{\bt,H}_\diamond(\rho)\|&\le\sum_{t=0}^{H-1}\gamma^t\big(\E\|\Phi_t\|+\E\|\Psi_t\|^2\big)\le(M_h+M_g^2)\sum_{t\ge0}\gamma^t(t+1)=\frac{M_g^2+M_h}{(1-\gamma)^2}=\Mv, \label{eq:smooth-proof}
\end{align}
where we used the Cauchy--Schwarz inequality and $\sum_{t\ge0}\gamma^t(t+1)=(1-\gamma)^{-2}$. Since $\Theta$ is convex, the mean value theorem yields that $V^{\bt,H}_\diamond(\rho)$ is $\Lv$-Lipschitz and that its gradient is $\Mv$-Lipschitz, i.e., $V^{\bt,H}_\diamond(\rho)$ is $\Mv$-smooth, so that for all $\bt,\bt'\in\Theta$,
\begin{align}
\big|V^{\bt',H}_\diamond(\rho)-V^{\bt,H}_\diamond(\rho)-\langle\nabla_{\bt} V^{\bt,H}_\diamond(\rho),\bt'-\bt\rangle\big|\le\frac{\Mv}{2}\|\bt'-\bt\|^2. \label{eq:quad-bound}
\end{align}

\emph{Step 2 (the infinite-horizon value).} The series $\sum_t\gamma^t\E_{p_{\bt}}[\diamond_t]$, $\sum_t\gamma^t\E_{p_{\bt}}[\diamond_t\Psi_t]$ and $\sum_t\gamma^t\E_{p_{\bt}}[\diamond_t(\Phi_t+\Psi_t\Psi_t^\top)]$ converge uniformly on $\Theta$, because their terms are bounded in norm by $\gamma^t$, $\gamma^t\sqrt{t+1}M_g$ and $\gamma^t(t+1)(M_g^2+M_h)$, respectively, which are summable and independent of $\bt$. By the theorem on term-wise differentiation of uniformly convergent series, $V^{\bt}_\diamond(\rho)$ is twice continuously differentiable with $\nabla_{\bt} V^{\bt}_\diamond(\rho)=\sum_{t\ge0}\gamma^t\E[\diamond_t\Psi_t]$ and $\nabla^2_{\bt} V^{\bt}_\diamond(\rho)=\sum_{t\ge0}\gamma^t\E[\diamond_t(\Phi_t+\Psi_t\Psi_t^\top)]$, and the bounds \eqref{eq:lip-proof}, \eqref{eq:smooth-proof} and \eqref{eq:quad-bound} hold verbatim for $H=\infty$.

\emph{Step 3 (truncation errors).} By definition, $V^{\bt}_\diamond(\rho)-V^{\bt,H}_\diamond(\rho)=\sum_{t\ge H}\gamma^t\E[\diamond_t]\in[0,\sum_{t\ge H}\gamma^t]=[0,\epsilon_V(H)]$. For the gradient, by Step 2 and Lemma~\ref{lem:score},
\begin{align}
\|\nabla_{\bt} V^{\bt}_\diamond(\rho)-\nabla_{\bt} V^{\bt,H}_\diamond(\rho)\|
\le M_g\sum_{t\ge H}\gamma^t\sqrt{t+1}
\le M_g\sqrt{\sum_{t\ge H}\gamma^t}\sqrt{\sum_{t\ge H}\gamma^t(t+1)}.
\end{align}
Since $\sum_{t\ge H}\gamma^t=\frac{\gamma^H}{1-\gamma}$ and $\sum_{t\ge H}\gamma^t(t+1)=\gamma^H\sum_{k\ge0}\gamma^k(k+1+H)=\gamma^H\big(\frac{1}{(1-\gamma)^2}+\frac{H}{1-\gamma}\big)$, we obtain
\begin{align}
\|\nabla_{\bt} V^{\bt}_\diamond(\rho)-\nabla_{\bt} V^{\bt,H}_\diamond(\rho)\|\le M_g\sqrt{\frac{\gamma^{2H}(1+H(1-\gamma))}{(1-\gamma)^3}}=\frac{M_g\gamma^H\sqrt{1+H(1-\gamma)}}{(1-\gamma)^{3/2}}=\epsilon_\nabla(H). \qquad \qedhere
\end{align}

\end{proof}

%%%%%%%%%%%%%%%%%%%%%%%%%%%%%%%%%%%%%%%%%%%%%%%%%%%%%%%%%%
\section{Estimation Guarantees}
\label{app:estimation}

This appendix proves the estimation results of Sections~\ref{sec:estimators} and~\ref{sec:estimation_error}. We first establish the two properties of the extended inverse link (Lemma~\ref{lem:link}), then two auxiliary tools: a vector Bernstein inequality and the spherical-smoothing identity, then the coupling lemma (Lemma~\ref{lem:coupling}), which bounds the variance reduction achieved by the coupled rollouts, then the accuracy of the estimators (Lemma~\ref{lem:est}) and finally the error bound for the barrier gradient estimate (Theorem~\ref{theorem:barrier-err}), including the explicit constants $C_1,C_2$ of \eqref{eq:width-req}.

\subsection{Proof of Lemma~\ref{lem:link} (inverse-link estimation)}

\begin{proof}
Write $p^* \DefinedAs \sigma(x)\in\mathcal P_\sigma$ and $f \DefinedAs \sinv$, so that $x=\sigma^{-1}(p^*)=f(p^*)$ and $\hat x=f(\hat p)$.

\emph{Regularity of the extension.} On $\mathcal P_\sigma=[\underline{p},\overline{p}]$ we have $f=\sigma^{-1}$, hence $|f'|\le L$ and $f'$ is $L_2$-Lipschitz there (Assumption~\ref{assum:link}). On $[0,\underline{p})$ and $(\overline{p},1]$, $f$ is affine with slopes $(\sigma^{-1})'(\underline{p})$ and $(\sigma^{-1})'(\overline{p})$, both bounded by $L$ in absolute value. By construction $f$ and $f'$ are continuous at $\underline{p}$ and $\overline{p}$, so $f\in C^1([0,1])$ with $|f'|\le L$ on $[0,1]$, i.e., $f$ is $L$-Lipschitz on $[0,1]$.

We now show that $f'$ is $L_2$-Lipschitz on $[0,1]$, i.e., $|f'(b)-f'(a)|\le L_2(b-a)$ for all $0\le a<b\le1$. The interval $[0,1]$ is the union of the three pieces $[0,\underline p)$, $[\underline p,\overline p]$ and $(\overline p,1]$, on which $f'$ equals the constant $(\sigma^{-1})'(\underline p)$, the function $(\sigma^{-1})'$ and the constant $(\sigma^{-1})'(\overline p)$, respectively. We distinguish the following cases according to the pieces that contain $a$ and $b$.
\begin{enumerate}
\item[$(1)$] $a$ and $b$ lie in the same affine piece, i.e., $a<b<\underline p$ or $\overline p<a<b$. Then $f'$ is constant on the piece, so $f'(b)-f'(a)=0$.
\item[$(2)$] $a$ and $b$ lie in $[\underline p,\overline p]$. Then $f'=(\sigma^{-1})'$ on $[a,b]$, and since $|(\sigma^{-1})''|\le L_2$ on $\mathcal P_\sigma$ by Assumption~\ref{assum:link}, the mean value theorem gives $|f'(b)-f'(a)|\le L_2(b-a)$.
\item[$(3)$] $a<\underline p\le b\le\overline p$. Then $f'(a)=(\sigma^{-1})'(\underline p)=f'(\underline p)$ by construction, and case $(2)$ applied to the points $\underline p\le b$ gives
\begin{align}
|f'(b)-f'(a)|=|f'(b)-f'(\underline p)|\le L_2(b-\underline p)\le L_2(b-a).
\end{align}
\item[$(4)$] $\underline p\le a\le\overline p<b$. Symmetrically, $f'(b)=f'(\overline p)$ and case $(2)$ gives $|f'(b)-f'(a)|=|f'(\overline p)-f'(a)|\le L_2(\overline p-a)\le L_2(b-a)$.
\item[$(5)$] $a<\underline p$ and $b>\overline p$. Then $f'(a)=f'(\underline p)$ and $f'(b)=f'(\overline p)$, and case $(2)$ gives $|f'(b)-f'(a)|=|f'(\overline p)-f'(\underline p)|\le L_2(\overline p-\underline p)\le L_2(b-a)$.
\end{enumerate}
In every case $|f'(b)-f'(a)|\le L_2(b-a)$, which proves the claim.

\emph{Almost-sure bound.} Since $f$ is $L$-Lipschitz on $[0,1]$ and $\hat p,p^*\in[0,1]$, $|\hat x-x|=|f(\hat p)-f(p^*)|\le L|\hat p-p^*|\le L$.

\emph{Bias bound.} Since $f'$ is $L_2$-Lipschitz on $[0,1]$, for every $p\in[0,1]$,
\begin{align}
\big|f(p)-f(p^*)-f'(p^*)(p-p^*)\big|=\Big|\int_{p^*}^{p}\big(f'(u)-f'(p^*)\big)\,du\Big|\le\frac{L_2}{2}(p-p^*)^2.
\end{align}
Taking $p=\hat p$ and expectations, and using $\E[\hat p]=p^*$ and $\E[(\hat p-p^*)^2]=\frac{p^*(1-p^*)}{M}\le\frac{1}{4M}$,
\begin{align}
|\E[\hat x]-x|
&=\big|\E[f(\hat p)]-f(p^*)\big|
=\big|\E[f(\hat p)-f(p^*)]-f'(p^*)\,\E[\hat p-p^*]\big| \notag\\
&=\big|\E[f(\hat p)-f(p^*)-f'(p^*)(\hat p-p^*)]\big|
\le\E\big[\big|f(\hat p)-f(p^*)-f'(p^*)(\hat p-p^*)\big|\big] \notag\\
&\le\frac{L_2}{2}\E[(\hat p-p^*)^2]
=\frac{L_2\,p^*(1-p^*)}{2M}
\le\frac{L_2}{8M}=\beta_M.
\end{align}

\end{proof}

\subsection{Auxiliary lemmas}

\begin{Lemma}[Vector Bernstein inequality {\citep[Lemma~18]{kohler2017sub}}]
\label{lem:bernstein}
Let $Y_1,\dots,Y_N$ be independent random vectors in $\R^d$ with $\E[Y_n]=0$, $\|Y_n\|\le B$ and $\E\|Y_n\|^2\le\sigma^2$. Then for every $0<\epsilon\le\sigma^2/B$,
\begin{align}
\PP\Big(\Big\|\frac1N\sum_{n=1}^NY_n\Big\|\ge\epsilon\Big)\le\exp\Big(\frac14-\frac{N\epsilon^2}{8\sigma^2}\Big).
\end{align}
Equivalently, for $\delta'\in(0,1)$, with probability at least $1-\delta'$, it holds that $\|\frac1N\sum_{n=1}^N Y_n\|\le \epsilon = \sqrt{8\sigma^2(\log\frac{1}{\delta'}+\frac14)/N}$. Given that $\epsilon \leq \sigma^2/B$, it follows that $N\ge8B^2(\log\frac1{\delta'}+\frac14)/\sigma^2$.
\end{Lemma}

\begin{Lemma}[Spherical smoothing]
\label{lem:zo}
Let $f:\Theta\to\R$ be $M$-smooth on a convex set $\Theta\subseteq\R^d$, let $\bt\in\Theta$ and $\mu>0$ be such that the closed ball of radius $\mu$ around $\bt$ is contained in $\Theta$, and let $\bv\sim\mathrm{Unif}(\mathbb V^{d-1})$ be uniformly distributed on the unit sphere. Then
\begin{align}
\Big\|\E_{\bv}\Big[\frac{d}{\mu}\big(f(\bt+\mu\bv)-f(\bt)\big)\bv\Big]-\nabla f(\bt)\Big\|\le\frac{dM\mu}{2}.
\end{align}
\end{Lemma}
\begin{proof}
The assumption on the ball guarantees that for every unit vector $\bv$ the whole segment $\{\bt+u\bv:u\in[0,\mu]\}$ lies in $\Theta$, so that the smoothness of $f$ can be used along it. For $\Theta=\R^d$, as in the main text, this assumption is void. We proceed in three steps.

\emph{Step 1 (Taylor expansion along each direction).} Fix a unit vector $\bv$ and define $\zeta(\bv) \DefinedAs f(\bt+\mu\bv)-f(\bt)-\mu\langle\nabla f(\bt),\bv\rangle$. Since $f$ is $M$-smooth and the segment between $\bt$ and $\bt+\mu\bv$ lies in $\Theta$, it holds that
\begin{align}
|\zeta(\bv)|\le\frac{M}{2}\|\mu\bv\|^2=\frac{M\mu^2}{2}, \qquad\text{for every unit vector }\bv .
\end{align}

\emph{Step 2 (second moment of a uniform unit vector).} We claim that $\E_{\bv}[\bv\bv^\top]=\frac1dI_d$. The matrix $\Sigma \DefinedAs \E_{\bv}[\bv\bv^\top]$ is symmetric, and for every orthogonal matrix $Q$ the vector $Q\bv$ is again uniformly distributed on the unit sphere, so $\Sigma=\E[(Q\bv)(Q\bv)^\top]=Q\Sigma Q^\top$. A symmetric matrix that commutes with every orthogonal matrix is a multiple of the identity, $\Sigma=\kappa I_d$, and taking the trace gives $d\kappa=\E[\|\bv\|^2]=1$, hence $\kappa=\frac1d$.

\emph{Step 3 (combining the two steps).} Multiplying the identity $f(\bt+\mu\bv)-f(\bt)=\mu\langle\nabla f(\bt),\bv\rangle+\zeta(\bv)$ by $\frac d\mu\bv$ and taking expectations yields
\begin{align}
\E_{\bv}\Big[\frac{d}{\mu}\big(f(\bt+\mu\bv)-f(\bt)\big)\bv\Big]
=d\,\E_{\bv}\big[\bv\bv^\top\big]\nabla f(\bt)+\frac d\mu\E_{\bv}[\zeta(\bv)\bv]
=\nabla f(\bt)+\frac d\mu\E_{\bv}[\zeta(\bv)\bv],
\end{align}
where we used $\langle\nabla f(\bt),\bv\rangle\bv=\bv\bv^\top\nabla f(\bt)$ and Step~2. Therefore, by Jensen's inequality, $\|\bv\|=1$ and Step~1,
\begin{align}
\Big\|\E_{\bv}\Big[\frac{d}{\mu}\big(f(\bt+\mu\bv)-f(\bt)\big)\bv\Big]-\nabla f(\bt)\Big\|
=\frac d\mu\big\|\E_{\bv}[\zeta(\bv)\bv]\big\|
\le\frac d\mu\E_{\bv}\big[|\zeta(\bv)|\,\|\bv\|\big]
\le\frac d\mu\cdot\frac{M\mu^2}{2}=\frac{dM\mu}{2}. \qquad \qedhere
\end{align}
\end{proof}

\subsection{Coupled trajectories rarely differ}

The following lemma is the source of the term $\Gbar^2H|\Ac|M_g\mu$ in the variance proxy $\sigma_g^2$ of \eqref{eq:sigma-g}, since the two trajectories of a pair in Phase 2 are generated at parameters that are $\mu$ apart.
\begin{Lemma}[Coupled trajectories]
\label{lem:coupling}
Under Assumptions~\ref{assum:seeds} and~\ref{assum:smooth}, for $\bt,\bt'\in\Theta$ and a seed $\omega$,
\begin{align}
\PP\big(\tau(\bt,\omega)\ne\tau(\bt',\omega)\big)\le H|\Ac|M_g\|\bt-\bt'\|,\quad
\E\big[(R^H_\diamond(\tau(\bt',\omega))-R^H_\diamond(\tau(\bt,\omega)))^2\big]\le\Gbar^2\min\{1,H|\Ac|M_g\|\bt-\bt'\|\}.
\end{align}
\end{Lemma}
\begin{proof}
Throughout the proof, $\omega=(s_0,(u_h,u'_h)_{h<H})$ is a fixed seed with i.i.d.\ $\mathrm{Unif}[0,1]$ components, and all probabilities and expectations are with respect to $\omega$. Write $\tau=\tau(\bt,\omega)=\{(s_h,a_h)\}_{h<H}$ and $\tau'=\tau(\bt',\omega)=\{(s'_h,a'_h)\}_{h<H}$ for the two coupled trajectories, and recall from Assumption~\ref{assum:seeds} that both are generated by the same deterministic rules $a_h=F^{-1}_{\pi_{\bt}(\cdot\,\vert\,s_h)}(u_h)$, $s_{h+1}=\Phi(s_h,a_h,u'_h)$ and $a'_h=F^{-1}_{\pi_{\bt'}(\cdot\,\vert\,s'_h)}(u_h)$, $s'_{h+1}=\Phi(s'_h,a'_h,u'_h)$, with $s_0=s'_0$. Here, for a distribution $q$ on $\Ac$ and the fixed ordering $a^{(1)}\prec\cdots\prec a^{(|\Ac|)}$ of $\Ac$, $F_q(a^{(k)}) \DefinedAs \sum_{j\le k}q(a^{(j)})$ is the cumulative distribution function and $F_q^{-1}(u) \DefinedAs a^{(k)}$ for the smallest $k$ with $F_q(a^{(k)})\ge u$. We proceed in four steps.

\emph{Step 1 (the two trajectories agree until the first disagreement of the actions).} Define the first disagreement time $h^* \DefinedAs \min\{h<H:a_h\ne a'_h\}$, with $h^* \DefinedAs H$ if the actions agree at every step. We claim that $s_h=s'_h$ and $a_h=a'_h$ for all $h<h^*$, and that $s_h=s'_h$ also for $h=h^*$. This follows by induction on $h$. For $h=0$, $s_0=s'_0$ by construction. If $s_h=s'_h$ for some $h<h^*$, then $a_h=a'_h$ by the definition of $h^*$, and since the next states are computed from the same state, the same action and the same uniform $u'_h$ through the same function $\Phi$, we get $s_{h+1}=\Phi(s_h,a_h,u'_h)=\Phi(s'_h,a'_h,u'_h)=s'_{h+1}$. In particular, $\tau\ne\tau'$ implies $h^*<H$, so that
\begin{align}
\label{eq:coupling-union}
\PP(\tau\ne\tau')\le\PP(h^*<H)\le\sum_{h=0}^{H-1}\PP\big(a_h\ne a'_h,\ s_h=s'_h\big),
\end{align}
where we used that $h^*=h$ requires $a_h\ne a'_h$ together with $s_h=s'_h$, and a union bound over $h$.

\emph{Step 2 (probability of a disagreement at a given step).} Fix $h<H$ and condition on the event $\{s_h=s'_h=s\}$ for some state $s$. This event depends only on $(s_0,(u_k,u'_k)_{k<h})$, so it is independent of the uniform $u_h$. On this event, both actions are obtained from the same uniform $u_h$ by inverse-CDF sampling from $p \DefinedAs \pi_{\bt}(\cdot\,\vert\,s)$ and $p' \DefinedAs \pi_{\bt'}(\cdot\,\vert\,s)$, respectively. Inverse-CDF sampling returns $a^{(k)}$ exactly when $u_h\in(F_p(a^{(k-1)}),F_p(a^{(k)})]$ (with $F_p(a^{(0)}) \DefinedAs 0$), and similarly for $p'$. If $u_h$ belongs to the interval $(F_p(a^{(k-1)}),F_p(a^{(k)})]$ for $p$ and to the interval $(F_{p'}(a^{(k-1)}),F_{p'}(a^{(k)})]$ for $p'$ with the same index $k$, the two actions coincide. Hence the actions can differ only if $u_h$ lies between $F_p(a^{(k)})$ and $F_{p'}(a^{(k)})$ for some $k$, i.e., in the set $\bigcup_{k=1}^{|\Ac|}\big(\min\{F_p(a^{(k)}),F_{p'}(a^{(k)})\},\max\{F_p(a^{(k)}),F_{p'}(a^{(k)})\}\big]$, whose Lebesgue measure is at most $\sum_{k=1}^{|\Ac|}|F_p(a^{(k)})-F_{p'}(a^{(k)})|$. Since $u_h$ is uniform on $[0,1]$ and independent of the conditioning event,
\begin{align}
\label{eq:coupling-step}
\PP\big(a_h\ne a'_h\,\big\vert\,s_h=s'_h=s\big)\le\sum_{k=1}^{|\Ac|}\big|F_p(a^{(k)})-F_{p'}(a^{(k)})\big|
\le\sum_{k=1}^{|\Ac|}\sum_{j\le k}\big|p(a^{(j)})-p'(a^{(j)})\big|
\le|\Ac|\,\|p-p'\|_1,
\end{align}
where the second inequality is the triangle inequality applied to the partial sums that define $F_p$ and $F_{p'}$, and the third one bounds each of the $|\Ac|$ partial sums by the full sum $\|p-p'\|_1=\sum_{a\in\Ac}|p(a)-p'(a)|$.

\emph{Step 3 (the total variation distance between the two policies at a state).} We bound $\|\pi_{\bt}(\cdot\,\vert\,s)-\pi_{\bt'}(\cdot\,\vert\,s)\|_1$. Since $\Theta$ is convex, the segment $\bar\bt(r) \DefinedAs \bt'+r(\bt-\bt')$, $r\in[0,1]$, lies in $\Theta$, and for every action $a$ the fundamental theorem of calculus gives
\begin{align}
\pi_{\bt}(a\,\vert\,s)-\pi_{\bt'}(a\,\vert\,s)=\int_0^1\big\langle\nabla_{\bt}\pi_{\bar\bt(r)}(a\,\vert\,s),\bt-\bt'\big\rangle\,dr .
\end{align}
Taking absolute values, summing over $a$, using the Cauchy--Schwarz inequality and exchanging sum and integral,
\begin{align}
\|\pi_{\bt}(\cdot\,\vert\,s)-\pi_{\bt'}(\cdot\,\vert\,s)\|_1
\le\|\bt-\bt'\|\int_0^1\sum_{a\in\Ac}\|\nabla_{\bt}\pi_{\bar\bt(r)}(a\,\vert\,s)\|\,dr
\le\|\bt-\bt'\|\sup_{r\in[0,1]}\sum_{a\in\Ac}\|\nabla_{\bt}\pi_{\bar\bt(r)}(a\,\vert\,s)\| .
\end{align}
For every parameter $\bar\bt$ we have $\nabla_{\bt}\pi_{\bar\bt}(a\,\vert\,s)=\pi_{\bar\bt}(a\,\vert\,s)\nabla_{\bt}\log\pi_{\bar\bt}(a\,\vert\,s)$, so by Assumption~\ref{assum:smooth}
\begin{align}
\sum_{a\in\Ac}\|\nabla_{\bt}\pi_{\bar\bt}(a\,\vert\,s)\|=\sum_{a\in\Ac}\pi_{\bar\bt}(a\,\vert\,s)\,\|\nabla_{\bt}\log\pi_{\bar\bt}(a\,\vert\,s)\|\le M_g\sum_{a\in\Ac}\pi_{\bar\bt}(a\,\vert\,s)=M_g .
\end{align}
Note that the supremum over $r$ is taken of the whole sum, which is what makes the bound $M_g$ rather than $|\Ac|M_g$. Combining the last two displays,
\begin{align}
\label{eq:coupling-tv}
\|\pi_{\bt}(\cdot\,\vert\,s)-\pi_{\bt'}(\cdot\,\vert\,s)\|_1\le M_g\|\bt-\bt'\|, \qquad\text{for every }s\in\St .
\end{align}

\emph{Step 4 (conclusion).} By \eqref{eq:coupling-step} and \eqref{eq:coupling-tv}, for every $h<H$,
\begin{align}
\PP\big(a_h\ne a'_h,\ s_h=s'_h\big)=\sum_{s\in\St}\PP(s_h=s'_h=s)\,\PP\big(a_h\ne a'_h\,\big\vert\,s_h=s'_h=s\big)&\le|\Ac|M_g\|\bt-\bt'\|\sum_{s\in\St}\PP(s_h=s'_h=s) \notag \\ 
&\le|\Ac|M_g\|\bt-\bt'\| .
\end{align}
Inserting the previous expression into \eqref{eq:coupling-union} gives the first claim, $\PP(\tau\ne\tau')\le H|\Ac|M_g\|\bt-\bt'\|$. For the second claim, let $D \DefinedAs R^H_\diamond(\tau')-R^H_\diamond(\tau)$. On the event $\{\tau=\tau'\}$ we have $D=0$, and on its complement $|D|\le G(H)\le\Gbar$ because both returns lie in $[0,G(H)]$. Hence
\begin{align}
\E[D^2]=\E\big[D^2\mathbf 1\{\tau\ne\tau'\}\big]\le\Gbar^2\,\PP(\tau\ne\tau')\le\Gbar^2H|\Ac|M_g\|\bt-\bt'\| .
\end{align}
Since $\E[D^2]\le\Gbar^2$ trivially as well, the second claim follows by taking the minimum of the two bounds.

\end{proof}

\subsection{Proof of Lemma~\ref{lem:est} (accuracy of the estimators)}

We decompose the widths of Section~\ref{sec:stepsize} into their bias and concentration parts, $\wc=b_c+s_c\sqrt{\iot}$ and $\wg=b_g+s_g\sqrt{\iot}$ with
\begin{align}
b_c \DefinedAs \epsilon_V(H)+\beta_{M_c},\quad s_c \DefinedAs (\Gbar+L)\sqrt{2/N_c},\quad
b_g \DefinedAs \epsilon_\nabla(H)+\frac{d\Mv\mu}{2}+\frac{d\beta_M}{\mu},\quad s_g \DefinedAs \sqrt{8\sigma_g^2/N},
\end{align}
and write $B_g \DefinedAs \frac{d(\Gbar+L)}{\mu}$ for the range bound of the gradient estimator. Lemma~\ref{lem:est} follows from the following more detailed statement by a union bound.

\begin{Lemma}[Bias and concentration of the estimators]
\label{lem:est-detailed}
Let Assumptions~\ref{assum:link},~\ref{assum:seeds} and~\ref{assum:smooth} hold and let \eqref{eq:range} hold. Conditionally on $\mathcal F_t$, the following holds for every $\diamond\in\{r\}\cup\{g_i\}_{i=1}^m$ and $i\in[m]$.
\begin{enumerate}
\item[$\text{\normalfont i)}$] $\|\E[\hb^{(t)}_\diamond]-\nabla_{\bt}V^{\bt^{(t)}}_\diamond(\rho)\|\le b_g$ and $|\E[\widehat h^{(t)}_{c_i}]-\bar V^{\bt^{(t)}}_{g_i}(\rho)|\le b_c$.
\item[$\text{\normalfont ii)}$] For each fixed $\diamond$, $\|\hb^{(t)}_\diamond-\E[\hb^{(t)}_\diamond]\|\le s_g\sqrt{\iot}$ with probability at least $1-\frac{\delta}{2m+1}$, and for each fixed $i$, $|\widehat h^{(t)}_{c_i}-\E[\widehat h^{(t)}_{c_i}]|\le s_c\sqrt{\iot}$ with probability at least $1-\frac{\delta}{2m+1}$.
\end{enumerate}
\end{Lemma}

\begin{proof}[Proof of Lemmas~\ref{lem:est-detailed} and~\ref{lem:est}]
Condition on $\mathcal F_t$, write $\bt \DefinedAs \bt^{(t)}$ and drop the superscript $(t)$. Within Phases 1 and 2, the tuples $\xi_n \DefinedAs (\omega_n,\bv_n,\tau_{n,0},\tau_{n,1},\{o_{\diamond,n,j}\}_{\diamond,j},\{o_{c_i,n,j}\}_{i,j})$, $n\in[N]$, are i.i.d., and so are the base tuples $(\omega_n,\tau_{n,0},\{o_{c_i,n,j}\}_{i,j})$, $n\in[N_c]$. The coupling of $\tau_{n,0}$ and $\tau_{n,1}$ through the common seed $\omega_n$ does not affect the marginal law of either trajectory, so $\tau_{n,0}\sim\pi_{\bt}$ and, conditionally on $\bv_n$, $\tau_{n,1}\sim\pi_{\bt+\mu\bv_n}$, both of length $H$ with initial state drawn from $\rho$.

\emph{Gradient estimators, bias.} Fix $\diamond$. Let $\Delta R_n \DefinedAs R^H_\diamond(\tau_{n,1})-R^H_\diamond(\tau_{n,0})\in[-G(H),G(H)]\subset\bar{\mathcal I}$, $\hat x_n \DefinedAs \sinv_\diamond(\hat p_{\diamond,n})$ and $X_n \DefinedAs \frac d\mu\hat x_n\bv_n$, so that $\hb_\diamond=\frac1N\sum_{n=1}^NX_n$. Conditionally on $(\bv_n,\tau_{n,0},\tau_{n,1})$, the feedback bits $o_{\diamond,n,1},\dots,o_{\diamond,n,M}$ are i.i.d.\ Bernoulli$(\sigma_\diamond(\Delta R_n))$, so Lemma~\ref{lem:link} gives $\E[\hat x_n \,\vert\, \bv_n,\tau_{n,0},\tau_{n,1}]=\Delta R_n+\beta_n$ with $|\beta_n|\le\beta_M$. By the marginal laws above, $\E[\Delta R_n \,\vert\, \bv_n]=V^{\bt+\mu\bv_n,H}_\diamond(\rho)-V^{\bt,H}_\diamond(\rho)$. By the tower property,
\begin{align}
\E[X_n]=\frac d\mu\E_{\bv_n}\Big[\big(V^{\bt+\mu\bv_n,H}_\diamond(\rho)-V^{\bt,H}_\diamond(\rho)\big)\bv_n\Big]+\frac d\mu\E[\beta_n\bv_n].
\end{align}
By Lemma~\ref{lem:smooth}, $\bt\mapsto V^{\bt,H}_\diamond(\rho)$ is $\Mv$-smooth, so Lemma~\ref{lem:zo} bounds the distance of the first term to $\nabla_{\bt}V^{\bt,H}_\diamond(\rho)$ by $\frac{d\Mv\mu}{2}$, Lemma~\ref{lem:smooth} bounds $\|\nabla_{\bt}V^{\bt,H}_\diamond(\rho)-\nabla_{\bt}V^{\bt}_\diamond(\rho)\|\le\epsilon_\nabla(H)$, and $\|\frac d\mu\E[\beta_n\bv_n]\|\le\frac{d\beta_M}{\mu}$. Since $\E[\hb_\diamond]=\E[X_n]$, the triangle inequality gives $\|\E[\hb_\diamond]-\nabla_{\bt}V^{\bt}_\diamond(\rho)\|\le b_g$.

\emph{Gradient estimators, concentration.} By Lemma~\ref{lem:link}, $|\hat x_n|\le|\Delta R_n|+|\hat x_n-\Delta R_n|\le\Gbar+L$, so $\|X_n\|\le B_g=\frac{d(\Gbar+L)}{\mu}$ and $\|X_n-\E X_n\|\le2B_g$. For the second moment, the pair $(\tau_{n,0},\tau_{n,1})$ is coupled with parameter distance $\|(\bt+\mu\bv_n)-\bt\|=\mu$, so Lemma~\ref{lem:coupling} gives $\E[(\Delta R_n)^2]\le\Gbar^2H|\Ac|M_g\mu$, and Lemma~\ref{lem:link} gives $\E[(\hat x_n-\Delta R_n)^2 \,\vert\, \tau_{n,0},\tau_{n,1}]\le L^2\E[(\hat p_{\diamond,n}-\sigma_\diamond(\Delta R_n))^2 \,\vert\, \tau_{n,0},\tau_{n,1}]\le\frac{L^2}{4M}$. Since $\|\bv_n\|=1$ and $(a+b)^2\le2a^2+2b^2$,
\begin{align}
\E\|X_n-\E X_n\|^2\le\E\|X_n\|^2=\frac{d^2}{\mu^2}\E[\hat x_n^2]&\le\frac{2d^2}{\mu^2}\Big(\E[(\Delta R_n)^2]+\E[(\hat x_n-\Delta R_n)^2]\Big)\le\sigma_g^2.
\end{align}
Lemma~\ref{lem:bernstein} applied to the i.i.d.\ centered vectors $Y_n=X_n-\E X_n$ with $B=2B_g$ and $\delta'=\frac{\delta}{2m+1}$ yields, with probability at least $1-\delta'$, $\|\hb_\diamond-\E\hb_\diamond\|\le\sqrt{8\sigma_g^2(\log\frac1{\delta'}+\frac14)/N}\le s_g\sqrt{\iot}$. Its range condition $N\ge8(2B_g)^2(\log\frac1{\delta'}+\frac14)/\sigma_g^2$ is implied by \eqref{eq:range}.

\emph{Constraint margin estimators.} Fix $i\in[m]$, let $x_n \DefinedAs R^H_{g_i}(\tau_{n,0})-b_i$ and $Z_n \DefinedAs \sinv_{c_i}(\hat p_{c_i,n})$ for $n\in[N_c]$, so that $\widehat h_{c_i}=\frac1{N_c}\sum_nZ_n$. Since $R^H_{g_i}(\tau_{n,0})\in[0,G(H)]$ and $b_i\in(0,\Gbar)$, $x_n\in\bar{\mathcal I}$. Conditionally on $\tau_{n,0}$, the bits $o_{c_i,n,j}$ are i.i.d.\ Bernoulli$(\sigma_{c_i}(x_n))$, so by Lemma~\ref{lem:link} (with $M_c$ evaluators), $|Z_n-x_n|\le L$ almost surely and $|\E[Z_n \,\vert\, \tau_{n,0}]-x_n|\le\beta_{M_c}$. Moreover $\E[x_n]=V^{\bt,H}_{g_i}(\rho)-b_i$, which differs from $\bar V^{\bt}_{g_i}(\rho)$ by at most $\epsilon_V(H)$ (Lemma~\ref{lem:smooth}). Hence $|\E[\widehat h_{c_i}]-\bar V^{\bt}_{g_i}(\rho)|\le b_c$. For the concentration, $Z_n\in[-\Gbar-L,\Gbar+L]$, an interval of length $2(\Gbar+L)$, and the $Z_n$ are i.i.d., so Hoeffding's inequality gives $\PP(|\widehat h_{c_i}-\E\widehat h_{c_i}|\ge\epsilon)\le2\exp(-\frac{2N_c\epsilon^2}{4(\Gbar+L)^2})$, i.e., with probability at least $1-\delta'$,
\begin{align}
|\widehat h_{c_i}-\E[\widehat h_{c_i}]|\le(\Gbar+L)\sqrt{\frac{2\log(2/\delta')}{N_c}}\le s_c\sqrt{\iot}.
\end{align}

\emph{Union bound.} The $2m+1$ concentration events ($\hb_r$, $\hb_{g_1},\dots,\hb_{g_m}$, $\widehat h_{c_1},\dots,\widehat h_{c_m}$) each fail with conditional probability at most $\frac{\delta}{2m+1}$. On their intersection, which has conditional probability at least $1-\delta$ given $\mathcal F_t$, the triangle inequality with the bias bounds gives $\|\hb_\diamond-\nabla_{\bt} V_\diamond^{\bt}(\rho)\|\le b_g+s_g\sqrt{\iot}=\wg$ and $|\widehat h_{c_i}-\bar V^{\bt}_{g_i}(\rho)|\le b_c+s_c\sqrt{\iot}=\wc$, i.e., $\mathcal E_t$ holds.

\end{proof}

% \begin{remark}[Independent instead of coupled trajectories]
% \label{rem:independent}
% If the two trajectories of a pair are sampled independently, the bias bounds above are unchanged (they only use marginal laws), and the second moment is bounded by the trivial estimate $\E\|X_n\|^2\le B_g^2$. Lemma~\ref{lem:bernstein} with $\sigma^2=B^2$ (whose range condition reduces to $N\ge8\iot$) then yields the width $\sqrt{8B_g^2\iot/N}=\frac{2\sqrt2d(\Gbar+L)}{\mu\sqrt N}\sqrt{\iot}$, i.e., Lemma~\ref{lem:est} with $\sigma_g^2$ replaced by $B_g^2$. This wider width is a valid confidence width, so the safety argument of Appendix~\ref{app:safe} applies to it unchanged. The variant of ZO-LBPG without coupling in Table~\ref{tab:results} uses it.
% \end{remark}

\subsection{Proof of Theorem~\ref{theorem:barrier-err} (log-barrier gradient error)}

The constants in \eqref{eq:width-req} are
\begin{align}
\label{eq:C12}
C_1 \DefinedAs \frac{3}{8(3c+4m)},\qquad C_2 \DefinedAs \frac{3c}{64m\Lv},
\end{align}
so that $\bar w_g=\frac{3\min\{c\eta,2\ell\}}{8(3c+4m)}$ and $\bar w_c=\frac{3c\eta\min\{c\eta,2\ell\}}{64m\Lv}$. For $\eta\le2\ell/c$ these reduce to $\bar w_g=\frac{3c\eta}{8(3c+4m)}$ and $\bar w_c=\frac{3c^2\eta^2}{64m\Lv}$. The form $\min\{c\eta,2\ell\}$ ensures that the conditions of Lemma~\ref{lem:product} below are met for every $\eta\le1$.

\begin{proof}
On $\mathcal E_t$, $|\widehat h^{(t)}_{c_i}-\alpha_i(t)|\le\wc$ for all $i$, so $\underline\alpha_i(t)=\widehat h^{(t)}_{c_i}-\wc\in[\alpha_i(t)-2\wc,\alpha_i(t)]$. If $\wc\le\alpha_{\min}(t)/8$, then $\underline\alpha_i(t)\ge\alpha_i(t)-\frac{\alpha_i(t)}{4}=\frac34\alpha_i(t)$, which proves the first claim. Next, decompose
\begin{align}
\Delta_t=\big(\hb^{(t)}_r-\nabla_{\bt} V^{\bt^{(t)}}_r(\rho)\big)+\eta\sum_{i=1}^m\Big[\frac{\hb^{(t)}_{g_i}-\nabla_{\bt} V^{\bt^{(t)}}_{g_i}(\rho)}{\underline\alpha_i(t)}+\nabla_{\bt} V^{\bt^{(t)}}_{g_i}(\rho)\Big(\frac{1}{\underline\alpha_i(t)}-\frac{1}{\alpha_i(t)}\Big)\Big].
\end{align}
On $\mathcal E_t$ the first term has norm at most $\wg$, the numerators in the sum have norm at most $\wg$, and $\|\nabla_{\bt} V^{\bt^{(t)}}_{g_i}(\rho)\|\le\Lv$ (Lemma~\ref{lem:smooth}). Using $\underline\alpha_i(t)\ge\frac34\alpha_i(t)$ and $0\le\alpha_i(t)-\underline\alpha_i(t)\le2\wc$,
\begin{align}
\frac{1}{\underline\alpha_i(t)}\le\frac{4}{3\alpha_i(t)}, \qquad
0\le\frac{1}{\underline\alpha_i(t)}-\frac{1}{\alpha_i(t)}=\frac{\alpha_i(t)-\underline\alpha_i(t)}{\alpha_i(t)\underline\alpha_i(t)}\le\frac{2\wc}{\frac34\alpha_i(t)^2}=\frac{8\wc}{3\alpha_i(t)^2},
\end{align}
and the triangle inequality yields \eqref{eq:delta-bound}.

For the second claim, note first that $\bar w_c\le\frac{3c\eta\cdot2\ell}{64m\Lv}\le\frac{c\eta}{8}\cdot\frac{3\ell}{4m\Lv}\le\frac{c\eta}{8}\le\frac{\alpha_{\min}(t)}{8}$, using $\ell\le\Lv$ and $m\ge1$, so the first claim applies. Then, by \eqref{eq:delta-bound}, $\alpha_i(t)\ge c\eta$, $\wg\le\frac{3c\eta}{8(3c+4m)}$ and $\wc\le\frac{3c^2\eta^2}{64m\Lv}$,
\begin{align}
\|\Delta_t\|\le\wg\Big(1+\frac{4m}{3c}\Big)+\frac{8m\eta\Lv\wc}{3c^2\eta^2}
\le\frac{3c\eta}{8(3c+4m)}\cdot\frac{3c+4m}{3c}+\frac{8m\Lv}{3c^2\eta}\cdot\frac{3c^2\eta^2}{64m\Lv}
=\frac{\eta}{8}+\frac{\eta}{8}=\frac{\eta}{4}.  \qquad \qedhere
\end{align}
\end{proof}

%%%%%%%%%%%%%%%%%%%%%%%%%%%%%%%%%%%%%%%%%%%%%%%%%%%%%%%%%%
\section{Safe Exploration (Proof of Theorem~\ref{thm:safe})}
\label{app:safe}

Throughout this appendix we use the elementary facts $\ell\le\Lv$, $\bar c\le\frac16$, $c\le\frac{\bar c}{2}\le\frac{\ell}{12m\Lv}$ (since $\bar c/3<1$ and $m\ge1$), and $\Gbar\ge1$. We fix an iteration $t$ at which Algorithm~\ref{alg_zolbpg} performs an update (i.e., $\|\hg_t\|>\eta/2$), assume that $\bt^{(t)}$ is strictly feasible ($\alpha_{\min}(t)>0$), and work on the good event $\mathcal E_t$ of \eqref{eq:good-event}. We write $u^*_t \DefinedAs \psi_t\|\hg_t\|$ for the length of the update, $\bt(u) \DefinedAs \bt^{(t)}+u\hat u_t$ for $u\ge0$, so that $\bt^{(t+1)}=\bt(u^*_t)$, and we use the following standing conditions on the widths,
\begin{align}
\label{eq:standing}
\wc\le\frac{\alpha_{\min}(t)}{8}, \qquad \wg\le\frac{\Lv}{2}.
\end{align}
By Theorem~\ref{theorem:barrier-err}, \eqref{eq:standing} implies $\frac34\alpha_i(t)\le\underline\alpha_i(t)\le\alpha_i(t)$ for all $i$. Moreover, on $\mathcal E_t$, $\|\hb^{(t)}_\diamond\|\le\Lv+\wg\le\frac32\Lv$ for every $\diamond$, and since $|\langle\hb^{(t)}_{g_i},\hat u_t\rangle-\beta_i(t)|\le\|\hb^{(t)}_{g_i}-\nabla_{\bt}V^{\bt^{(t)}}_{g_i}(\rho)\|\le\wg$ with $|\beta_i(t)|=|\langle\nabla_{\bt} V^{\bt^{(t)}}_{g_i}(\rho),\hat u_t\rangle|\le\Lv$,
\begin{align}
\label{eq:beta-sandwich}
|\beta_i(t)|\le\overline\beta_i(t)\le|\beta_i(t)|+2\wg\le\Lv+2\wg\le2\Lv.
\end{align}

\subsection{No overshoot}

\begin{Lemma}[No overshoot and bounded constraint margin growth]
\label{lem:noovershoot}
Let Assumptions~\ref{assum:link} and~\ref{assum:smooth} hold, and let $a_1,\dots,a_m\ge0$ satisfy $a_i\ge|\beta_i(t)|$. If $0\le\bar u\le\min_i\frac{\underline\alpha_i(t)}{\sqrt{\Mv\underline\alpha_i(t)}+2a_i}$ and $\underline\alpha_i(t)\le\alpha_i(t)$ for all $i$, then for every $i\in[m]$ and every $u\in[0,\bar u]$,
\begin{align}
\alpha_i(t)-\frac{\underline\alpha_i(t)}{2}\ \le\ \bar V^{\bt(u)}_{g_i}(\rho)\ \le\ \alpha_i(t)+\frac{\underline\alpha_i(t)}{2}.
\end{align}
In particular, on $\mathcal E_t$ and under \eqref{eq:standing}, with $a_i=\overline\beta_i(t)$ and $\bar u=u^*_t$ (admissible by \eqref{eq:stepsize} and \eqref{eq:beta-sandwich}), $\frac12\alpha_i(t)\le\alpha_i(t+1)\le\frac32\alpha_i(t)$, so $\bt^{(t+1)}$ is strictly feasible and $\bar V^{\bt(u)}_{g_i}(\rho)\ge\frac12\underline\alpha_i(t)$ along the whole update segment.
\end{Lemma}
\begin{proof}
By the quadratic bound \eqref{eq:quad-bound} (Lemma~\ref{lem:smooth}) applied to $V_{g_i}$ and by $a_i\ge|\beta_i(t)|$,
\begin{align}
\big|\bar V^{\bt(u)}_{g_i}(\rho)-\alpha_i(t)\big|\le u\,|\beta_i(t)|+\frac{\Mv}{2}u^2\le u\,a_i+\frac{\Mv}{2}u^2=:q_i(u).
\end{align}
The function $q_i$ is increasing on $[0,\infty)$, so it suffices to show $q_i(\bar u)\le\frac12\underline\alpha_i(t)$. By assumption,
\begin{align}
\bar u\le\frac{\underline\alpha_i(t)}{\sqrt{\Mv\underline\alpha_i(t)}+2a_i}=\frac{\underline\alpha_i(t)}{s+2a}, \qquad a \DefinedAs a_i,\ s \DefinedAs \sqrt{\Mv\underline\alpha_i(t)}.
\end{align}
Since $q_i$ is increasing and $\Mv\underline\alpha_i(t)=s^2$,
\begin{align}
q_i(\bar u)\le\frac{\underline\alpha_i(t)\,a}{s+2a}+\frac{\Mv\,\underline\alpha_i(t)^2}{2(s+2a)^2}
=\underline\alpha_i(t)\,\frac{2a(s+2a)+s^2}{2(s+2a)^2}
=\underline\alpha_i(t)\,\frac{4a^2+2as+s^2}{2(4a^2+4as+s^2)}\le\frac{\underline\alpha_i(t)}{2},
\end{align}
because $4a^2+2as+s^2\le4a^2+4as+s^2$. The consequence follows from $\underline\alpha_i(t)\le\alpha_i(t)$ (so $\alpha_i(t)\pm\frac12\underline\alpha_i(t)$ lies in $[\frac12\alpha_i(t),\frac32\alpha_i(t)]$) and from $\alpha_i(t)-\frac12\underline\alpha_i(t)\ge\frac12\underline\alpha_i(t)$.

\end{proof}

\subsection{Local smoothness of the barrier along the update}

\begin{Lemma}[Local smoothness]
\label{lem:local}
Let Assumptions~\ref{assum:link} and~\ref{assum:smooth} hold. On $\mathcal E_t$ and under \eqref{eq:standing}, define $\phi(u) \DefinedAs \Bbar^{\bt(u)}(\rho)$ and, for a nonempty $\Iset\subseteq[m]$, $\phi_\Iset(u) \DefinedAs \eta\sum_{i\in\Iset}\log\bar V^{\bt(u)}_{g_i}(\rho)$, $u\in[0,u^*_t]$. Then $\phi$ and $\phi_\Iset$ are twice differentiable on $[0,u^*_t]$ with
\begin{align}
|\phi''(u)|\le\widehat M_t, \qquad |\phi''_\Iset(u)|\le\sum_{i\in\Iset}\Big(\frac{10\Mv\eta}{\underline\alpha_i(t)}+\frac{8\eta\overline\beta_i(t)^2}{\underline\alpha_i(t)^2}\Big)\le\widehat M_t,
\end{align}
where $\widehat M_t$ is defined in \eqref{eq:Mhat}. Consequently,
\begin{align}
\Bbar^{\bt^{(t+1)}}(\rho)&\ge\Bbar^{\bt^{(t)}}(\rho)+\psi_t\big\langle\nabla_{\bt}\Bbar^{\bt^{(t)}}(\rho),\hg_t\big\rangle-\frac{\widehat M_t\psi_t^2}{2}\|\hg_t\|^2, \label{eq:ascent-B}\\
\eta\sum_{i\in\Iset}\log\alpha_i(t+1)&\ge\eta\sum_{i\in\Iset}\log\alpha_i(t)+\psi_t\Big\langle\eta\sum_{i\in\Iset}\frac{\nabla_{\bt} V^{\bt^{(t)}}_{g_i}(\rho)}{\alpha_i(t)},\hg_t\Big\rangle-\frac{\widehat M_t\psi_t^2}{2}\|\hg_t\|^2. \label{eq:ascent-I}
\end{align}
\end{Lemma}
\begin{proof}
By Lemma~\ref{lem:noovershoot}, $\bar V^{\bt(u)}_{g_i}(\rho)\ge\frac12\underline\alpha_i(t)>0$ on $[0,u^*_t]$, so the logarithms are well defined and, by Lemma~\ref{lem:smooth}, $\phi$ and $\phi_\Iset$ are twice differentiable. Differentiating $\log\bar V^{\bt}_{g_i}(\rho)$ twice along $\hat u_t$ gives
\begin{align}
\frac{d^2}{du^2}\log\bar V^{\bt(u)}_{g_i}(\rho)=\frac{\hat u_t^\top\nabla^2_{\bt} V^{\bt(u)}_{g_i}(\rho)\hat u_t}{\bar V^{\bt(u)}_{g_i}(\rho)}-\frac{\langle\nabla_{\bt} V^{\bt(u)}_{g_i}(\rho),\hat u_t\rangle^2}{\bar V^{\bt(u)}_{g_i}(\rho)^2}.
\end{align}
We bound the two terms. First, $|\hat u_t^\top\nabla^2_{\bt} V^{\bt(u)}_{g_i}(\rho)\hat u_t|\le\Mv$ and $\bar V^{\bt(u)}_{g_i}(\rho)\ge\frac12\underline\alpha_i(t)$, so the first term is at most $\frac{2\Mv}{\underline\alpha_i(t)}$ in absolute value. Second, by $\Mv$-smoothness and \eqref{eq:beta-sandwich},
\begin{align}
\big|\langle\nabla_{\bt} V^{\bt(u)}_{g_i}(\rho),\hat u_t\rangle\big|\le\big|\langle\nabla_{\bt} V^{\bt^{(t)}}_{g_i}(\rho),\hat u_t\rangle\big|+\Mv u\le\overline\beta_i(t)+\Mv u^*_t\le\overline\beta_i(t)+\sqrt{\Mv\underline\alpha_i(t)},
\end{align}
where the last step uses $u^*_t\le\underline\alpha_i(t)/\sqrt{\Mv\underline\alpha_i(t)}$, a consequence of \eqref{eq:stepsize}. Hence $\langle\nabla_{\bt} V^{\bt(u)}_{g_i}(\rho),\hat u_t\rangle^2\le2\overline\beta_i(t)^2+2\Mv\underline\alpha_i(t)$, and the second term is at most $\frac{4(2\overline\beta_i(t)^2+2\Mv\underline\alpha_i(t))}{\underline\alpha_i(t)^2}=\frac{8\overline\beta_i(t)^2}{\underline\alpha_i(t)^2}+\frac{8\Mv}{\underline\alpha_i(t)}$ in absolute value. Summing,
\begin{align}
\Big|\frac{d^2}{du^2}\,\eta\log\bar V^{\bt(u)}_{g_i}(\rho)\Big|\le\frac{10\Mv\eta}{\underline\alpha_i(t)}+\frac{8\eta\overline\beta_i(t)^2}{\underline\alpha_i(t)^2},
\end{align}
which gives the bound on $|\phi''_\Iset|$, and adding $|\hat u_t^\top\nabla^2_{\bt} V^{\bt(u)}_r(\rho)\hat u_t|\le\Mv$ gives $|\phi''|\le\widehat M_t$. Finally, Taylor's theorem with the second-derivative bound yields $\phi(u^*_t)\ge\phi(0)+u^*_t\phi'(0)-\frac{\widehat M_t}{2}(u^*_t)^2$ with $\phi'(0)=\langle\nabla_{\bt}\Bbar^{\bt^{(t)}}(\rho),\hat u_t\rangle$ and $u^*_t=\psi_t\|\hg_t\|$, which is \eqref{eq:ascent-B}, and \eqref{eq:ascent-I} follows in the same way from $\phi_\Iset'(0)=\langle\eta\sum_{i\in\Iset}\nabla_{\bt} V^{\bt^{(t)}}_{g_i}(\rho)/\alpha_i(t),\hat u_t\rangle$.

\end{proof}

\subsection{Boundary repulsion through the product of the near-active constraint margins}

\begin{Lemma}[Product of near-active constraint margins is non-decreasing near the boundary]
\label{lem:product}
Let Assumptions~\ref{assum:link},~\ref{assum:smooth} and~\ref{assum:mfcq} hold, let $\eta\le\nu_{\mathrm{emf}}$, and let $\Iset_t \DefinedAs \{i\in[m]:\alpha_i(t)\le\eta\}$. Suppose that on $\mathcal E_t$
\begin{align}
\label{eq:product-widths}
\wg\le\frac{3\ell}{16m}, \qquad \wc\le\frac{3\ell\,\alpha_{\min}(t)}{32\,m\,\Lv}, \qquad\text{and}\qquad \alpha_{\min}(t)\le\bar c\eta=\frac{\ell\eta}{6m\Lv}.
\end{align}
Then for every index set $\Iset$ with $\Iset_t\subseteq\Iset\subseteq[m]$,
\begin{align}
\prod_{i\in\Iset}\alpha_i(t+1)\ \ge\ \prod_{i\in\Iset}\alpha_i(t).
\end{align}
\end{Lemma}
\begin{proof}
The conditions \eqref{eq:product-widths} imply \eqref{eq:standing}, because $\frac{3\ell}{32m\Lv}\le\frac{3}{32}<\frac18$ since $\ell\le\Lv$, and $\frac{3\ell}{16m}\le\frac{3\Lv}{16}<\frac{\Lv}{2}$. Hence Lemmas~\ref{lem:noovershoot} and~\ref{lem:local} apply. Note also that $\Iset_t\neq\emptyset$, because the index attaining $\alpha_{\min}(t)\le\bar c\eta<\eta$ belongs to $\Iset_t$.

\emph{Step 1 (reduction to a comparison of two vectors).} Write $\nabla V_{g_i} \DefinedAs \nabla_{\bt} V^{\bt^{(t)}}_{g_i}(\rho)$ and define
\begin{align}
D_1 \DefinedAs \sum_{i\in\Iset}\frac{\nabla V_{g_i}}{\alpha_i(t)}, \qquad D_2 \DefinedAs \frac{\hg_t}{\eta}-D_1, \qquad\text{so that}\qquad \hg_t=\eta(D_1+D_2).
\end{align}
By \eqref{eq:ascent-I} and $\psi_t\le1/\widehat M_t$ (definition \eqref{eq:stepsize}),
\begin{align}
\eta\sum_{i\in\Iset}\big(\log\alpha_i(t+1)-\log\alpha_i(t)\big)
&\ge\psi_t\langle\eta D_1,\hg_t\rangle-\frac{\psi_t}{2}\|\hg_t\|^2 \notag\\
&=\frac{\psi_t\eta^2}{2}\Big(2\langle D_1,D_1+D_2\rangle-\|D_1+D_2\|^2\Big) \notag\\
&=\frac{\psi_t\eta^2}{2}\big(\|D_1\|^2-\|D_2\|^2\big).
\end{align}
It therefore suffices to prove $\|D_2\|\le\|D_1\|$.

\emph{Step 2 (lower bound on $\|D_1\|$).} Since $\bt^{(t)}$ is strictly feasible and $\eta\le\nu_{\mathrm{emf}}$, every $i\in\Iset_t$ satisfies $0<\alpha_i(t)\le\nu_{\mathrm{emf}}$, i.e., $\Iset_t\subseteq\Iset_{\nu_{\mathrm{emf}}}(\bt^{(t)})$. Let $s \DefinedAs s_{\bt^{(t)}}$ be the unit vector of Assumption~\ref{assum:mfcq}, so that $\langle s,\nabla V_{g_i}\rangle\ge\ell$ for $i\in\Iset_t$, while $|\langle s,\nabla V_{g_i}\rangle|\le\Lv$ and $\alpha_i(t)>\eta$ for $i\in\Iset\setminus\Iset_t$. Hence
\begin{align}
\|D_1\|\ge\langle s,D_1\rangle=\sum_{i\in\Iset_t}\frac{\langle s,\nabla V_{g_i}\rangle}{\alpha_i(t)}+\sum_{i\in\Iset\setminus\Iset_t}\frac{\langle s,\nabla V_{g_i}\rangle}{\alpha_i(t)}
\ge\frac{\ell}{\alpha_{\min}(t)}-\frac{(m-|\Iset_t|)\Lv}{\eta}, \label{eq:D1-lb}
\end{align}
where we kept only the term of the minimizing index in the first sum (all its terms are positive) and used $|\Iset\setminus\Iset_t|\le m-|\Iset_t|$.

\emph{Step 3 (upper bound on $\|D_2\|$).} By the definition \eqref{eq:barrier-grad-hat-new} of $\hg_t$,
\begin{align}
D_2=\frac{\hb^{(t)}_r}{\eta}+\sum_{j\notin\Iset}\frac{\hb^{(t)}_{g_j}}{\underline\alpha_j(t)}+\sum_{j\in\Iset}\Big(\frac{\hb^{(t)}_{g_j}}{\underline\alpha_j(t)}-\frac{\nabla V_{g_j}}{\alpha_j(t)}\Big).
\end{align}
On $\mathcal E_t$ we have $\|\hb^{(t)}_r\|\le\frac32\Lv$. For $j\notin\Iset$ (hence $j\notin\Iset_t$, so $\alpha_j(t)>\eta$) we have $\underline\alpha_j(t)\ge\frac34\alpha_j(t)>\frac34\eta$ and thus $\|\hb^{(t)}_{g_j}\|/\underline\alpha_j(t)\le\frac{3\Lv/2}{3\eta/4}=\frac{2\Lv}{\eta}$. For $j\in\Iset$, exactly as in the proof of Theorem~\ref{theorem:barrier-err},
\begin{align}
\Big\|\frac{\hb^{(t)}_{g_j}}{\underline\alpha_j(t)}-\frac{\nabla V_{g_j}}{\alpha_j(t)}\Big\|&\le\frac{\wg}{\underline\alpha_j(t)}+\Lv\Big(\frac{1}{\underline\alpha_j(t)}-\frac{1}{\alpha_j(t)}\Big) \notag\\
&\le\frac{4\wg}{3\alpha_j(t)}+\frac{8\Lv\wc}{3\alpha_j(t)^2}\le\frac{4\wg}{3\alpha_{\min}(t)}+\frac{8\Lv\wc}{3\alpha_{\min}(t)^2}.
\end{align}
Since $|\Iset|\le m$, the width conditions in \eqref{eq:product-widths} give
\begin{align}
\sum_{j\in\Iset}\Big(\frac{4\wg}{3\alpha_{\min}(t)}+\frac{8\Lv\wc}{3\alpha_{\min}(t)^2}\Big)&\le\frac{4m}{3\alpha_{\min}(t)}\cdot\frac{3\ell}{16m}+\frac{8m\Lv}{3\alpha_{\min}(t)^2}\cdot\frac{3\ell\alpha_{\min}(t)}{32m\Lv} \notag\\
&=\frac{\ell}{4\alpha_{\min}(t)}+\frac{\ell}{4\alpha_{\min}(t)}=\frac{\ell}{2\alpha_{\min}(t)}.
\end{align}
Therefore
\begin{align}
\|D_2\|\le\frac{3\Lv}{2\eta}+\frac{2(m-|\Iset|)\Lv}{\eta}+\frac{\ell}{2\alpha_{\min}(t)}. \label{eq:D2-ub}
\end{align}

\emph{Step 4 (comparison).} Combining \eqref{eq:D1-lb} and \eqref{eq:D2-ub}, and using $|\Iset|\ge|\Iset_t|\ge1$,
\begin{align}
\|D_1\|-\|D_2\|&\ge\frac{\ell}{2\alpha_{\min}(t)}-\frac{\Lv}{\eta}\Big(\frac32+2(m-|\Iset|)+(m-|\Iset_t|)\Big) \notag\\
&\ge\frac{\ell}{2\alpha_{\min}(t)}-\frac{\Lv}{\eta}\Big(3m-\frac32\Big)\ge\frac{\ell}{2\alpha_{\min}(t)}-\frac{3m\Lv}{\eta}.
\end{align}
Finally, $\alpha_{\min}(t)\le\bar c\eta=\frac{\ell\eta}{6m\Lv}$ is equivalent to $\frac{3m\Lv}{\eta}\le\frac{\ell}{2\alpha_{\min}(t)}$, so $\|D_1\|\ge\|D_2\|$. By Step 1, $\sum_{i\in\Iset}\log\alpha_i(t+1)\ge\sum_{i\in\Iset}\log\alpha_i(t)$, which is the claim (all constraint margins involved are positive by Lemma~\ref{lem:noovershoot}).

\end{proof}

\subsection{Lower bound on the step size}

The constant $C$ of Theorem~\ref{thm:safe} is
\begin{align}
\label{eq:Cconst}
C \DefinedAs \min\Big\{\Big[\Mv\Big(1+\frac{40m}{3c}\Big)+\frac{512m\Lv^2}{9c^2}\Big]^{-1},\ \frac{c}{2\Lv\big(1+\frac{2m}{c}\big)\big(\sqrt{\Mv\Gbar}+4\Lv\big)}\Big\},
\end{align}
and its two terms correspond to the two terms of the step size \eqref{eq:stepsize}.

\begin{Lemma}[Step-size lower bound]
\label{lem:steplb}
Let Assumptions~\ref{assum:link} and~\ref{assum:smooth} hold and $\eta\le1$. On $\mathcal E_t$, if $\alpha_{\min}(t)\ge c\eta$ and \eqref{eq:standing} holds, then $\psi_t\ge C\eta$ with $C$ as in \eqref{eq:Cconst}.
\end{Lemma}
\begin{proof}
On $\mathcal E_t$ under \eqref{eq:standing}, $\frac34c\eta\le\frac34\alpha_i(t)\le\underline\alpha_i(t)\le\alpha_i(t)\le\Gbar$ and $\overline\beta_i(t)\le2\Lv$ by \eqref{eq:beta-sandwich}. Hence, using $\eta\le1$ in the last step,
\begin{align}
\widehat M_t\le\Mv+\sum_{i=1}^m\Big(\frac{10\Mv\eta}{\frac34c\eta}+\frac{8\eta(2\Lv)^2}{(\frac34c\eta)^2}\Big)&=\Mv\Big(1+\frac{40m}{3c}\Big)+\frac{512m\Lv^2}{9c^2\eta} \notag\\
&\le\frac1\eta\Big[\Mv\Big(1+\frac{40m}{3c}\Big)+\frac{512m\Lv^2}{9c^2}\Big],
\end{align}
so the first term in \eqref{eq:stepsize} is at least $C_{\mathrm a}\eta$ with $C_{\mathrm a} \DefinedAs [\Mv(1+\frac{40m}{3c})+\frac{512m\Lv^2}{9c^2}]^{-1}$. For the second term, $\|\hg_t\|\le\|\hb^{(t)}_r\|+\eta\sum_i\|\hb^{(t)}_{g_i}\|/\underline\alpha_i(t)\le\frac32\Lv+\eta m\cdot\frac{3\Lv/2}{3c\eta/4}=\frac32\Lv\big(1+\frac{2m}{c}\big)$. The map $x\mapsto\frac{x}{\sqrt{\Mv x}+2\overline\beta_i(t)}$ is increasing on $(0,\infty)$, so, using $\underline\alpha_i(t)\ge\frac34c\eta$, $\overline\beta_i(t)\le2\Lv$ and $\frac34c\eta\le\Gbar$,
\begin{align}
\frac{\underline\alpha_i(t)}{\sqrt{\Mv\underline\alpha_i(t)}+2\overline\beta_i(t)}\ge\frac{\frac34c\eta}{\sqrt{\Mv\cdot\frac34c\eta}+4\Lv}\ge\frac{3c\eta}{4\big(\sqrt{\Mv\Gbar}+4\Lv\big)}.
\end{align}
Dividing by $\|\hg_t\|\le\frac32\Lv(1+\frac{2m}{c})$ shows that the second term in \eqref{eq:stepsize} is at least $C_{\mathrm b}\eta$ with $C_{\mathrm b} \DefinedAs \frac{c}{2\Lv(1+\frac{2m}{c})(\sqrt{\Mv\Gbar}+4\Lv)}$. Thus $\psi_t\ge\min\{C_{\mathrm a},C_{\mathrm b}\}\eta=C\eta$ by \eqref{eq:Cconst}.

\end{proof}

\subsection{Proof of Theorem~\ref{thm:safe}}

\begin{proof}
Let $T'\le T$ denote the number of iterations at which Algorithm~\ref{alg_zolbpg} forms estimates (the algorithm either updates at every $t<T$, in which case $T'=T$, or stops at the first $t_b<T$ with $\|\hg_{t_b}\|\le\eta/2$, in which case $T'=t_b+1$). Let $\mathcal E \DefinedAs \bigcap_{t<T'}\mathcal E_t$. Conditionally on the history up to iteration $t$, the estimators at $\bt^{(t)}$ are formed from fresh samples, so Lemma~\ref{lem:est} gives $\PP(\mathcal E_t^c \,\vert\, \text{history})\le\delta$, hence $\PP(\mathcal E_t^c)\le\delta$ and, by the union bound, $\PP(\mathcal E)\ge1-T\delta$. We show that all four claims hold deterministically on $\mathcal E$.

\emph{Preliminaries on the widths.} By assumption, \eqref{eq:range} holds (so that Lemma~\ref{lem:est} applies at every iteration), and $\wg\le\bar w_g$, $\wc\le\bar w_c$ with $\bar w_g,\bar w_c$ from \eqref{eq:width-req}. We record the consequences of the width bounds,
\begin{align}
\wg&\le\frac{3\cdot2\ell}{8(3c+4m)}\le\frac{3\ell}{16m}\le\frac{\Lv}{2}, \label{eq:wg-cons}\\
\wc&\le\frac{3c\eta\cdot2\ell}{64m\Lv}=\frac{3\ell\,c\eta}{32m\Lv}\le\frac{c\eta}{8}. \label{eq:wc-cons}
\end{align}
In particular, whenever $\alpha_{\min}(t)\ge c\eta$, the standing conditions \eqref{eq:standing} hold and $\wc\le\frac{3\ell\alpha_{\min}(t)}{32m\Lv}$, so that Lemmas~\ref{lem:noovershoot}, \ref{lem:local}, \ref{lem:steplb} and (if additionally $\alpha_{\min}(t)\le\bar c\eta$) Lemma~\ref{lem:product} are applicable at iteration $t$.

\emph{Claim (i), $\alpha_{\min}(t)\ge c\eta$ for all $t\le T'$.} We argue by induction on $t$. For $t=0$, Assumption~\ref{assum:slater} gives $\alpha_i(0)\ge\nu\ge\nu_{\mathrm{emf}}\ge\eta\ge c\eta$ because $c<1$. Suppose that $\alpha_{\min}(u)\ge c\eta$ for all $u\le t$, where $t<T'$ is an iteration at which an update is performed (if no update is performed at $t$, there is nothing to prove). Then the lemmas above apply at every $u\le t$.

\emph{Case A, $\alpha_{\min}(t)>\bar c\eta$.} By Lemma~\ref{lem:noovershoot}, $\alpha_i(t+1)\ge\frac12\alpha_i(t)>\frac{\bar c\eta}{2}\ge c\eta$ for all $i$, since $c\le\bar c/2$.

\emph{Case B, $\alpha_{\min}(t)\le\bar c\eta$.} Let $t_0\le t$ be the smallest index such that $\alpha_{\min}(u)\le\bar c\eta$ for every $u\in\{t_0,\dots,t\}$ (the current ``danger phase''). Since $\alpha_{\min}(0)\ge\eta>\bar c\eta$, we have $t_0\ge1$ and $\alpha_{\min}(t_0-1)>\bar c\eta$, and Lemma~\ref{lem:noovershoot} at iteration $t_0-1$ then yields
\begin{align}
\alpha_i(t_0)\ge\frac12\alpha_i(t_0-1)>\frac{\bar c\eta}{2}\quad\text{for all } i\in[m]. \label{eq:phase-start}
\end{align}
For $v\in\{t_0,\dots,t+1\}$ let $\Iset_v \DefinedAs \{i:\alpha_i(v)\le\eta\}$ be the near-active set at iteration $v$ (for $v=t+1$ this refers to the new iterate), and for $u\in\{t_0,\dots,t+1\}$ define the nested sets
\begin{align}
\Iset^{(u)} \DefinedAs \bigcup_{v=u}^{t+1}\Iset_v, \qquad\text{so that}\qquad \Iset^{(u)}=\Iset_u\cup\Iset^{(u+1)}\supseteq\Iset_u \quad\text{and}\quad \Iset^{(t+1)}=\Iset_{t+1}.
\end{align}
For every $u\in\{t_0,\dots,t\}$, the hypotheses of Lemma~\ref{lem:product} hold at iteration $u$ (we are on $\mathcal E_u$, $\alpha_{\min}(u)\ge c\eta$ by the induction hypothesis, and $\alpha_{\min}(u)\le\bar c\eta$ by the definition of the phase), and $\Iset^{(u)}\supseteq\Iset_u$. Hence
\begin{align}
\prod_{i\in\Iset^{(u)}}\alpha_i(u+1)\ge\prod_{i\in\Iset^{(u)}}\alpha_i(u). \label{eq:prod-step}
\end{align}
Every $i\in\Iset^{(u)}\setminus\Iset^{(u+1)}$ belongs to $\Iset_u$, so $\alpha_i(u)\le\eta$ and, by Lemma~\ref{lem:noovershoot}, $\alpha_i(u+1)\le\frac32\alpha_i(u)\le\frac32\eta$. Dividing \eqref{eq:prod-step} by $\prod_{i\in\Iset^{(u)}\setminus\Iset^{(u+1)}}\alpha_i(u+1)\le(\frac32\eta)^{|\Iset^{(u)}\setminus\Iset^{(u+1)}|}$ gives
\begin{align}
\prod_{i\in\Iset^{(u+1)}}\alpha_i(u+1)\ge\frac{\prod_{i\in\Iset^{(u)}}\alpha_i(u)}{(\frac32\eta)^{|\Iset^{(u)}\setminus\Iset^{(u+1)}|}}, \qquad u=t_0,\dots,t.
\end{align}
Chaining these inequalities from $u=t_0$ to $u=t$, and noting that the sets $\Iset^{(u)}\setminus\Iset^{(u+1)}$, $u=t_0,\dots,t$, are pairwise disjoint with union $\Iset^{(t_0)}\setminus\Iset^{(t+1)}$, we obtain
\begin{align}
\prod_{i\in\Iset_{t+1}}\alpha_i(t+1)\ge\frac{\prod_{i\in\Iset^{(t_0)}}\alpha_i(t_0)}{(\frac32\eta)^{|\Iset^{(t_0)}\setminus\Iset_{t+1}|}}. \label{eq:prod-chain}
\end{align}
Now fix $j\in\Iset_{t+1}$ (if $\Iset_{t+1}=\emptyset$ there is nothing to prove) and let $k \DefinedAs |\Iset^{(t_0)}|\in\{1,\dots,m\}$. Dividing \eqref{eq:prod-chain} by $\prod_{i\in\Iset_{t+1}\setminus\{j\}}\alpha_i(t+1)\le\eta^{|\Iset_{t+1}|-1}\le(\frac32\eta)^{|\Iset_{t+1}|-1}$ and using \eqref{eq:phase-start},
\begin{align}
\alpha_j(t+1)\ge\frac{\prod_{i\in\Iset^{(t_0)}}\alpha_i(t_0)}{(\frac32\eta)^{|\Iset^{(t_0)}|-1}}\ge\frac{(\bar c\eta/2)^{k}}{(\frac32\eta)^{k-1}}=\frac{\bar c\eta}{2}\Big(\frac{\bar c}{3}\Big)^{k-1}\ge\frac{\bar c\eta}{2}\Big(\frac{\bar c}{3}\Big)^{m-1}=c\eta,
\end{align}
where the last inequality uses $\bar c/3<1$ and $k\le m$. For $j\notin\Iset_{t+1}$ we have $\alpha_j(t+1)>\eta\ge c\eta$ directly. This completes the induction and proves (i) for all iterates generated by the algorithm.

\emph{Claim (ii), safety of the exploration policies.} For $t<T'$, $i\in[m]$ and $n\in[N]$, Lemma~\ref{lem:smooth} ($\Lv$-Lipschitz continuity of $V_{g_i}$) and $\|\mu\bv_n^{(t)}\|=\mu\le\frac{c\eta}{2\Lv}$ give
\begin{align}
\bar V^{\bt^{(t)}+\mu\bv_n^{(t)}}_{g_i}(\rho)\ge\alpha_i(t)-\mu\Lv\ge c\eta-\frac{c\eta}{2}=\frac{c\eta}{2}.
\end{align}

\emph{Claim (iii), barrier-gradient error.} For $t<T'$, by (i), \eqref{eq:width-req} and Theorem~\ref{theorem:barrier-err}, $\|\Delta_t\|\le\eta/4$.

\emph{Claim (iv), step size.} For every $t<T'$ at which an update is performed, (i), \eqref{eq:wg-cons}, \eqref{eq:wc-cons} and Lemma~\ref{lem:steplb} give $\psi_t\ge C\eta$.

Finally, (i) and (ii) show that every policy executed by ZO-LBPG is strictly feasible on $\mathcal E$, i.e., safe exploration holds with probability at least $1-T\delta$.

\end{proof}

%%%%%%%%%%%%%%%%%%%%%%%%%%%%%%%%%%%%%%%%%%%%%%%%%%%%%%%%%%
\section{Global Convergence and Sample Complexity}
\label{app:gd}

This appendix proves the results of Section~\ref{sec:convergence}. We recall the performance difference lemma, prove gradient dominance (Lemma~\ref{lem:gd}), derive the linear contraction of the optimality gap (Theorem~\ref{thm:main}) and translate the accuracy requirements into sample sizes (Corollary~\ref{cor:samples}).

\subsection{Performance difference lemma}

\begin{Lemma}[Performance difference \citep{kakade2002approximately}]
\label{lem:pdl}
For any policies $\pi,\pi'$ and $\diamond \in \{r\} \cup \{g_i\}_{i=1}^m$,
\begin{align}
V_\diamond^{\pi'}(\rho) - V_\diamond^{\pi}(\rho) = \frac{1}{1-\gamma}\E_{(s,a) \sim d^{\pi}_\rho}\!\big[A_\diamond^{\pi'}(s,a)\big],
\end{align}
where $A_\diamond^{\pi'}(s,a) = Q_\diamond^{\pi'}(s,a) - V_\diamond^{\pi'}(s)$ is the advantage function of $\pi'$.
\end{Lemma}

\subsection{Proof of Lemma~\ref{lem:gd} (gradient dominance)}

\begin{proof}
Fix a strictly feasible $\bt$ and write $F \DefinedAs F^{\bt}(\rho)$, $\alpha_i \DefinedAs \bar V^{\bt}_{g_i}(\rho)>0$, $\nabla V_\diamond \DefinedAs \nabla_{\bt} V^{\bt}_\diamond(\rho)$ and $\nabla B \DefinedAs \nabla_{\bt}\Bbar^{\bt}(\rho)=\nabla V_r+\sum_i\frac{\eta}{\alpha_i}\nabla V_{g_i}$.

\emph{Step 1 (performance-difference expansion).} Applying Lemma~\ref{lem:pdl} with $\pi=\pi^*$ and $\pi'=\pi_{\bt}$ to $r$ and to each $g_i$,
\begin{align}
V_r^{\pi^*}(\rho) - V_r^{\bt}(\rho) + \sum_{i=1}^m \frac{\eta}{\alpha_i}\big(V_{g_i}^{\pi^*}(\rho) - V_{g_i}^{\bt}(\rho)\big)
= \frac{1}{1-\gamma}\E_{(s,a) \sim d^{\pi^*}_\rho}\Big[A_r^{\bt}(s,a) + \sum_{i=1}^m \frac{\eta}{\alpha_i}A_{g_i}^{\bt}(s,a)\Big]. \label{eq:gd1-app}
\end{align}

\emph{Step 2 (approximation by score functions).} By Assumption~\ref{assum:transfer} and Jensen's inequality, for each $\diamond$,
\begin{align}
\E_{d^{\pi^*}_\rho}\!\big[A_\diamond^{\bt}(s,a) - (1-\gamma){\mu_\diamond^*}^\top\nabla_{\bt}\log\pi_{\bt}(a|s)\big]&\le\E_{d^{\pi^*}_\rho}\!\big|A_\diamond^{\bt}(s,a) - (1-\gamma){\mu_\diamond^*}^\top\nabla_{\bt}\log\pi_{\bt}(a|s)\big| \notag\\
&\le \sqrt{\varepsilon_{\mathrm{bias}}}.
\end{align}
Adding and subtracting $(1-\gamma){\mu_\diamond^*}^\top\nabla_{\bt}\log\pi_{\bt}$ inside \eqref{eq:gd1-app} and using that all coefficients $\eta/\alpha_i$ are positive,
\begin{align}
&V_r^{\pi^*}(\rho) - V_r^{\bt}(\rho) + \sum_{i=1}^m \frac{\eta}{\alpha_i}\big(V_{g_i}^{\pi^*}(\rho) - V_{g_i}^{\bt}(\rho)\big)
\notag\\
&\qquad\le \E_{d^{\pi^*}_\rho}\!\Big[\Big(\mu_r^* + \sum_{i=1}^m \frac{\eta}{\alpha_i}\mu_{g_i}^*\Big)^\top \nabla_{\bt}\log\pi_{\bt}(a|s)\Big] + \frac{\sqrt{\varepsilon_{\mathrm{bias}}}}{1-\gamma}\Big(1 + \sum_{i=1}^m \frac{\eta}{\alpha_i}\Big). \label{eq:gd2-app}
\end{align}

\emph{Step 3 (identification of the barrier gradient).} Since $\mu_\diamond^* = F^{\dagger} \nabla V_\diamond$ and $F^\dagger$ is linear,
\begin{align}
\mu_r^* + \sum_{i=1}^m \frac{\eta}{\alpha_i}\mu_{g_i}^* = F^{\dagger}\Big(\nabla V_r+\sum_{i=1}^m\frac{\eta}{\alpha_i}\nabla V_{g_i}\Big)=F^{\dagger} \nabla B.
\end{align}

\emph{Step 4 (Fisher non-degeneracy).} Let $P_{\mathrm{Im}}$ denote the orthogonal projection onto $\mathrm{Im}(F)$. Since $F$ is symmetric positive semidefinite, $F^\dagger=F^\dagger P_{\mathrm{Im}}$ and the nonzero eigenvalues of $F^\dagger$ are the reciprocals of the nonzero eigenvalues of $F$, which are at least $\mu_F$ by Assumption~\ref{assum:fisher}. Hence $\|F^\dagger\nabla B\|=\|F^\dagger P_{\mathrm{Im}}\nabla B\|\le\frac{1}{\mu_F}\|P_{\mathrm{Im}}\nabla B\|\le\frac{\|\nabla B\|}{\mu_F}$. Together with $\|\nabla_{\bt}\log\pi_{\bt}(a|s)\|\le M_g$ (Assumption~\ref{assum:smooth}) and the Cauchy--Schwarz inequality,
\begin{align}
\E_{d^{\pi^*}_\rho}\!\Big[\big(F^{\dagger}\nabla B\big)^\top\nabla_{\bt}\log\pi_{\bt}(a|s)\Big]\le\|F^\dagger\nabla B\|\,\E_{d^{\pi^*}_\rho}\|\nabla_{\bt}\log\pi_{\bt}(a|s)\|\le\frac{M_g}{\mu_F}\|\nabla B\|.
\end{align}

\emph{Step 5 (feasibility of $\pi^*$).} Since $\pi^*$ is feasible, $V_{g_i}^{\pi^*}(\rho) \ge b_i$, hence $\frac{V_{g_i}^{\pi^*}(\rho) - V_{g_i}^{\bt}(\rho)}{\alpha_i}=\frac{V_{g_i}^{\pi^*}(\rho) - b_i}{\alpha_i} - 1 \ge -1$ and $\sum_i\frac{\eta}{\alpha_i}(V_{g_i}^{\pi^*}(\rho) - V_{g_i}^{\bt}(\rho))\ge-m\eta$.

\emph{Step 6 (conclusion).} Combining \eqref{eq:gd2-app} with Steps 3, 4 and 5,
\begin{align}
V_r^{\pi^*}(\rho) - V_r^{\bt}(\rho) - m\eta \le \frac{M_g}{\mu_F}\|\nabla_{\bt}\Bbar^{\bt}(\rho)\| + \frac{\sqrt{\varepsilon_{\mathrm{bias}}}}{1-\gamma}\Big(1 + \sum_{i=1}^m \frac{\eta}{\bar V_{g_i}^{\bt}(\rho)}\Big),
\end{align}
which is the claim.

\end{proof}

\subsection{Proof of Theorem~\ref{thm:main} (global convergence)}

\begin{proof}
Let $T'$ and $\mathcal E=\bigcap_{t<T'}\mathcal E_t$ be as in the proof of Theorem~\ref{thm:safe}, so that $\PP(\mathcal E)\ge1-T\delta$, and work on $\mathcal E$. By Theorem~\ref{thm:safe}, for every $t<T'$ we have $\alpha_{\min}(t)\ge c\eta$ and $\|\Delta_t\|\le\eta/4$, and, whenever an update is performed, $\psi_t\ge C\eta\ge\bar C\eta$ and the ascent inequality \eqref{eq:ascent-B} holds. Write $B_t \DefinedAs \Bbar^{\bt^{(t)}}(\rho)$, $\nabla B_t \DefinedAs \nabla_{\bt}\Bbar^{\bt^{(t)}}(\rho)$, $D_t \DefinedAs V_r^{\pi^*}(\rho)-V_r^{\bt^{(t)}}(\rho)$ and
\begin{align}
a \DefinedAs m\eta+\frac{\sqrt{\varepsilon_{\mathrm{bias}}}}{1-\gamma}\Big(1+\frac mc\Big).
\end{align}
Since $\eta/\alpha_i(t)\le1/c$, Lemma~\ref{lem:gd} yields, for every $t<T'$,
\begin{align}
D_t\le a+\frac{M_g}{\mu_F}\|\nabla B_t\|. \label{eq:gd-used}
\end{align}

\emph{Step 1 (one-step ascent).} Suppose an update is performed at iteration $t$, i.e., $\|\hg_t\|>\eta/2$. From \eqref{eq:ascent-B}, $\psi_t\le1/\widehat M_t$ and $\hg_t=\nabla B_t+\Delta_t$,
\begin{align}
B_{t+1}-B_t\ge\psi_t\langle\nabla B_t,\hg_t\rangle-\frac{\psi_t}{2}\|\hg_t\|^2&=\frac{\psi_t}{2}\Big(2\langle\nabla B_t,\nabla B_t+\Delta_t\rangle-\|\nabla B_t+\Delta_t\|^2\Big) \notag\\
&=\frac{\psi_t}{2}\big(\|\nabla B_t\|^2-\|\Delta_t\|^2\big).
\end{align}
Since $\|\nabla B_t\|\ge\|\hg_t\|-\|\Delta_t\|>\frac\eta2-\frac\eta4=\frac\eta4$, we have $\|\Delta_t\|^2\le\frac{\eta^2}{16}\le\frac\eta4\|\nabla B_t\|$ and $\|\nabla B_t\|^2-\|\Delta_t\|^2\ge\frac{\eta}{4}\|\nabla B_t\|-\frac{\eta^2}{16}\ge0$. Using $\psi_t\ge\bar C\eta$ and then \eqref{eq:gd-used},
\begin{align}
B_{t+1}-B_t\ge\frac{\bar C\eta}{2}\Big(\frac{\eta}{4}\|\nabla B_t\|-\frac{\eta^2}{16}\Big)=\frac{\bar C\eta^2}{8}\|\nabla B_t\|-\frac{\bar C\eta^3}{32}\ge\kappa\,(D_t-a)-\frac{\bar C\eta^3}{32}, \qquad \kappa=\frac{\bar C\mu_F\eta^2}{8M_g}. \label{eq:ascent-used}
\end{align}
On the other hand, by the definition of the barrier,
\begin{align}
B_{t+1}-B_t=\big(V_r^{\bt^{(t+1)}}(\rho)-V_r^{\bt^{(t)}}(\rho)\big)+\eta\sum_{i=1}^m\Lambda_i(t)=D_t-D_{t+1}+\eta\sum_{i=1}^m\Lambda_i(t), \notag\\
\Lambda_i(t) \DefinedAs \log\alpha_i(t+1)-\log\alpha_i(t).
\end{align}
Combining with \eqref{eq:ascent-used},
\begin{align}
D_{t+1}\le(1-\kappa)D_t+\kappa a+\frac{\bar C\eta^3}{32}+\eta\sum_{i=1}^m\Lambda_i(t). \label{eq:recursion}
\end{align}

\emph{Step 2 (no early stop).} Suppose the algorithm performs all $T$ updates ($T'=T$, $\bt_{\mathrm{out}}=\bt^{(T)}$). Since $\kappa\in(0,1]$, unrolling \eqref{eq:recursion} from $t=0$ to $T-1$ gives
\begin{align}
D_T\le(1-\kappa)^TD_0+\Big(\kappa a+\frac{\bar C\eta^3}{32}\Big)\sum_{j=0}^{T-1}(1-\kappa)^{T-1-j}+\eta\sum_{i=1}^m S_i, \notag\\
S_i \DefinedAs \sum_{j=0}^{T-1}(1-\kappa)^{T-1-j}\Lambda_i(j).
\end{align}
The geometric sum is at most $1/\kappa$, so the middle term is at most $a+\frac{\bar C\eta^3}{32\kappa}=a+\frac{M_g\eta}{4\mu_F}$. For the weighted telescoping sums $S_i$, write $\ell_j \DefinedAs \log\alpha_i(j)$ and $w_j \DefinedAs (1-\kappa)^{T-1-j}$ for $0\le j\le T-1$. Then $w_{T-1}=1$ and $w_{j-1}-w_j=-\kappa(1-\kappa)^{T-1-j}\le0$ for $1\le j\le T-1$, and Abel summation gives
\begin{align}
S_i=\sum_{j=0}^{T-1}w_j(\ell_{j+1}-\ell_j)&=w_{T-1}\ell_T-w_0\ell_0+\sum_{j=1}^{T-1}(w_{j-1}-w_j)\ell_j \notag\\
&=\ell_T-(1-\kappa)^{T-1}\ell_0-\kappa\sum_{j=1}^{T-1}(1-\kappa)^{T-1-j}\ell_j.
\end{align}
Now $\ell_T=\log\alpha_i(T)\le\log\Gbar$ because $\alpha_i(T)\le V^{\bt^{(T)}}_{g_i}(\rho)\le\Gbar$, and $\ell_j\ge\log(c\eta)$ for $0\le j\le T-1$ by Theorem~\ref{thm:safe}. Since the coefficients $(1-\kappa)^{T-1}$ and $\kappa(1-\kappa)^{T-1-j}$ are nonnegative and sum to $(1-\kappa)^{T-1}+\kappa\sum_{k=0}^{T-2}(1-\kappa)^k=(1-\kappa)^{T-1}+1-(1-\kappa)^{T-1}=1$,
\begin{align}
S_i\le\log\Gbar-\log(c\eta)=\log\frac{1}{c\eta(1-\gamma)}.
\end{align}
Therefore
\begin{align}
D_T\le(1-\kappa)^TD_0+a+\frac{M_g\eta}{4\mu_F}+m\eta\log\frac{1}{c\eta(1-\gamma)}. \label{eq:caseA}
\end{align}

\emph{Step 3 (early stop).} Suppose instead that the algorithm stops at iteration $t_b<T$, returning $\bt_{\mathrm{out}}=\bt^{(t_b)}$, because $\|\hg_{t_b}\|\le\eta/2$. Since $\mathcal E_{t_b}$ holds (the estimates at $t_b$ are formed before the stopping test, so $t_b<T'$), Theorem~\ref{thm:safe}(iii) gives $\|\Delta_{t_b}\|\le\eta/4$ and thus $\|\nabla B_{t_b}\|\le\|\hg_{t_b}\|+\|\Delta_{t_b}\|\le\frac{3\eta}{4}$. By \eqref{eq:gd-used},
\begin{align}
D_{t_b}\le a+\frac{3M_g\eta}{4\mu_F}. \label{eq:caseB}
\end{align}

\emph{Step 4 (conclusion).} In both cases, \eqref{eq:caseA} and \eqref{eq:caseB} imply
\begin{align}
V_r^{\pi^*}(\rho)-V_r^{\bt_{\mathrm{out}}}(\rho)\le(1-\kappa)^TD_0+a+\frac{3M_g\eta}{4\mu_F}+m\eta\log\frac{1}{c\eta(1-\gamma)},
\end{align}
and substituting $a$ yields the statement of the theorem (note that $c\eta(1-\gamma)<1$, so the logarithm is positive).

\end{proof}

\subsection{Proof of Corollary~\ref{cor:samples} (sample complexity)}

The hyperparameters of Corollary~\ref{cor:samples} are, with $\eta=\varepsilon/m$ and $\delta=\delta_0/T$,
\begin{align}
\label{eq:param-choice}
T=\Big\lceil\frac{8M_gm^2}{\bar C\mu_F\varepsilon^2}\log\frac{\Gbar}{\varepsilon}\Big\rceil, \quad
\mu=\min\Big\{\frac{\bar w_g}{2d\Mv},\frac{c\eta}{2\Lv}\Big\}, \quad
M_c=\Big\lceil\frac{3L_2}{8\bar w_c}\Big\rceil, \quad
M=\Big\lceil\frac{dL_2}{2\mu\bar w_g}\Big\rceil, \notag\\
N_c=\Big\lceil\frac{18(\Gbar+L)^2\iot}{\bar w_c^2}\Big\rceil, \quad
N=\Big\lceil\max\Big\{\frac{128\sigma_g^2\iot}{\bar w_g^2},\ \frac{16(\Gbar+L)^2\iot}{\Gbar^2H|\Ac|M_g\,\mu}\Big\}\Big\rceil,
\end{align}
and $H$ is the smallest integer with $\epsilon_V(H)\le\bar w_c/3$ and $\epsilon_\nabla(H)\le\bar w_g/4$. The proof verifies the hypotheses of Theorems~\ref{thm:safe} and~\ref{thm:main} for these choices, derives the optimality gap, and then computes the orders of magnitude, including the dependence on $m$, $c$, $\bar C$ and $\mu_F$ that is hidden in the statement of the corollary.

\begin{proof}
\emph{Width requirements.} With $\eta=\varepsilon/m\le\min\{1,\nu_{\mathrm{emf}}\}$ (as $\varepsilon\le1$) and $\eta\le2\ell/c$, the widths \eqref{eq:width-req} read $\bar w_g=\frac{3c\eta}{8(3c+4m)}$ and $\bar w_c=\frac{3c^2\eta^2}{64m\Lv}$. We verify the range condition \eqref{eq:range}, $\wc\le\bar w_c$, $\wg\le\bar w_g$ and $\mu\le c\eta/(2\Lv)$, the hypotheses of Theorem~\ref{thm:safe} and Theorem~\ref{thm:main}.

\emph{Range condition.} Since $\mu^2\sigma_g^2=2d^2(\Gbar^2H|\Ac|M_g\mu+\frac{L^2}{4M})\ge2d^2\Gbar^2H|\Ac|M_g\mu$, condition \eqref{eq:range} is implied by $N\ge\frac{16(\Gbar+L)^2\iot}{\Gbar^2H|\Ac|M_g\mu}$, which the choice of $N$ in \eqref{eq:param-choice} ensures.

\emph{Constraint margins.} $\epsilon_V(H)\le\bar w_c/3$ by the choice of $H$, $\beta_{M_c}=\frac{L_2}{8M_c}\le\frac{\bar w_c}{3}$ by the choice of $M_c$, and $s_c\sqrt{\iot}=(\Gbar+L)\sqrt{2\iot/N_c}\le\frac{\sqrt2\bar w_c}{\sqrt{18}}=\frac{\bar w_c}{3}$ by the choice of $N_c$. Hence $\wc\le\bar w_c$.

\emph{Gradients.} $\epsilon_\nabla(H)\le\bar w_g/4$ by the choice of $H$, $\frac{d\Mv\mu}{2}\le\frac{\bar w_g}{4}$ and $\mu\le\frac{c\eta}{2\Lv}$ by the choice of $\mu$, $\frac{d\beta_M}{\mu}=\frac{dL_2}{8M\mu}\le\frac{\bar w_g}{4}$ since $M\ge\frac{dL_2}{2\mu\bar w_g}$, and $s_g\sqrt{\iot}=\sqrt{8\sigma_g^2\iot/N}\le\sqrt{8\bar w_g^2/128}=\frac{\bar w_g}{4}$ since $N\ge\frac{128\sigma_g^2\iot}{\bar w_g^2}$. Hence $\wg\le\bar w_g$.

With $\delta=\delta_0/T$ the failure probability is at most $T\delta=\delta_0$.

\emph{Optimality gap.} With $\kappa=\frac{\bar C\mu_F\varepsilon^2}{8M_gm^2}$ and $T\ge\frac1\kappa\log\frac{\Gbar}{\varepsilon}$, $(1-\kappa)^TD_0\le e^{-\kappa T}\Gbar\le\varepsilon$, using $D_0\le V_r^{\pi^*}(\rho)\le\Gbar$. Theorem~\ref{thm:main} with $\eta=\varepsilon/m$ and $1+\frac mc\le\frac{2m}{c}$ (as $c<1\le m$) gives
\begin{align}
V_r^{\pi^*}(\rho)-V_r^{\bt_{\mathrm{out}}}(\rho)\le\varepsilon+\frac{2m\sqrt{\varepsilon_{\mathrm{bias}}}}{c(1-\gamma)}+\frac{3M_g\varepsilon}{4m\mu_F}+\varepsilon\Big(1+\log\frac{m}{c\varepsilon(1-\gamma)}\Big),
\end{align}
which is the stated bound.

\emph{Orders of magnitude.} Since $3c\le3m$, $\frac{3c\varepsilon}{56m^2}\le\bar w_g\le\frac{3c\varepsilon}{32m^2}$, i.e., $\bar w_g=\Theta(c\varepsilon/m^2)$, and $\bar w_c=\frac{3c^2\varepsilon^2}{64m^3\Lv}$. Hence $\mu=\Theta(\min\{\frac{c\varepsilon}{m^2d\Mv},\frac{c\varepsilon}{m\Lv}\})$, $\frac{1}{\mu\bar w_g}=\mathcal O(\frac{m^4d\Mv+m^3\Lv}{c^2\varepsilon^2})$, and
\begin{align}
M_c&=\mathcal O\Big(\frac{m^3\Lv L_2}{c^2\varepsilon^2}\Big),\qquad
M=\mathcal O\Big(\frac{m^4d^2\Mv L_2+m^3d\Lv L_2}{c^2\varepsilon^2}\Big)=\mathcal O\Big(\frac{m^4d^2}{c^2}\varepsilon^{-2}\Big),\\
N_c&=\mathcal O\Big(\frac{m^6\Lv^2(\Gbar+L)^2\iot}{c^4\varepsilon^4}\Big),\\
N&=\mathcal O\Big(\frac{\iot}{\bar w_g^2}\Big(\frac{d^2\Gbar^2H|\Ac|M_g}{\mu}+\frac{d^2L^2}{\mu^2M}\Big)+\frac{(\Gbar+L)^2\iot}{\Gbar^2H|\Ac|M_g\mu}\Big)
\notag\\
&=\widetilde{\mathcal O}\Big(\frac{m^6d^3}{c^3}\varepsilon^{-3}+\frac{m^4d^2}{c^2}\varepsilon^{-2}+\frac{m^2d}{c}\varepsilon^{-1}\Big)=\widetilde{\mathcal O}\Big(\frac{m^6d^3}{c^3}\varepsilon^{-3}\Big),\\
T&=\mathcal O\Big(\frac{M_gm^2}{\bar C\mu_F\varepsilon^2}\log\frac{\Gbar}{\varepsilon}\Big),
\end{align}
where for $N$ we used $\frac{d^2}{\mu}=\mathcal O(\frac{m^2d^3\Mv+md^2\Lv}{c\varepsilon})$ and $\mu^2M\ge\frac{\mu dL_2}{2\bar w_g}=\Omega(\frac{L_2}{\Mv})$ (from $M\ge\frac{dL_2}{2\mu\bar w_g}$ and $\mu\le\frac{\bar w_g}{2d\Mv}$). For $H$, using $\gamma^H\le e^{-(1-\gamma)H}$ and $\sqrt{1+x}\le e^{x/2}$, it suffices that $H\ge\frac{1}{1-\gamma}\log\frac{3}{(1-\gamma)\bar w_c}$ and $H\ge\frac{2}{1-\gamma}\log\frac{4M_g}{(1-\gamma)^{3/2}\bar w_g}$, so $H=\mathcal O\big(\frac{1}{1-\gamma}\log\frac{mdM_g\Lv}{c(1-\gamma)\varepsilon}\big)$. Each iteration uses $N_c$ base trajectories and $N$ perturbed trajectories, and $mN_cM_c+(1+m)NM$ human queries. Multiplying by $T$, the constraint margin channel contributes $TN_c=\widetilde{\mathcal O}(\frac{m^8}{\bar C\mu_Fc^4}\varepsilon^{-6})$ trajectories and $TmN_cM_c=\widetilde{\mathcal O}(\frac{m^{12}}{\bar C\mu_Fc^6}\varepsilon^{-8})$ queries, the gradient channel $TN=\widetilde{\mathcal O}(\frac{m^8d^3}{\bar C\mu_Fc^3}\varepsilon^{-5})$ trajectories and $T(1+m)NM=\widetilde{\mathcal O}(\frac{m^{13}d^5}{\bar C\mu_Fc^5}\varepsilon^{-7})$ queries. Here $\widetilde{\mathcal O}$ hides $\iot$, $\log\frac{\Gbar}{\varepsilon}$ and the problem constants.

\end{proof}

%%%%%%%%
\section{Experimental Details}
\label{app:experiments}

This appendix completes Section~\ref{sec:sync-data}. It specifies the environment, the simulated evaluators and the hyperparameters of both methods, reports the full results in Table~\ref{tab:results}, and discusses two variants of the estimator. 
% The code reproduces every number and figure of the paper.

\textbf{Environment.} The CMDP is generated with a fixed seed, with $P(\cdot \,\vert\,  s,a)\sim\mathrm{Dirichlet}(0.5,\dots,0.5)$ independently for each $(s,a)$, $r(s,a)\sim\mathrm{Unif}[0,1]$, $g_i(s,a)=\mathrm{clip}(0.4(1-r(s,a))+0.6\xi_i(s,a),0,1)$ with $\xi_i\sim\mathrm{Unif}[0,1]$, $\rho$ uniform, $\gamma=0.85$ and $b=(3,3,3)$. The constrained optimum $V^\star_r(\rho)=5.306$ is computed by the linear program over occupancy measures $q\in\R_+^{|\St||\Ac|}$ that maximizes $\langle q,r\rangle$ subject to $\sum_aq(s',a)-\gamma\sum_{s,a}P(s' \,\vert\,  s,a)q(s,a)=\rho(s')$ and $\langle q,g_i\rangle\ge b_i$. All three constraints are active at the optimum (utilities $(3.00,3.00,3.00)$), whereas the unconstrained optimum has $V_r=5.88$ and utilities $(2.34,2.64,2.60)$. The uniform policy has $V_r=3.87$ and constraint margins $(0.518,0.398,0.262)$.

\textbf{Sampler and simulated feedback.} Trajectories of length $H=40$ ($\epsilon_V(H)=0.01$) are generated by the seeded sampler of Assumption~\ref{assum:seeds}, with actions drawn by inverse-CDF sampling from a uniform $u_h$ and next states from a uniform $u'_h$. For ZO-LBPG the two trajectories of every pair in Phase~2 share the seed $(s_0,u_h,u'_h)_{h<H}$. Each pairwise query on $(\tau_1,\tau_0)$ and channel $\diamond$ is answered by $M$ evaluators with success probability $\sigma(R^H_\diamond(\tau_1)-R^H_\diamond(\tau_0))$, each absolute query on $\tau_0$ and constraint $i$ with probability $\sigma(R^H_{g_i}(\tau_0)-b_i)$, where $\sigma$ is the Bradley--Terry link, and the frequencies are inverted with \eqref{eq:ext-inv} on $\mathcal P_\sigma=[\sigma(-\Gbar),\sigma(\Gbar)]$. The same number of evaluators is used for all channels ($M_c=M$). The returns are used only to simulate the evaluators and to evaluate the policies, and neither method observes them.

\textbf{ZO-LBPG.} Algorithm~\ref{alg_zolbpg} with $\eta=0.05$, $\delta=0.05$ and data-driven confidence widths as in \citet{ni2025safe}, all with the confidence multiplier $\sqrt{\iot}\approx2.50$ of Section~\ref{sec:stepsize}. Specifically, $\wc$ is the empirical standard error of the $N_c$ constraint margin samples $\{\sinv_{c_i}(\hat p_{c_i,n})\}_n$ times $\sqrt{\iot}$, plus $\epsilon_V(H)$, and the width in $\overline\beta_i(t)$ is the empirical standard error of the $N$ projected samples $\{\frac{d}{\mu_t}\sinv_{g_i}(\hat p_{g_i,n})\langle\bv^{(t)}_n,\hat u_t\rangle\}_n$ times $\sqrt{\iot}$. The smoothness constant is $\Mv=0.5$, five times the largest Hessian norm of the value functions measured along the runs ($\le0.1$), and $\hat L=0.5$ upper-bounds the measured gradient norms. After Phase~1 the perturbation magnitude is set to $\mu_t=\min\{0.5,\min_i\underline\alpha_i(t)/\hat L\}$, floored at $0.05$ (Section~\ref{sec:stepsize}). The stopping rule was never triggered within $T=2000$ iterations.

\textbf{ZPGPD-HF.} Algorithm~2 of \citet{liu2026policy} extended to $m$ constraints, with $N$ independent directions, independently sampled trajectories in each pair, the same simulated evaluators, the softmax parameterization (no projection of $\bt$), $\mu=0.5$, one dual variable per constraint, $\lambda^{(0)}=0$, and dual projection onto $[0,\frac{2}{(1-\gamma)\xi}]$ with $\xi=0.262$ the initial Slater margin. The theoretical step sizes of \citet{liu2026policy} are stated for the direct parameterization and are too small to be useful here, so we tuned the primal and dual step sizes $(\psi_1,\psi_2\sqrt T)$ over $\{0.01,0.02\}\times\{1,4\}$ (one seed, $T\in\{1000,1500\}$) and selected $\psi_1=0.02$, $\psi_2=4/\sqrt T$, the setting whose average constraint margin over the second half of training is closest to zero, i.e., the one that comes closest to the average-feasibility guarantee of the method. All settings produce between $33\%$ and $83\%$ infeasible iterates, and those with smaller final gaps have average constraint margins between $-0.05$ and $-0.09$.

\begin{figure}[ht]
    \centering
    \includegraphics[width=0.7\linewidth]{figures/fig_legend.pdf}\\
    \includegraphics[width=0.35\linewidth]{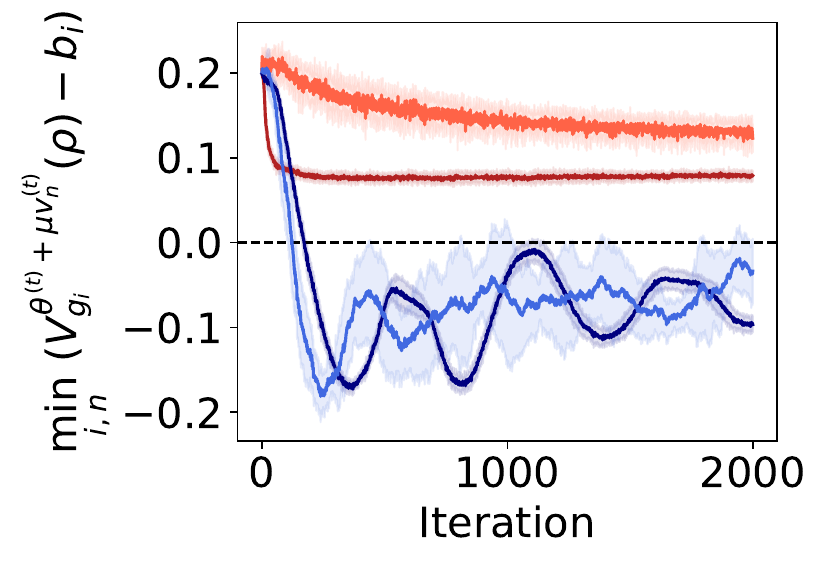}
    \caption{Minimum constraint margin over all policies executed for data collection at iteration $t$ (the $N$ exploration policies $\pi_{\bt^{(t)}+\mu_t\bv^{(t)}_n}$). Means over ten seeds with one-standard-deviation bands. ZO-LBPG keeps every executed policy feasible.}
    \label{fig:pert}
\end{figure}

\textbf{Results.} Table~\ref{tab:results} reports all runs after $T=2000$ iterations with ten seeds, and Figure~\ref{fig:pert} shows the smallest constraint margin over the exploration policies executed at each iteration. ZO-LBPG never violates a constraint, neither at its iterates nor at any of the $N$ exploration policies of each iteration, while ZPGPD-HF violates the constraints on about two thirds of its iterates in both regimes. The average constraint margin of ZPGPD-HF over the second half of training is close to zero, so it satisfies the constraints on average, exactly as its theory predicts.

\begin{table}[ht]
\centering
\footnotesize
\caption{Results after $T=2000$ iterations. ``Infeasible'' is the fraction of iterates that violate a constraint, ``Worst'' is the smallest constraint margin over all iterates and seeds, and ``Worst Executed'' is the smallest constraint margin over all executed exploration policies. All runs use 10 seeds. The last three rows are the variants discussed below.}
\label{tab:results}
\begin{tabular}{llccccc}
\toprule
Algorithm & Noise & Coupling & Infeasible & Worst & Worst Executed & Final Reward Gap\\
\midrule
ZO-LBPG & low & \ding{51} & $0.0\%$ & $+0.081$ & $+0.052$ & $0.58\pm0.05$\\
ZO-LBPG & high & \ding{51} & $0.0\%$ & $+0.127$ & $+0.048$ & $0.93\pm0.03$\\
ZPGPD-HF & low & \ding{55} & $66.6\%$ & $-0.118$ & $-0.192$ & $0.08\pm0.02$\\
ZPGPD-HF & high & \ding{55} & $68.9\%$ & $-0.179$ & $-0.242$ & $0.18\pm0.16$\\
\midrule
ZPGPD-HF & low & \ding{51} & $64.1\%$ & $-0.113$ & $-0.192$ & $0.08\pm0.01$\\
ZO-LBPG & low & \ding{55} & $0.0\%$ & $+0.088$ & $+0.057$ & $0.65\pm0.04$\\
ZO-LBPG, $N=2048$ & low & \ding{51} & $0.0\%$ & $+0.080$ & $+0.055$ & $0.48\pm0.02$\\
\bottomrule
\end{tabular}
\end{table}

\textbf{Coupling and batch size.} The last three rows of Table~\ref{tab:results} vary the estimator in the low-noise regime. Sampling the two trajectories of each pair of ZO-LBPG independently instead of coupling them, which is the estimator of prior work \citep{zhang2024zeroth,liu2026policy} with independent directions, raises the final gap from $0.58$ to $0.65$, since the wider confidence widths enter the step size quadratically through $\overline\beta_i(t)$, while safety is unaffected, as the theory predicts. Doubling the number of pairs to $N=2048$ lowers the gap to $0.48$ at the same level of safety. Both variants confirm that the reward performance of ZO-LBPG is governed by the width $\wg$ of the gradient estimate. Giving ZPGPD-HF the coupled estimator reduces its fraction of infeasible iterates only marginally (from $66.6\%$ to $64.1\%$) and leaves its gap unchanged, because the coupling reduces the variance of its gradient estimate but does nothing about the oscillations of the dual variables that drive its iterates across the boundary.

%%%%%%%%%%%%%%%%%%%%%%%%%%%%%%%%%%%%%%%%%%%%%%%%%%%%%%%%%%%%%%%%%%%%%%%%%%%%%%%%%%%%
%%%%%%%%%%%%%%%%%%%%%%%%%%%%%%%%%%%%%%%%%%%%%%%%%%%%%%%%%%%%%%%%%%%%%%%%%%%%%%%%%%%%
%%%%%%%%%%%%%%%%%%%%%%%%%%%%%%%%%%%%%%%%%%%%%%%%%%%%%%%%%%%%%%%%%%%%%%%%%%%%%%%%%%%%
%%%%%%%%%%%%%%%%%%%%%%%%%%%%%%%%%%%%%%%%%%%%%%%%%%%%%%%%%%%%%%%%%%%%%%%%%%%%%%%%%%%%
%%%%%%%%%%%%%%%%%%%%%%%%%%%%%%%%%%%%%%%%%%%%%%%%%%%%%%%%%%%%%%%%%%%%%%%%%%%%%%%%%%%%

%%%%%%%%%%%%%%%%%%%%%%%%%%%%%%%%%%%%%%%%%%%%%%%%%%%%%%%%%%%%%%%%%%%%%%%%%%%%%%%%%%%%
%%%%%%%%%%%%%%%%%%%%%%%%%%%%%%%%%%%%%%%%%%%%%%%%%%%%%%%%%%%%%%%%%%%%%%%%%%%%%%%%%%%%
%%%%%%%%%%%%%%%%%%%%%%%%%%%%%%%%%%%%%%%%%%%%%%%%%%%%%%%%%%%%%%%%%%%%%%%%%%%%%%%%%%%%
%%%%%%%%%%%%%%%%%%%%%%%%%%%%%%%%%%%%%%%%%%%%%%%%%%%%%%%%%%%%%%%%%%%%%%%%%%%%%%%%%%%%
%%%%%%%%%%%%%%%%%%%%%%%%%%%%%%%%%%%%%%%%%%%%%%%%%%%%%%%%%%%%%%%%%%%%%%%%%%%%%%%%%%%%

\section{Details of the LLM experiment}\label{app:llm}

\textbf{Model and sampling.} The base model Qwen2.5-3B-Instruct is kept frozen and sampled at temperature $0.7$ with
at most $128$ new tokens. Responses are generated by inverse-CDF sampling over the vocabulary in a fixed order, with
one uniform per token position drawn before generation, in single precision. We verified that with $\theta'=\theta$
coupled pairs coincide token by token, also across separate generation calls with different batch sizes, and the
fraction of identical pairs decreases smoothly with $\|\theta'-\theta\|$. Prompts are drawn uniformly from the
$32{,}374$ training prompts of PKU-SafeRLHF.

\textbf{Policy class.} Since the variance of the zeroth-order estimator grows with the dimension, the policy parameter
is low-dimensional. For each of the four targeted linear maps $W$ (the attention output projection and the MLP down
projection in layers $14$ and $18$ of the $36$-layer model), the adapter is $W+BRA$ with frozen
$B\in\mathbb{R}^{2048\times 2}$, $A\in\mathbb{R}^{1\times d_{\mathrm{in}}}$ and trainable $R\in\mathbb{R}^{2\times 1}$,
so that $d=8$. The factors are computed once from PKU-SafeRLHF training responses before training. The first column of
$B$ is the top principal direction of the per-pair difference of the layer output, averaged over response tokens,
between the unsafe and the safe response of $256$ pairs that differ in safety. The second column is the analogous
direction between the more and the less helpful response of $256$ pairs in which both responses are safe,
orthogonalized against the first. Each column is scaled to the root-mean-square size of its difference component. The
row of $A$ is the top uncentered principal direction of the module input on response tokens, scaled to unit
root-mean-square coefficient, so that $A$ acts as a nearly constant gain and a unit entry of $R$ shifts the activations
by about one typical contrastive difference. The factors use responses from the dataset only, never the reward or cost
models or the evaluators. With factors taken instead from the singular value decomposition of $W$, a
finite-difference probe showed that the helpfulness and harmlessness values are nearly constant in $\theta$ even when
almost every response changes, so we use data-informed factors. The initial parameter $\theta^{(0)}$ sets each entry
of $R$ that maps the input direction to the safety direction to $0.5$ (a unit vector along this axis) and the remaining
entries to $0$. At $\theta^{(0)}$ helpfulness and harmlessness conflict, so that ascending the reward alone decreases
the margin.

\textbf{Latent signals and evaluators.} The Beaver-7B reward and cost scores $s_r,s_c$ of a prompt--response pair are
mapped to returns $R_r=5\,\mathrm{sigmoid}((s_r+6)/2)$ and $R_g=5\,\mathrm{sigmoid}(-(s_c+2)/3)$ in $[0,5]$, with
centers and scales calibrated on responses of the base model. The pairwise and absolute channels
follow~\eqref{eq:channels} with the Bradley--Terry link $\sigma(x)=1/(1+e^{-x})$. Each query is answered by $256$
independent Bernoulli draws, and the answers are inverted with~\eqref{eq:hc}. The threshold $b=2.25$ gives the
initial margin $\bar V^{\theta^{(0)}}_{g}(\rho)\approx 0.89$.

\textbf{Hyperparameters.} ZO-LBPG: $\eta=0.05$; $\delta=0.1$, with confidence widths built from empirical standard
errors as in Section~\ref{sec:sync-data} (so $z=2.86$ standard errors); $M_V=1$, five times the largest curvature
measured by finite differences; perturbation radius $\mu_t=\max\{10^{-3},\min\{0.5,\min_i\alpha_i(t)/\hat L\}\}$ with
$\hat L=0.5$, which keeps every exploration policy within the certified margin. ZPGPD-HF uses the same estimator and
batches, primal step $0.15$, dual step $0.4$, and a dual variable projected onto $[0,50]$ and initialized at $0$; the
unconstrained reference is ZPGPD-HF with the dual variable fixed at $0$. Shared: $N=512$, $N_c=1024$, $M=M_c=256$,
$T=300$, and the same three seeds for both methods, so that both see the same prompts at every iteration. Each
iteration takes about two minutes on one H100 GPU.

\textbf{Evaluation.} The true values of each iterate are estimated from the oracle scores of its $N_c$ on-policy
responses (standard error about $0.05$), and the average value of its exploration policies from the $N$ perturbed
responses.

\textbf{Additional results.} Table~\ref{tab:llm} summarizes the runs. Ascending the helpfulness alone drives the
margin to about $-0.9$ within $60$ iterations. ZPGPD-HF oscillates around the boundary as its dual variable builds up
and decays, and its exploration policies are on average significantly infeasible at $26.6\%$ of the iterations.
ZO-LBPG raises the helpfulness from $2.39$ to $2.74$ while its margin decreases from $0.89$ and settles near $0.17$,
roughly one confidence width from the boundary. ZPGPD-HF reaches a higher average helpfulness ($2.94$ over the last
$100$ iterations), partly through its infeasible iterates; restricted to its feasible iterates, its average
helpfulness is $2.86$. The distance at which ZO-LBPG settles is set by the width of the margin confidence bounds,
consistent with the role of the margin-estimation cost in Corollary~\ref{cor:samples}.

\begin{table}[ht]
  \caption{LLM experiment after $T=300$ iterations (three seeds, one seed for the unconstrained reference).
  ``Infeasible'' is the fraction of iterates whose estimated margin is negative, with the fraction significantly
  negative at two standard errors in parentheses. ``Exploration'' is the fraction of iterations whose exploration
  policies have a significantly negative average margin. ``Worst'' is the smallest estimated margin over all iterates
  and seeds. ``Helpfulness'' is the mean over the last $100$ iterations, and ``Feasible'' restricts it to feasible
  iterates.}
  \label{tab:llm}
  \centering
  \small
  \begin{tabular}{lccccc}
    \toprule
    Algorithm & Infeasible & Exploration & Worst & Helpfulness & Feasible \\
    \midrule
    ZO-LBPG       & $0\%$ ($0\%$)          & $0\%$     & $+0.025$ & $2.74 \pm 0.01$ & $2.74$ \\
    ZPGPD-HF      & $43.5\%$ ($31.3\%$)    & $26.6\%$  & $-0.512$ & $2.94 \pm 0.04$ & $2.86$ \\
    Unconstrained & $91.7\%$ ($91.4\%$)    & $90.7\%$  & $-1.029$ & $3.52$          & ---    \\
    \bottomrule
  \end{tabular}
\end{table}

%%%%%%%%%%%%%%%%%%%%%%%%%%%%%%%%%%%%%%%%%%%%%%%%%%%%%%%%%%%%%%%%%%%%%%%%%%%%%%%%%%%%%%%%%%%%%%%%%%%%%%%%%%%%%%%%%%%%%%%%%%%%%%%%%%%%%%%%%%%%%%%%%%%%%%%%%%%%%%%%%%%%%%%%%%%%%%%%%%%%%%%%%%%%%%%%%%%%%%%%%%%%%%%%%%%%%%%%%%%%%%%%%%%%%%%%%%%%%%%%%%%%%%%%%%%%%%%%%%%%%%%%%%%%%%%%%%%%%%%%%%%%%%%%%%%%%%%%%%%%%%%%%%%%%%%%%%%%%%%%%%%%%%%%%%%%%%%%%%%%%%%%%%%%%%%%%%%%%%%%%%%%%%%%%%%%%%%%%%%%%%%%%

\end{document}